\documentclass[twoside,11pt]{article}

\usepackage{jmlr2e}

\usepackage[utf8]{inputenc}

\usepackage{url}
\usepackage[T1]{fontenc}
\usepackage{xcolor}
\usepackage{booktabs}       % professional-quality tables
\usepackage{amsfonts}       % blackboard math symbols
\usepackage{nicefrac}       % compact symbols for 1/2, etc.
\usepackage{microtype}      % microtypography
\usepackage{amssymb}
\usepackage{amsmath}
\usepackage{amsthm}
\usepackage{amsfonts}
\usepackage{natbib}
\usepackage{color}
\usepackage{graphicx,enumitem}
\usepackage{colordvi}
\usepackage{subfigure}
\usepackage{bbm}
\usepackage{algorithm}% http://ctan.org/pkg/algorithms
\usepackage{algorithmic}
\usepackage{multirow}
\usepackage{wrapfig}
\usepackage{makecell}
\usepackage{pifont}
\usepackage{hyperref}
\usepackage{cleveref}

\usepackage{mathtools}

\hypersetup{citecolor=blue}
\usepackage{natbib}
\jmlrheading{0}{0000}{0-00}{0/00}{00/00}{meila00a}{Soufiane Hayou, Arnaud Doucet, and Judith Rousseau}

\ShortHeadings{Exact convergence rates of the NTK regime}{Hayou, Doucet, and Rousseau}
\firstpageno{0}

\newcommand{\bigO}{\mathcal{O}}

\newtheorem{prop}{Proposition}
\newtheorem{thm}{Theorem}
\newtheorem{fact}{Fact}
\newtheorem{lemma}{Lemma}

\newtheorem{assumption}{Assumption}

\newtheorem{lemma2}{Appendix Lemma}
\newtheorem{lemma3}{Lemma}

\newcommand{\tickNo}{\textcolor{red}{\ding{55}}}
\newcommand{\tickYes}{\textcolor{blue}{\checkmark}}

\newcommand{\Sd}{\mathbb{S}^{d-1}}

\newcommand*{\email}[1]{%
    \normalsize\href{mailto:#1}{#1}\par
    }
\begin{document}

\title{Exact Convergence Rates of the Neural Tangent Kernel in the Large Depth Limit}
\author{\name Soufiane Hayou \email hayou@nus.edu.sg \\
       \addr Department of Mathematics\\
       National University of Singapore
       \AND
       \name Arnaud Doucet \email arnaud.doucet@stats.ox.ac.uk \\
       \addr Department of Statistics\\
       University of Oxford
       \AND
       \name Judith Rousseau \email judith.rousseau@stats.ox.ac.uk \\
       \addr Department of Statistics\\
       University of Oxford\\}

\editor{}

\maketitle

\begin{abstract}
Recent work by \citet{jacot} has shown that training a neural network using gradient descent in parameter space is related to kernel gradient descent in function space with respect to the Neural Tangent Kernel (NTK). \citet{lee_wide_neural_nets_linear_models} built on this result by establishing that the output of a neural network trained using gradient descent can be approximated by a linear model when the network width is large. Indeed, under regularity conditions, the NTK converges to a time-independent kernel in the infinite-width limit. This regime is often called the NTK regime. In parallel, recent works on signal propagation \citep{poole, samuel,hayou} studied the impact of the initialization and the activation function on \emph{signal propagation} in deep neural networks. In this paper, we connect these two theories by quantifying the impact of the initialization and the activation function on the NTK when the network depth becomes large. In particular, we provide a comprehensive analysis of the convergence rates of the NTK regime to the \emph{infinite depth} regime.\\
\end{abstract}

\section{Introduction}
The theoretical analysis of the generalization capacity of Deep Neural Networks (DNNs) has been the topic of a multitude of work; see e.g. \cite{simon,nguyen,zhang, zou}). Recently, there has been an growing interest in the implicit bias of DNNs in the infinite-width limit (e.g. infinite number of neurons per layer for a fully connected neural network and infinite number of channels for a convolutional neural network). \citet{jacot} introduced the Neural Tangent Kernel (NTK) that characterises DNN training with gradient flow. The infinite-width limit of these dynamics is referred to as the NTK regime or Lazy training regime. This nomenclature follows from the fact that, in this regime, the NTK remains constant (with respect to training time) during training and the training procedure is well approximated by a first order Taylor expansion of the output function near its initial value. \cite{lee_wide_neural_nets_linear_models} demonstrated that such a simple model could lead to surprisingly good performance. However, most experiments in the NTK regime have been conducted with shallow neural networks and do not extend to DNNs. In this paper, we address this topic by establishing the limitations of the NTK regime for DNNs, showing how it differs from the actual training of DNNs with Stochastic Gradient Descent. 
\paragraph{Neural Tangent Kernel.}
\citet{jacot} showed that training a neural network (NN) with GD (Gradient Descent) in parameter space is equivalent to GD in a function space with respect to the NTK. \citet{simon2} used a similar approach to prove that full batch GD converges to global minima for shallow neural networks, and \citet{karakida2018universal} linked the Fisher information matrix to the NTK, studying its spectral distribution for infinite-width NN. The infinite-width limit for different architectures was studied by \citet{yang2019scaling}, who introduced a tensor formalism that can express neural computations for a variety of architectures. \citet{lee_wide_neural_nets_linear_models} studied a linear approximation of the full batch GD dynamics based on the NTK, and gave a method to approximate the NTK for different architectures. Finally, \citet{arora2019exact} proposed an efficient algorithm to compute the NTK for convolutional architectures (Convolutional NTK). In all of these papers, the authors only studied the effect of the infinite-width limit (NTK regime) on relatively shallow networks. 
\paragraph{Information propagation.} 
In parallel, information propagation in infinite-width DNNs has been studied by \cite{hayou,lee_nngp,samuel,yang2017meanfield}. These works provide an analysis of the signal propagation at initialization. In particular, they study the impact of the choice of initialization hyper-parameters (variance of the initial random weights). They identify a set of hyper-parameters known as the Edge of Chaos (EOC) and a class of activation functions ensuring `deeper' propagation of the information carried by the input, so that the network output still carries some information about the input. In this paper, we prove that the EOC initialization has also some benefits on the NTK at initialization, which, in the infinite-width limit, remains constant during training.
% \begin{table}
% \caption{\small{"Does the model learn?". We train a FeedForward Neural Network on MNIST using both standard SGD training and NTK training defined in section \ref{section:ntk}. For Shallow networks, both SGD and NTK yield good performance (See section \ref{section:experiments}). However, for Deep networks, the NTK training yields trivial accuracy of around $\sim 10 \%$ for any initialization scheme.}}
% \label{table:summary_of_findings}
% \begin{center}
% \begin{small}
% \resizebox{8.cm}{!}{
% \begin{tabular}{lccccr}
% & &  \makecell{Initialization on\\ the Edge of Chaos} & Other Initialization \\
% \midrule
% \multirow{2}{*}{\makecell{Shallow Network\\ (depth $L=3$)}} & NTK & \tickYes &  \tickYes \\
% \cmidrule{2-4}
% &SGD  & \tickYes &  \tickYes \\
% \midrule
% \multirow{2}{*}{\makecell{Medium Network\\ (depth $L=30$)}} & NTK & \tickYes &  \tickNo \\
% \cmidrule{2-4}
% &SGD  & \tickYes &  \tickNo \\
% \midrule
% \multirow{2}{*}{\makecell{Deep Network\\ (depth $L=300$)}} & NTK  & \tickNo &  \tickNo \\
% \cmidrule{2-4}
% &SGD   & \tickYes &  \tickNo \\
% \midrule
% \end{tabular}
% }
% \end{small}
% \end{center}
% \vspace{-1.cm}
% %\vskip -0.1in
\paragraph{NTK training and SGD training.} Stochastic Gradient Descent (SGD) has been successfully used to train DNNs. Following the introduction of the NTK in \cite{jacot}, \citet{lee_wide_neural_nets_linear_models} suggested a different approach to training overparameterized neural networks. The idea originates from the conjecture that, in overparameterized models, a local minimum exists near the initial weights. Thus, using a first order Taylor expansion at these initial weights, the NN reduces to a simple linear model, and this linear model is trained\footnote{With quadratic loss, the linear model has an analytic solution and no training is required.} instead of the original NN. Hereafter, we refer to this training procedure as the \emph{NTK training} and the trained model as the \emph{NTK regime}. If we replace SGD with the (continuous-time) gradient flow and consider the infinite-width limit, then these two training regimes become identical under some mild conditions. We provide further details in Section \ref{section:ntk}. 
\paragraph{Contributions.} The NTK is usually studied in the context of finite depth and infinite-width. The large depth limit is often overlooked. Our aim in this paper is to provide a comprehensive theoretical analysis of the NTK regime in the large depth limit. This limit should be interpreted as infinite-width-then-infinite-depth limit of the NTK. Our contributions are as follows
\begin{itemize}
    \item We prove that the NTK regime is always trivial in the sense that the NTK is degenerate in the infinite depth limit. However, the convergence rate to this trivial regime is controlled by the initialization hyper-parameters. 
    \item We prove that for some standard neural architectures (Fully-connected and convolutional NNs), the EOC initialization provides sub-exponential convergence rates to this trivial regime, while other initialization schemes yield an exponential rate. More precisely, we obtain an exact convergence rate of $\Theta(\log(L) L^{-1})$ when the hyper-parameters are chosen in the EOC, where the notation $g(x) = \Theta(m(x))$ means there exist two constants $A,B > 0$ such that $ A \, m(x) \leq g(x) \leq B \, m(x)$. On the other hand, this rate becomes $\bigO(e^{-\gamma L}))$ when the initialization hyper-parameters are not in the EOC. 
    For the same depth $L$, the NTK regime is thus \emph{``less trivial''} for an EOC initialization.
    
    \item We characterize the convergence rate of the NTK for ResNet architectures. We show that, contrary to fully-connected and convolution NNs, this rate is always  $\Theta(\log(L) L^{-1})$ irrespective of the initialization.
    \item We leverage our theoretical results on the asymptotic behaviour of the NTK to obtain practical guidelines for the choice of the learning rate for SGD training. We establish the existence of a learning rate \emph{passband} for SGD training; any choice of the learning rate outside of this passband yields poor performance.
\end{itemize}
\begin{table}
\caption{\small{``Does the model learn?''. We train a Feedforward NN on CIFAR10 using both standard SGD training and NTK training defined in Section \ref{section:ntk}. For shallow networks, both SGD and NTK yield good performance (See Section \ref{section:experiments}). However, for deep networks, the NTK training yields trivial accuracy of around $\sim 10 \%$ for any initialization scheme.}}
\label{table:summary_of_findings}
\begin{center}
\begin{small}
\resizebox{6.8cm}{!}{
\begin{tabular}{lccccr}
& &  EOC Init & Other Init \\
\midrule
\multirow{2}{*}{$L=3$} & NTK & \tickYes &  \tickYes \\
\cmidrule{2-4}
&SGD  & \tickYes &  \tickYes \\
\midrule
\multirow{2}{*}{$L=30$} & NTK & \tickYes &  \tickNo \\
\cmidrule{2-4}
&SGD  & \tickYes &  \tickNo \\
\midrule
\multirow{2}{*}{$L=300$} & NTK  & \tickNo &  \tickNo \\
\cmidrule{2-4}
&SGD   & \tickYes &  \tickNo \\
\midrule
\end{tabular}
}
\end{small}
\end{center}
%\vskip -0.1in
\end{table}
Table \ref{table:summary_of_findings} summarizes the behaviour of NTK and SGD training for different depths and initialization schemes of a Fully-connected Feedforward Neural Network (FFNN) on the CIFAR10 dataset. We show whether the model test accuracy is significantly larger than $10\%$, which is the accuracy of the trivial random classifier. The results displayed in this table illustrate that for shallow FFNN ($L =3$), the model learns to classify with both NTK training and SGD training for any initialization scheme. For a medium depth network ($L=30$), NTK training and SGD training both succeed in training the model with an initialization in the EOC, while they both fail with other initializations. It has already been observed that, with SGD, an EOC initialization is beneficial for the training of DNNs \citep{hayou, samuel}. Our results show that the EOC initialization is also beneficial for NTK training (Section \ref{section:ntk}).
However, for a deeper network with $L = 300$, NTK training fails for any initialization, while SGD training succeeds in training the model with EOC initialization. This confirms the limitations of NTK training for DNNs, suggesting that parameter updates might play a more significant role in generalization of DNNs compared to the implicit architectural bias\footnote{We refer to the implicit bias solely induced by the architecture and not the training algorithm.}. However, although the large depth NTK regime is trivial, we leverage this asymptotic analysis to obtain a theoretical upper bound on the learning rate (Section \ref{section:learning_rate}).
We illustrate our theoretical results through extensive simulations. All the proofs are detailed in the appendix.

\paragraph{Related work.} After the first version of this paper was made publicly available (\cite{hayou2019meanfield}), similar albeit weaker results on the behaviour of the NTK in the large depth limit have been derived independently. \cite{xiao2020disentangling} studied the effect of the initialization on the trainability/generalization of the NTK regime in the case of FFNN, via the condition number of the kernel matrix. Their results agree with ours in the case of FFNN. However, we also provide here uniform convergence rates of the kernel. \cite{huang_ntk} considered a scaled version of ResNet (where the blocks are scaled) with ReLU and proved an upper bound on the convergence rate of order $\mathcal{O}\left(\frac{\text{polylog} L}{L}\right)$, while our results give the exact rate of $\Theta(\log(L) L^{-1})$ for both ReLU and Tanh. Note that for NTK training, a lower bound on the rate is more useful than an upper bound as we want to ensure that the convergence rate of the NTK to the trivial infinite-depth regime is not fast. Our $\Theta(\log(L) L^{-1})$ rate has the advantage of being both an upper and a lower bound. More generally, scaling ResNet was shown to be an effective way to avoid degeneracy in the large depth limit \citep{hayou2020stable}. Other papers studied the spectrum of the NTK for fixed depth $L$, e.g. \citep{bietti2021deep, chen2021deep}.
\section{Neural Networks and Neural Tangent Kernel}\label{section:ntk}
%\subsection{Setup and notations: the infinite-width limit }
Consider a neural network consisting of $L$ layers of widths $(n_l)_{1 \leq l \leq L}$, $n_0 = d $ , and let $\theta = (\theta^l)_{1 \leq l \leq L}$ be the flattened vector of weights and bias indexed by the layer's index, and $p$ be the dimension of $\theta$.  
The output function $f$ of the neural network is given by some mapping $s : \mathbb{R}^{n_L} \rightarrow \mathbb{R}^o$ of the last layer $y^L(x)$; $o$ being the dimension of the output (e.g. number of classes for a classification problem). For any input $x \in \mathbb{R}^d$, we thus have $f(x, \theta) = s(y^L(x)) \in \mathbb{R}^o$. As we train the model, $\theta$ changes with time $t$, and we denote by $\theta_t$ the value of $\theta $ at training time $t$ and $f_t(x) = f(x, \theta_t)$. Let $\mathcal{D} = \{(x_i, z_i) : 1 \leq i \leq N\}$ be a general dataset, and let $\mathcal{X} = (x_i)_{1 \leq  i \leq N}$, $\mathcal{Z} = (z_j)_{1 \leq  j \leq N}$ be the sequences of inputs and outputs respectively.  For all functions $g: \mathbb R^{d \times o} \rightarrow \mathbb R^k$ we denote by $g(\mathcal X, \mathcal Z)$ the matrix in $\mathbb R^{k \times N} $ with entries $g(x_i,z_i), i \in [N] \overset{\small{def}}{=} \{1, \dots, N\}$.\\
% We assume that there is no colinearity in the input dataset $\mathcal{X}$.
% \begin{assumption}\label{assumption:collinearity}
% There is no two inputs $x,x' \in \mathcal{X}$ such that $x' = \alpha x$ for some $\alpha \in \mathbb{R}$. 
% \end{assumption}
% \begin{assumption}[Compact dataset]
% There exists a compact set $E \subset \mathbb{R}^d$ such that $\mathcal{X} \subset E$.
% \end{assumption}

The NTK is defined as the $o \times o$ dimensional kernel $K^L_{\theta}$ given for all $x,x'  \in \mathbb R^d$ by
\begin{equation*}
   K^L_{\theta_t}(x, x') = \nabla_{\theta} f(x, \theta_t) \nabla_{\theta} f(x', \theta_t)^T = \sum_{l=1}^L \nabla_{\theta^l} f(x, \theta_t) \nabla_{\theta^l} f(x', \theta_t)^T \in \mathbb R^{o \times o}. 
\end{equation*}
\\
\cite{jacot} studied the dynamics of the output of the NN as a function of the training time $t$ when the network is trained using a GD algorithm. For parameters $\theta$, the empirical loss is given by $\mathcal{L}(\theta) = \frac{1}{N} \sum_{i=1}^N \ell(f(x_i, \theta), y_i)$. The GD updates are given by
\begin{equation} \label{equ:discrete_gd1}
    \hat{\theta}_{t+1} = \hat{\theta}_t - \eta \nabla_{\theta} \mathcal{L}(\hat{\theta}_t),
\end{equation}
where $\eta>0$ is the learning rate.\\
Let $T>0$ be the training time and $N_s = T/\eta$ be the number of steps of the discrete GD \eqref{equ:discrete_gd1}. \cite{jacot} studied the continuous time limit of \eqref{equ:discrete_gd1}, often called gradient flow, with discretization step $\Delta t = \eta $, which yields
\begin{equation}\label{equa:continuous_time_gd1}
    d{\theta_t} = - \nabla_{\theta} \mathcal{L}(\theta_t)dt.
\end{equation}
This can be written as 
\begin{equation*}
    d\theta_t = - \frac{1}{N} \nabla_{\theta} f(\mathcal{X}, \theta_t)^T \nabla_{z}\ell(f(\mathcal{X}, \theta_t), \mathcal{Z}) dt,
\end{equation*}
where %$f(\mathcal{X}, \theta_t) \in \mathbb{R}^{N\times o}$ is the flattened vector of $(f(x, \theta_t))_{x \in \mathcal{X}}$ and
$\nabla_{\theta} f(\mathcal{X}, \theta_t)$ is a matrix of dimension $oN \times p $ and $\nabla_{z}\ell(f(\mathcal{X}, \theta_t), \mathcal{Z}) $ is the flattened vector of dimension $ oN$ constructed from the concatenation of the vectors $\nabla_z \ell(z,y_i)_{| z = f(x_i, \theta_t)}, i\leq N$. 
As a result, the output function $f_t(x) = f(x,\theta_t) \in \mathbb{R}^o$ satisfies the following ordinary differential equation (ODE)
\begin{equation}\label{equation:f_t_dynamics1}
    df_t(x)  = - \frac{1}{N} \nabla_{\theta} f(x, \theta_t) \nabla_{\theta} f(\mathcal{X}, \theta_t)^T \nabla_{z}\ell(f_t(\mathcal{X}), \mathcal{Z}) dt.
\end{equation}
Given the dataset $\mathcal{D}$, the NTK matrix at training time $t$, denoted by $K^L_{\theta_t}(\mathcal{X}, \mathcal{X})$, is the $oN \times oN$ matrix defined blockwise by 
$$ K^L_{\theta_t}(\mathcal{X}, \mathcal{X}) 
 = \left( \begin{array}{cccc}
 K^L_{\theta_t}(x_1, x_1) &    \cdots & K^L_{\theta_t}(x_1, x_N) \\
 K^L_{\theta_t}(x_2, x_1)  & \cdots & K^L_{\theta_t}(x_2, x_N)\\
 \vdots & \ddots & \vdots \\
K^L_{\theta_t}(x_N, x_1) &   \cdots & K^L_{\theta_t}(x_N, x_N)
 \end{array}\right). 
$$\\
By applying  Eq. \eqref{equation:f_t_dynamics1} to the vector $\mathcal{X}$, we obtain
\begin{equation}\label{equation:infinite_width_dynamics0}
    df_t(x_j) = - \frac{1}{N}  K_{\theta_t}^L(x_j, \mathcal{X}) \nabla_{z}\ell(f_t(\mathcal{X}), \mathcal{Z}) dt,\quad j =1,\cdots, N.
\end{equation}
In the case of FFNN, \cite{jacot} proved that with gradient descent,  $ K_{\theta_t}^L$ converges to a kernel function $K^L$ (independent on $\theta_t$) for all $t\leq T$ (where $T$ is an upper bound on the training time) as the widths $n_1, \cdots, n_L$ go to infinity sequentially, i.e. $n_1 \to \infty$, then $n_2 \to \infty$, $\dots$, then $n_L \to \infty$. This was later generalized by \cite{yang_tensor3_2020} to the case where the limits can be taken simultaneously. This result holds under the technical assumption $\int_0^T\|\nabla_{z}\ell(f_t(\mathcal{X}), \mathcal{Z})\|_2dt<\infty $ a.s. with respect to the initialization weights. In this limit, Eq. \eqref{equation:infinite_width_dynamics0} becomes
\begin{equation}\label{equation:infinite_width_dynamics}
 df_t(x_j) = - \frac{1}{N}   K^L(x_j, \mathcal{X}) \nabla_{z}\ell(f_t(\mathcal{X}), \mathcal{Z}) dt,\quad j \in [N].   
\end{equation} 
We note hereafter $\hat{K}^L = K^L(\mathcal{X}, \mathcal{X})$.
For instance, with the quadratic loss $\ell(z',z) = \frac{1}{2} ||z'-z||^2$, Eq. \eqref{equation:infinite_width_dynamics} becomes
\begin{equation}\label{equation:f_t_SDE_quadratic}
df_t(x_j) = - \frac{1}{N} \hat{K}^L (f_t(\mathcal{X}) -  \mathcal{Z}) dt , \quad j \in [N],
\end{equation}
which is the ODE of a simple linear model that has the following closed-form solution
\begin{equation}\label{equation:infinite_width_dynamics_squared_loss}
    f_t(\mathcal{X}) = e^{-\frac{t}{N} \hat{K}^L } f_0(\mathcal{X}) + (I - e^{-\frac{t}{N} \hat{K}^L }) \mathcal{Z}.
\end{equation}
For general input $x \in \mathbb{R}^d$ (not necessarily in the dataset), we have
\begin{equation}\label{equation:generalization_formula}
f_t(x) = f_0(x) + \gamma(x, \mathcal{X}) (I - e^{-\frac{t}{N} \hat{K}^L }) (\mathcal{Z} - f_0(\mathcal{X}))
\end{equation}
where $\gamma(x) = K^L(x, \mathcal{X}) (\hat{K}^L )^{-1}$.\\
The infinite-width limit of the training dynamics is given by Eq. \eqref{equation:infinite_width_dynamics}, which leads to Eq. \eqref{equation:generalization_formula} under the quadratic loss. 
For other loss functions such as the cross-entropy loss (often used for classification tasks), some approximations can be used to obtain the infinite-width limit (see \cite{lee_wide_neural_nets_linear_models} for more details). These approximations are implemented in the Python package Neural-Tangents \citep{neuraltangents2020}. Hereafter we refer to $f_\infty$, i.e. the limit of $f_t$ in Eq. \eqref{equation:f_t_SDE_quadratic} as $t$ goes to infinity by the \emph{NTK regime solution} or simply the \emph{NTK regime} when there is no confusion.

%\paragraph{Role of the NTK in NTK training.} 
As observed in \citet{du2018gradient}, the convergence rate (w.r.t time) of $f_t$ to $f_\infty$ (infinite training time) is given by the smallest eigenvalue of $\hat{K}^L$. If the NTK becomes singular in the large depth limit, then the NTK training fails (the kernel matrix $\hat{K}^L$ converges to a singular matrix).
Importantly,  since $K^L$ is constant during training, its value is fixed at initialization. Therefore, we would naturally expect the initialization hyper-parameters to impact the infinite-width NTK; in Section \ref{section:large_depth_ntk_behaviour}, we prove several results on this link between initialization and the NTK. 

%\paragraph{NTK regime is scale invariant.} 
An important aspect which will be of interest in the interpretation of our theoretical results is that, with the quadratic loss, the NTK regime solution $f_\infty$ is \emph{scale invariant} in the sense that $f_\infty$ does not change if we scale the kernel by some scalar. In particular, this scalar can be depth-dependent, i.e. for any scaling factor $a_L$,
\begin{equation}\label{eq:scale_invariance1}
\gamma(x, \mathcal{X}) = K^L(x, \mathcal{X}) (\hat{K}^L )^{-1} = (K^L(x, \mathcal{X})/ a_L) (\hat{K}^L /a_L )^{-1} .
\end{equation}
Thus, \emph{studying the NTK regime with kernel $K^L$ is equivalent to studying the NTK regime with any scaled kernel $K^L/a_L$}, as both kernels yield the same solution $f_\infty$. In Theorems \ref{thm:ntk_eoc} and \ref{thm:ntk_residual}, we study scaled kernels to mitigate an exploding kernel effect in the limit of large depth. With the scale-invariance property stated above, the NTK regime solution remains unchanged in this case. 

% \begin{lemma}
% \label{lemma:invertibility_ntk}
% Assume the network is initialized with Gaussian weights and the mapping $s$ is such that $s(\mathbb{R}^{n_L})$ is not a subset in any hyperplane of $\mathbb{R}^o$. Then, with dynamics (\ref{equation:infinite_width_dynamics_squared_loss}) we have\\
% $\bullet$~$||f_t(\mathcal{X}) - \mathcal{Z}||$ converges almost surely to 0 as $t \rightarrow \infty$ if $\hat{K}^L$ is non-singular.\\
% $\bullet$~If $\hat{K}^L$ is singular, then almost surely, there exists a constant $C>0$ such that for all $t>0$, 
% $$
% ||f_t(\mathcal{X}) - \mathcal{Z}|| \geq C.
% $$
% \end{lemma}
% Lemma \ref{lemma:invertibility_ntk} shows that an invertible NTK is crucial in order for $f_t(\mathcal{X})$ to converge to $\mathcal{Z}$. More generally,
Another important aspect of the NTK regime is the generalization properties of $f_\infty$. From Eq. \eqref{equation:generalization_formula}, we can see that the term $\gamma$ plays a crucial role in the generalization capacity of the NTK regime. Different works \citep{du2018gradient,arora_fine_grained} have demonstrated that the inverse of the NTK matrix plays a crucial role in the generalization error of wide NN. Indeed, \citet{cao} proved that training a network of depth $L$ in the Lazy training regime yields a generalization bound of the form $\mathcal{O}(L\sqrt{\mathcal{Z}^T (\hat{K}^L)^{-1} \mathcal{Z} / N})$ in the infinite-width limit, where $\mathcal{Z}$ is the vector of training data outputs. Moreover, Eq. \eqref{equation:generalization_formula} shows that the Reproducing Kernel Hilbert Space (RKHS) generated by $K^L$ controls $f_t$. To see this, observe that for any $t \in (0,T)$ we can deduce from Eq. \eqref{equation:generalization_formula} that there exist coefficients $a_1, ..., a_N \in \mathbb{R}$ such that for all $x\in \mathbb{R}^d$, $f_t(x) - f_0(x) = \sum_{i=1}^N a_i K^L(x_i,x)$, showing that the `training residual' $f_t - f_0$ belongs to the RKHS of the NTK. In other words, with NTK training, $K^L$ controls what the network learns beyond initialization.

\section{Asymptotic Neural Tangent Kernel}\label{section:large_depth_ntk_behaviour}
In this section, we study the behaviour of $K^L$ as $L$ goes to $\infty$. We prove that $K^L$ converges in this limit to a degenerate kernel (i.e. rank-one kernel) and therefore the NTK regime becomes trivial. However, with an EOC initialization, this convergence is slow, which allows the use of NTK training for deeper NNs (NTK training is successful with $L = 30$, see \Cref{section:experiments}). However, since the limiting NTK is trivial, NTK training necessarily fails for large depth DNNs. We restrict our analysis to the case where the output dimension is 1 ($o=1$), the generalization of our results to multi-dimensional outputs is straightforward.

\subsection{NTK parameterization and the Edge of Chaos: FFNN and CNN architectures}
Let $\phi$ be the activation function (e.g. ReLU, Tanh etc.). We first establish results for the following architectures:

\paragraph{Feedforward Fully-connected Neural Network (FFNN).} Given depth $L$, widths $(n_l)_{1\leq l \leq L}$, weights $w^l $ and bias $b^l$. For some input $x \in \mathbb{R}^d$, the forward propagation of the FFNN using the NTK parameterization\footnote{The NTK parameterization introduces the factors $1/\sqrt{n_l}$ in front of the weights, this is necessary for the infinite-width limit.} is given by
\begin{equation}\label{equation:ffnn_net}
\begin{aligned}
y^1_i(x) &= \frac{\sigma_w}{\sqrt{d}}\sum_{j=1}^{d} w^1_{ij} x_j + \sigma_b b^1_i, \\
y^l_i(x) &= \frac{\sigma_w}{\sqrt{n_{l-1}}}\sum_{j=1}^{n_{l-1}} w^l_{ij} \phi(y^{l-1}_j(x)) + \sigma_b b^l_i, \quad  2\leq l \leq L,
\end{aligned}
\end{equation}
where $\sigma_b, \sigma_w \geq 0$ are constants. In this standard parameterization of FFNNs, $\sigma_w/\sqrt{n_{l-1}}$ and $\sigma_b$ are absorbed in the weights, and $(\sigma_b, \sigma_w)$ are called \emph{initialization hyper-parameters}. In the NTK parameterization above, they are just constants in front of the weights. However, since these constants essentially represent the scale of the initialization weights, we preserve the nomenclature `initialization hyper-parameters' to refer to $(\sigma_b,\sigma_w)$.
\paragraph{Convolutional Neural Network (CNN).} Consider a 1-dimensional convolutional neural network of depth $L$, denoting by $[m:n]$ the set of integers $\{m,m+1, ..., n\}$ for $m\leq n$, the forward propagation is given by
\begin{equation}\label{equation:convolutional_net}
\begin{aligned}
y^1_{i,\alpha}(x) &= \frac{\sigma_w}{\sqrt{v_1}} \sum_{j=1}^{n_{0}} \sum_{\beta \in ker} w^1_{i,j,\beta} x_{j,\alpha+\beta} + \sigma_b b^1_i,\\
y^l_{i,\alpha}(x) &= \frac{\sigma_w}{\sqrt{v_l}} \sum_{j=1}^{n_{l-1}} \sum_{\beta \in ker} w^l_{i,j,\beta} \phi(y^{l-1}_{j,\alpha+\beta}(x)) + \sigma_b b^l_i, \quad  2\leq l \leq L,
\end{aligned}
\end{equation}
where $i \in [n_l]$ is the channel number, $\alpha \in [0:M-1]$ is the neuron location in the channel, $n_l$ is the number of channels in the $l^{th}$ layer, and $M$ is the number of neurons in each channel, $ker = [-k: k]$ is a filter with size $2k +1$ and $v_l = n_{l-1}(2k + 1)$. Here, $w^l \in \mathbb{R}^{n_l\times n_{l-1}\times(2k+1)}$. We assume periodic boundary conditions, so that $y^l_{i, \alpha}=y^l_{i, \alpha+M}=y^l_{i, \alpha-M}$, and for $l=0$, $x_{i,\alpha+M_0}=x_{i,\alpha}=x_{i,\alpha-M_0}$. For the sake of simplification, we only consider the case of 1D CNN, the generalization to a $m$-dimensional CNN for $m\in \mathbb{N}$ is straightforward.\\
% the scaling factor becomes $\frac{\sigma_b}{\sqrt{n_{l-1} (2k+1)^2}}$.
Hereafter we write $y^l(x) = \mathcal{F}(w^l, y^{l-1}(x))$ for either recursions \eqref{equation:ffnn_net} or \eqref{equation:convolutional_net}. 

We also denote by $x\cdot x'$ the inner product in $\mathbb{R}^d$ for $x,x' \in \mathbb{R}^d$ and for $x,x' \in \mathbb{R}^{n_0 \times (2k+1)}$, let $[x,x']_{\alpha,\alpha'}$ be a convolutional mapping defined by 
$$[x,x']_{\alpha, \alpha'}=\sum_{j=1}^{n_0}\sum_{\beta \in ker_0} x_{j, \alpha+\beta} x_{j, \alpha'+\beta}^{'}.$$

For both architectures, we initialize the weights and bias with $w^l_{ij}, b^l_i \stackrel{\text{iid}}\sim \mathcal{N}(0, 1)$, where $\mathcal{N}(\mu, \sigma^{2})$ denotes the Gaussian distribution of mean $\mu$ and variance $\sigma^{2}$. 
we are interested in infinite-width regime, which corresponds to an infinite number of neurons for fully-connected layers , and an infinite number of channels for convolutional layers, i.e. $\min_{l\geq 1}n_l$ goes to infinity. 

A classical result in the literature of signal propagation at initialization is the following: in the infinite-width limit, the neurons $(y^l_i(.))_{i,l}$ converge to zero-mean Gaussian processes, as proved by \cite{neal,lee_nngp, matthews, hayou, samuel}; hence, their covariance kernels characterize their behaviour. Hereafter, we denote by $q^l(x,x')$ (resp. $q^l_{\alpha,\alpha'}(x,x')$) the covariance between $y^l_1(x)$ and $y^l_1(x')$ (resp. $y^l_{1,\alpha}(x)$ and $y^l_{1,\alpha'}(x')$). The choice of the neuron index 1 is arbitrary since different neurons in the same layer of a FFNN are identically distributed (a similar result holds in CNNs for the channel index instead of the neuron index).
We denote the corresponding correlations by $c^l(x,x')$ for FFNN and $c^l_{\alpha,\alpha'}(x,x')$ for CNN. In particular, we easily obtain
$q^1(x,x') = \sigma_b^2 + \frac{\sigma_w^2}{d} x\cdot x'$ for FFNN and $q^1_{\alpha,\alpha'}(x,x') = \sigma_b^2 + \frac{\sigma_w^2}{n_0 (2k+1)} [x,x']_{\alpha,\alpha'}
$ for CNN.

 \citet{jacot} established a recursive formula for the infinite-width NTK of  an FFNN when $\sigma_w = 1$. We generalize the result to any $\sigma_w>0$ in the next Lemmas, and Lemmas   \ref{lemma:cnn_ntk}, \ref{lemma:resnet_ntk} and \ref{lemma:resnet_ntk_conv} generalize it to other architectures as well. These generalizations are quite straightforward and can be established using the same induction as in \cite{jacot}. 
 
\begin{lemma} [Generalization of Theorem 1 in \cite{jacot}]\label{lemma:ffnn_ntk}
Consider an FFNN of the form \eqref{equation:ffnn_net}. Then, as $n_1, n_2, ..., n_{L-1} \rightarrow \infty$ sequentially, we have for all $x,x' \in \mathbb{R}^d$, $i,i' \leq n_L$, $K^L_{ii'}(x,x') = \delta_{ii'} K^L(x,x')$, where $K^L(x,x')$ is given by the recursive formula
$$
K^{L} (x,x') = \dot{q}^L(x,x') K^{L-1} (x,x') + \hat{q}^L(x,x'), \quad \textrm{where}
$$
$\hat{q}^{l} (x,x')= \sigma_b^2 + \sigma_w^2 \mathbb{E}[\phi(y_1^{l-1}(x))\phi(y_1^{l-1}(x'))]$ and $\dot{q}^l(x,x') = \sigma_w^2 \mathbb{E}[\phi'(y_1^{l-1}(x))\phi'(y_1^{l-1}(x'))]$.
\end{lemma}
Lemma \ref{lemma:ffnn_ntk} is proved in Appendix \ref{proofsSection3}. \\
For CNNs, the next lemma shows that a similar recursion for the infinite-width NTK holds. The infinite-width for CNNs is obtained by taking the number of channels to infinity.
\begin{lemma}[Infinite-width NTK of a CNN]\label{lemma:cnn_ntk}
Consider a CNN of the form \eqref{equation:convolutional_net}, then we have that for all $x,x' \in \mathbb{R}^d$, $i,i' \leq n_1$ and  $\alpha,\alpha' \in [0: M - 1]$
$$
K^1_{(i,\alpha), (i',\alpha')}(x,x') = \delta_{ii'} \left(\frac{\sigma_w^2}{n_0(2k+1)} [x,x']_{\alpha, \alpha'} + \sigma_b^2 \right).
$$
For $l\geq2$, as $n_1, n_2, ..., n_{l-1} \rightarrow \infty$ sequentially, we have for all $i,i' \leq n_l$, $\alpha,\alpha' \in [0:M - 1]$,  $K^{l}_{(i,\alpha),(i',\alpha')}(x,x') = \delta_{ii'} K^{l}_{\alpha,\alpha'}(x,x')$, where $K^{ l}_{\alpha,\alpha'}$ is given by the recursion 
\begin{align*}
    K^l_{\alpha,\alpha'} &= \frac{1}{2k+1} \sum_{\beta \in ker_l } \Psi^{l-1}_{\alpha+\beta,\alpha'+\beta},
\end{align*} 
where  $\Psi^{l-1}_{\alpha,\alpha'} = \dot{q}^l_{\alpha, \alpha'} K^{ l-1}_{\alpha,\alpha'}
    + \hat{q}^{l}_{\alpha, \alpha'}$, and $\hat{q}^{l}_{\alpha,  \alpha}$, resp.  $\dot{q}^l_{\alpha, \alpha'} $ is defined as $q^l$, resp. $\dot{q}^l$ in Lemma \ref{lemma:ffnn_ntk}, with $y_{1,\alpha}^{l-1}(x), y_{1,\alpha'}^{l-1}(x')$ in place of $y_{1}^{l-1}(x),y_{1}^{l-1}(x')$.
\end{lemma}
Lemma \ref{lemma:cnn_ntk} is proved in Appendix \ref{proofsSection3}. 

In the remainder of the paper, we leverage the above recursive formulas for the NTK to analyse its dynamics as $L$ goes to infinity. To alleviate the notations, hereafter, we use the notation $K^L$ for the NTK of both FFNN and CNN. For FFNN,  $K^L$ is given in Lemma \ref{lemma:ffnn_ntk}, whereas for CNN, $K^L =K^L_{\alpha,\alpha'}$ and is given in Lemma \ref{lemma:cnn_ntk} for any $\alpha, \alpha'$, i.e. all results that follow are true for any $\alpha,\alpha'$. We now present a brief review the EOC theory.

\paragraph{Edge of Chaos (EOC).} Recall that the weights and bias are randomly initialized with a Gaussian distribution (see above). Given an input $x$, we denote by $q^l(x)$ the variance of $y^l(x)$. The convergence analysis of $q^l(x)$ as $l$ increases is provided in  \cite{lee_nngp}, \cite{samuel}, and \cite{hayou}. Under general regularity conditions, it is proved that $q^l(x)$ converges to a point $q(\sigma_b, \sigma_w)>0$ independent of $x$ as $l \rightarrow \infty$. The asymptotic behaviour of the correlation $c^l(x,x')$ between $y^l(x)$ and $y^l(x')$ for any two inputs $x$ and $x'$ is also driven by the choice of $(\sigma_b, \sigma_w)$;  \citet{samuel} show that if  $\sigma_w^2 \mathbb{E}[\phi'(\sqrt{q(\sigma_b, \sigma_w)}Z)^2] < 1$, where $Z \sim \mathcal{N}(0, 1)$  then $c^l(x,x')$ converges to 1 exponentially quickly, This is called the ordered phase. On the other hand, if $\sigma_w^2 \mathbb{E}[\phi'(\sqrt{q(\sigma_b, \sigma_w)}Z)^2] > 1$ then $c^l(x,x')$ converges to $c<1$, this is referred to as the chaotic phase.  The authors define the EOC as the set of parameters $(\sigma_b, \sigma_w)$, such that $\sigma_w^2 \mathbb{E}[\phi'(\sqrt{q(\sigma_b, \sigma_w)}Z)^2] =1$. The behaviour of $c^l(x,x')$ in the EOC is studied in \cite{hayou} where it is proved to converge to 1 at a polynomial rate. We provide a comprehensive review of the theory of signal propagation theory at initialization in Appendix \ref{app:warmup_mean_field}. Notice that most results on signal propagation are stated with standard parameterization ($\sigma_w/\sqrt{n_{l-1}}$ and $\sigma_b$ are absorbed in the weights and bias at initialization). The two parameterizations are actually closely related and yield similar results for a properly scaled learning rate. In this work, we focus exclusively on the NTK parameterization introduced above and derive our results in this setting. We will show in this section that a choice of $(\sigma_b, \sigma_W)$ satisfying the EOC condition is beneficial for the NTK. On the contrary, any initialization in the ordered or chaotic phase, leads to exponential convergence rate of $K^L$ to a constant kernel in the limit of infinite depth. 
% the following two classes of activation functions.
% \begin{definition}
% Let $\phi$ be an activation function. Then
% \begin{enumerate}
%     \item $\phi$ is said to be ReLU-like if there exist $\lambda, \beta \in \mathbb{R}$ such that $\phi(x)=\lambda x$ for $x>0$ and $\phi(x)=\beta x$ for $x\leq0$.
%     \item  $\phi$ is said to be of class $\mathcal{S}$ if $\phi(0)=0$, $\phi$ has polynomial growth, and $\phi \in \mathcal{C}^4(\mathbb{R})$.
% \end{enumerate}
% \end{definition}

%\cite{hayou}, \cite{samuel} and \cite{pool} studied an initialization scheme known as the Edge of Chaos and proved that it maximizes the signal propagation through the neural network. However, this has not been directly linked to the training dynamics, we address\\
Let us first introduce some notation for the uniform convergence results. For $\epsilon \in (0,1)$, we define the set $B_\epsilon$ by
\begin{align*}
\text{FFNN : }B_\epsilon &= \{ (x,x') \in \mathbb{R}^d : c^1(x,x') \leq 1-\epsilon\},\\
\text{CNN : }B_\epsilon &= \{ (x,x') \in \mathbb{R}^d : \forall \alpha, \alpha', c^1_{\alpha, \alpha'}(x,x') \leq 1-\epsilon\}.
\end{align*}
We prove uniform convergence results on $B_{\epsilon}$. Given a dataset $\mathcal{D}$ with inputs $\mathcal{X}$, the existence of $\epsilon>0$ such that for all $x\neq x' \in \mathcal{X}, (x,x') \in B_\epsilon$ is guaranteed provided that there is no collinearity between inputs in the dataset $\mathcal{D}$.\\
In the next proposition, we prove that with a choice of $(\sigma_b,\sigma_w)$ in the ordered or chaotic phase, the infinite-width kernel $K^L$ converges to a constant kernel exponentially quickly in the limit of large depth.
\begin{prop}[NTK with Ordered/Chaotic Initialization]\label{prop:ordered_chaotic_ntk}
Let $(\sigma_b, \sigma_w)$ be  either in the ordered or in the chaotic phase. Then, there exist $\lambda>0$ such that for all $\epsilon \in (0,1) $, there exists $\gamma > 0$ such that  
$$
\sup_{(x, x') \in B_\epsilon} |K^L(x,x') - \lambda | \leq e^{-\gamma L}.$$
\end{prop}
The proof of Proposition \ref{prop:ordered_chaotic_ntk} is provided in Section \ref{sec:prop:ordered}. It relies on the asymptotic analysis of the second moment of the gradient. 

Proposition \ref{prop:ordered_chaotic_ntk} shows that off-diagonal terms of the NTK matrix $\hat{K}^L$ converge to a constant $\lambda>0$ as the depth grows when $(\sigma_b, \sigma_w)$ is chosen to be in the ordered or chaotic phases. The exponential convergence rate implies that even with a small number of layers, the kernel $K^L$ is close to being degenerate. This suggests that NTK training fails in this case, and the performance of the NTK regime solution will be no better than that of a random classifier. Empirically, we find that with depth $L=30$, the NTK training fails when the network is initialized in the ordered phase (Section \ref{section:experiments}). However, this can be mitigated by using hyper-parameters $(\sigma_b,\sigma_w)$ in the EOC as we show in the next result. Before stating the results for the EOC initialization, we first introduce the following assumption on the input space of CNN.
\begin{assumption}\label{assumption:cnn}
For all $x,x' \in \mathcal{X}$, $q^1_{\alpha,\alpha'}(x,x')$ is independent of $\alpha,\alpha'$.
\end{assumption}
Under Assumption \ref{assumption:cnn}, there exists some function $e:(x,x')\mapsto e(x,x')$ such that for all $\alpha,\alpha'$ and $x,x' \in \mathcal{X}$
$$
\sum_{j} \sum_{\beta \in ker} x_{j,\alpha+\beta}x'_{j,\alpha'+\beta}= e(x,x')
$$
which combined with Proposition \ref{lemma:cnn_ntk} implies that the NTK in the CNN architecture is the same as that in the FFNN architecture with $e(x,x')$ replacing $x \cdot x'$.
It is a constraint on the set of inputs $\mathcal{X}$. 
% The above  system has $N^2 M^2$ equations and $N\times 2n_0\times M$ variables (the $x_{j,\alpha}$'s) . Therefore, in the case $n_0>>1$, the set of solutions $S$ is large . 
Hereafter, for all CNN analysis, for some function $G$ and set $E$, \emph{taking the supremum $\sup_{(x,x') \in E} G(x,x')$ should be interpreted as $\sup_{(x,x') \in E \cap \mathcal{X}^2} G(x,x')$}. In the following we will specify  clearly whenever we use this assumption.\\

When we choose $(\sigma_b, \sigma_w)$ in the EOC, the next result shows that the NTK explodes in the limit of large depth. However, leveraging our remark on the scale invariance property of the NTK (see Eq. \eqref{eq:scale_invariance1}), we show that a scaled version of the kernel converges at a polynomial rate to the degenerate kernel (this is better than the exponential rate in the Ordered/Chaotic phase). 
% In the next theorem, the notation $g(x) = \Theta(m(x))$ \textcolor{red}{ We need to introduce the notations $\Theta()$ and $\mathcal O( )$ earlier }means there exist two constants $A,B > 0$ such that $ A \, m(x) \leq g(x) \leq B \, m(x)$.
\begin{thm}[NTK in the EOC]\label{thm:ntk_eoc}
Let $(\sigma_b, \sigma_w) \in \text{EOC}$, $\tilde{K}^L = K^L/L$, and $E \subset \mathbb{R}^d$ any compact set. We have that
$$\sup_{x\in E}| \tilde{K}^L(x, x) - \tilde{K}^{\infty}(x,x)| = \Theta(L^{-1}).$$
Moreover, there exists a constant $ \lambda\in (0,1)$ such that for all $\epsilon \in (0,1)$
$$
\sup_{(x,x') \in B_\epsilon} \big| \tilde{K}^L(x, x') -  \tilde{K}^{\infty}(x,x') \big| = \Theta(\log(L) L^{-1}), \quad \textup{where}
$$
\begin{itemize}
    \item If $\phi(x)=\max(0,x)$, i.e. the ReLU activation function, then $\tilde{K}^{\infty}(x,x') = \frac{\sigma_w^2 \|x\|~\|x'\|}{d}( 1 -(1-\lambda) \mathbbm{1}_{x\neq x'} )$ with $\lambda=1/4$.
    \item If $\phi=\textrm{Tanh}(x)$, i.e. the Hyperbolic Tangent activation function, then $\tilde{K}^{\infty}(x,x') = q( 1 -(1-\lambda) \mathbbm{1}_{x\neq x'} )$ where $q>0$ is a constant and $\lambda=1/3$.
\end{itemize}
All results hold for CNN under Assumption \ref{assumption:cnn}.
\end{thm}
Theorem \ref{thm:ntk_eoc} is proved in Section \ref{sec:pr:thEOC}.
The proof requires a special form of inequalities to control the convergence rate (i.e. to obtain $\Theta$ instead of $\bigO$). 
% Since $0<\lambda <1 $, on the EOC there exists an invertible matrix $J$ such that $\hat K^L = L \times J(1 + o(1))$ as $L \rightarrow \infty$, so that $\hat K^L$ remains asymptotically invertible and thus makes the NTK training possible. Hence initializing on the EOC has a double effect : it leaves the $\hat K^L$ invertible and it 
Theorem \ref{thm:ntk_eoc} shows that the EOC initialization yields a polynomial convergence rate (w.r.t $L$) of $\tilde{K}^L$ to the trivial kernel $\tilde{K}^\infty$. This is important knowing that  $\tilde{K}^\infty$ is trivial and brings hardly any information on $x$\footnote{Notice that the RKHS of $\tilde{K}^\infty$ is restricted to constant functions for Tanh and functions of the form $x \to \alpha \|x\|, \alpha \in \mathbb{R}$, for ReLU.}. Indeed, the convergence rate of $\tilde{K}^L$  to $\tilde{K}^\infty$ is $\Theta(\log(L) L^{-1})$. This means that as $L$ grows, with a choice of $(\sigma_b, \sigma_w)$ in the EOC, the kernel $\tilde{K}$ is still much farther from the trivial kernel $\tilde{K}^\infty$ compared to the Ordered/Chaotic initialization. The EOC allows thus the use of NTK training on deeper networks. Empirical results in \Cref{section:experiments} confirm this theoretical findings. In the next section, we show how the network architecture can impact the behaviour of the infinite-width NTK in the large depth limit.
\subsection{Residual Neural Networks (ResNet)}
Another important feature of DNNs, which is known to be highly influential, is the network architecture. For residual networks (commonly known as ResNet), the NTK also follows a simple recursion expression in the infinite-width limit. 
Consider a ResNet of the form
\begin{equation}\label{resNN:def}
y^l(x) = y^{l-1}(x) + \mathcal{F}(w^l, y^{l-1}(x)), \quad l\geq 2,
\end{equation}
where $\mathcal{F}$ is either a  dense layer as in  \eqref{equation:ffnn_net} or a convolutional layer  as in  \eqref{equation:convolutional_net}). The NTK recursion is given in the following lemma. 
\begin{lemma}[NTK of a ResNet with fully connected layers in the infinite-width limit]\label{lemma:resnet_ntk}
Let $K^{res,1}$ be the exact NTK for the ResNet with 1 layer. Then\\
$\bullet$~For the first layer (without residual connections), we have for all $x,x' \in \mathbb{R}^d$
$$
K^{res,1}_{ii'}(x,x') = \delta_{ii'}\left( \sigma_b^2 + \frac{\sigma_w^2}{d} x\cdot x'\right).
$$

$\bullet$~For $l\geq2$, as $n_1, n_2, ..., n_{l-1} \rightarrow \infty$ recursively, we have for all $i,i' \in [n_l]$, $K^{res,l}_{ii'} = \delta_{ii'} K^l_{res}$, where $K^l_{res}$ is given by the recursive formula for all $x,x' \in \mathbb R^d$
$$
K_{res}^l(x, x') = K_{res}^{l-1}(x, x') (\dot{q}^l(x, x')+1) + q^l(x,x').
$$
\end{lemma}

Adding skip connections has a direct impact on the recursion of the infinite-width NTK; it explicitly adds a term $K_{res}^{l-1}$ to the recursion. A similar result holds for residual networks with convolutional layers. 
\begin{lemma}[NTK of a ResNet with convolutional layers in the infinite-width limit]\label{lemma:resnet_ntk_conv}
Let $K^{res,1}$ be the exact NTK for the ResNet with 1 layer. Then\\
$\bullet$~For the first layer (without residual connections), we have for all $x,x' \in \mathbb{R}^d$
$$
K^{res,1}_{(i,\alpha), (i',\alpha')}(x,x') = \delta_{ii'} \Big(\frac{\sigma_w^2}{n_0(2k+1)}[x,x']_{\alpha,\alpha'} + \sigma_b^2 \Big).
$$
$\bullet$~For $l\geq2$, as $n_1, n_2, ..., n_{l-1} \rightarrow \infty$ recursively, we have for all $i,i' \in [n_l]$, $\alpha,\alpha' \in [0:M-1]$, $K^{res,l}_{(i,\alpha),(i',\alpha')}(x,x') = \delta_{ii'} K^{res, l}_{\alpha,\alpha'}(x,x')$, where $K^{res, l}_{\alpha,\alpha'}$ is given by the recursive formula for all $x,x' \in \mathbb{R}^d$, using the same notations as in lemma \ref{lemma:cnn_ntk},
$$
K^{res, l}_{\alpha,\alpha'} = K^{res, l-1}_{\alpha,\alpha'} + \frac{1}{2k+1} \sum_{\beta} \Psi^{l-1}_{\alpha+\beta, \alpha'+\beta},
$$
where $\Psi^l_{\alpha,\alpha'} = \dot{q}^l_{\alpha, \alpha'} K^{res, l}_{\alpha,\alpha'}+ \hat{q}^{l}_{\alpha, \alpha'}$.
%where $\Sigma^{l}_{\alpha, \alpha'} = \sigma_b^2 + \sigma_w^2 \mathbb{E}[\phi(y^{l-1}_{1,\alpha}(x))\phi(y^{l-1}_{1,\alpha'}(x'))]$ 
%and $\dot{\Sigma}^l_{\alpha, \alpha'} = \sigma_w^2 \mathbb{E}[\phi'(y^{l-1}_{1,\alpha}(x))\phi'(y^{l-1}_{1,\alpha'}(x'))]$.

\end{lemma}

The additional terms $ K_{res}^{l-1}(x, x')$ (resp.   $K^{res, l-1}_{\alpha,\alpha'}$) in the recursive formulas of Lemma \ref{lemma:resnet_ntk} (resp. Lemma \ref{lemma:resnet_ntk_conv}) appear as a result of the skip connections in the ResNet architecture. It turns out that this term helps in slowing down the convergence rate of the NTK. The next proposition shows that for any $\sigma_w>0$\footnote{We omit the bias in the ResNet architecture for the sake of simplification.}, the infinite-width NTK of a ResNet explodes (exponentially) as $L$ grows. However, a normalized version $\Bar{K}^L = K^L / \alpha_L$ of the NTK of a ResNet will always have a polynomial convergence rate to a limiting trivial kernel. Recall that the NTK regime is scale-invariant (Eq. \eqref{eq:scale_invariance1}), i.e. the scaling factor $\alpha_L$ does not change this regime.
\begin{thm}[NTK for ResNet]
\label{thm:ntk_residual}
Consider a ResNet satisfying 
\begin{equation}
y^l(x) = y^{l-1}(x) + \mathcal{F}(w^l, y^{l-1}(x)), \quad l\geq 2,
\end{equation}
where $\mathcal{F}$ is either a convolutional or dense layer (Eqs. \eqref{equation:ffnn_net} and \eqref{equation:convolutional_net}) with ReLU activation. Let $K^L_{res}$ be the corresponding NTK, and $\Bar{K}^L_{res} = K^L_{res} / \alpha_L$ (Normalized NTK) with $\alpha_L = L (1 + \frac{\sigma_w^2}{2})^{L-1}$. Then, for all compact sets $E \subset \mathbb{R}^d$, we have 
$$\sup_{x\in E} |\Bar{K}^L_{res}(x,x) - \Bar{K}^{\infty}_{res}(x,x)| = \Theta(L^{-1}).$$
Moreover, there exists a constant $\lambda\in (0,1)$ such that for all $\epsilon \in (0,1)$
$$\sup_{x,x' \in B_\epsilon} \big|\Bar{K}^L_{res}(x,x') - \Bar{K}^\infty_{res}(x,x') \big| = \Theta(\log(L) L^{-1}),$$
where $\Bar{K}^{\infty}_{res}(x,x') =\frac{\sigma_w^2 \|x\|~ \|x'\|}{d}( 1 -(1-\lambda) \mathbbm{1}_{x\neq x'} )$.\\
All results hold for ResNet with Convolutional layers under Assumption \ref{assumption:cnn}.
\end{thm}
The proof techniques of Theorem \ref{thm:ntk_residual} are similar to that of Theorem \ref{thm:ntk_eoc}. Details are provided in Section \ref{proofsSection}.\\
Theorem \ref{thm:ntk_residual} shows that the NTK of a ReLU ResNet explodes exponentially w.r.t  $L$. However, the normalized kernel $\Bar{K}^L_{res} = K^L_{res}(x,x')/ \alpha_L$ converges to a limiting kernel $\Bar{K}^\infty_{res}$ at the exact polynomial rate $\Theta(\log(L)L^{-1})$ for all $\sigma_w>0$. This allows for NTK training of deep ResNet, similarly to the EOC initialization for the FFNN or the CNN networks. However, $\Bar{K}^L_{res}$ converges to a trivial kernel. Thus, NTK training will fail at some point as we increase the depth. This is empirically verified in Section \ref{section:experiments}. 
% Recent work by \cite{hayou2020stable} introduced a scaling trick to achieve a non-degenerate kernel in the limit of infinite depth.
% \begin{figure*}
%   \centering
%   \subfigure[EOC]{%
%   \label{fig:convergence_rate_eoc}
%   \includegraphics[width=0.28\linewidth]{eoc_conv_rate.png}
%   }%
%   \subfigure[Ordered phase]{%
%   \label{fig:convergence_rate_ord}
%   \includegraphics[width=0.28\linewidth]{ordered_conv_rate.png}
%   }%
%   \subfigure[ResNet with FFNN blocks]{%
%   \label{fig:convergence_rate_res}
%   \includegraphics[width=0.28\linewidth]{resnet_conv_rate.png}
%   }%
%  \caption{Convergence rates for different initializations and architectures. We plot the average convergence rate for 10 randomly selected datapoints from CIFAR10. Curves were shifted by constants for clarity purpose. (a) Edge of Chaos. (b) Ordered phase. (c) Adding residual connections.}
%  \label{fig:convergence_rate}
% \end{figure*}

\subsection{Spectral decomposition of the limiting NTK}\label{section:sperctral_decomposition_ntk}
To refine the analysis presented in Section \ref{section:large_depth_ntk_behaviour}, we study the limiting behaviour of the spectrum of the kernels studied in Theorems \ref{thm:ntk_eoc}, \ref{thm:ntk_residual} and Proposition \ref{prop:ordered_chaotic_ntk}, on the unit sphere $\Sd = \{ x \in \mathbb{R}^d : \|x\|_2 = 1\}$. On the sphere $\Sd$, all of these kernels (namely $K^L$ for FFNN in the Ordered/Chaotic phase, $\tilde{K}^L$ for FFNN\footnote{The NTK for CNN is generally not a dot-product kernel since it depends on the quantities $[x,x']_{\alpha, \alpha'}$. We restrict our analysis to FFNN and ResNet with fully-connected layers.} in the EOC, and $\bar{K}_{res}^L$ for ResNets with fully-connected layers) are dot-product kernels, i.e. for any of these kernels, denoted by $\kappa_L$, there exists a function $g_L$ such that $\kappa^L(x,x') = g_L(x \cdot x')$ for all $x,x' \in \Sd$. This type of kernels is known to be diagonalizable on the sphere $\Sd$ and its eigenfunctions are the so-called Spherical Harmonics of $\Sd$. This diagonalization result has been independently observed in other works \citep{geifman2020similarity, cao2020understanding, bietti2021deep}. In the next proposition, we leverage the results of Section \ref{section:large_depth_ntk_behaviour} to study the aforementioned kernels from a spectral perspective.  
\begin{prop}[Spectral decomposition on $\Sd$]\label{prop:spectral_decomposition_Sd}
Let $\kappa^L$ be either, $K^L$ for an FFNN with $L$ layers initialized in the Ordered phase (Proposition \ref{prop:ordered_chaotic_ntk}), $\tilde{K}^L$ for an FFNN with $L$ layers initialized in the EOC (Theorem \ref{thm:ntk_eoc}), or $\Bar{K}^L_{res}$ for a ResNet with $L$ Fully Connected layers (Theorem \ref{thm:ntk_residual}). Then, for all $L\geq1$, there exists $(\mu^L_k)_{k\geq}$ such that for all $x,x' \in \mathbb{S}^{d-1}$
$$
\kappa^L(x,x') = \sum_{k\geq 0 } \mu^L_k \sum_{j=1}^{N(d,k)} Y_{k,j}(x) Y_{k,j}(x').
$$
$(Y_{k,j})_{k\geq0, j\in [1:N(d,k)]}$ are spherical harmonics of  $\mathbb{S}^{d-1}$, and $N(d,k)$ is the number of harmonics of order $k$.

Moreover, we have that $0<\mu^\infty_0 = \lim\limits_{L \rightarrow \infty} \mu^L_0 < \infty$, and for all $k\geq 1,$ 
$
\lim\limits_{L \rightarrow \infty} \mu^L_k = 0
$.
\end{prop}
The proof of Proposition \ref{prop:spectral_decomposition_Sd} is based on a result from spectral theory analysis. The limiting eigenvalues are obtained by a simple application of the dominated convergence theorem.\\
Proposition \ref{prop:spectral_decomposition_Sd} shows that in the limit of large depth $L$, the kernel $\kappa^L$ becomes close to the rank-one kernel $\kappa^\infty$ $(x,x') \mapsto \mu^\infty_0 Y_{0}(x)Y_{0}(x')$, where $Y_{0}$ is a function (constant function for Tanh and $x\to \|x\|$ for ReLU). Therefore, in the limit of infinite depth, the RKHS of the kernel $\kappa^L$ is reduced to the space spanned by $Y_0$, confirming that the NTK regime solution is trivial in this limit (recall that $f_\infty - f_0$ is in the RKHS of $\kappa_L$), hence, NTK training fails. 

\subsection{Scaled ResNet}
A natural extension of the ResNet architecture is the introduction of scaling factors for the blocks of the residual branches. In this section, we study the impact of a particular scaling scheme and show that this simple procedure can significantly reduce the convergence rate to the infinite depth trivial NTK regime.
\begin{thm}[Scaled ResNet]\label{thm:scaled_resnet}
Consider a ResNet satisfying
\begin{equation}
y^l(x) = y^{l-1}(x) +\frac{1}{\sqrt{l}}\mathcal{F}(w^l, y^{l-1}(x)), \quad l\geq 2,
\end{equation}
where $\mathcal{F}$ is either a convolutional or dense layer (eq. \eqref{equation:ffnn_net} and eq. \eqref{equation:convolutional_net}) with ReLU activation. Then the results of Theorem \ref{thm:ntk_residual} apply with   $\alpha_L = L^{1 + \sigma_w^2 / 2}$ and the convergence rate becomes $\Theta(\log(L)^{-1})$.
\end{thm}
Theorem \ref{thm:scaled_resnet} shows that scaling the residual blocks by $1/\sqrt{l}$ has two important effects on the kernel $K^L$: first, it stabilizes the NTK which only grows as $L^{1+\frac{\sigma_w^2}{2}}$ instead of $L (1+\frac{\sigma_w^2}{2})^{L-1}$; second, it drastically slows down the convergence rate to the limiting (trivial) $\Bar{K}^{\infty}_{res}$. Both properties are highly desirable for NTK training. The second property in particular means that with this scaling, the RKHS of $K^L / \alpha_L$ is `richer' compared to the non-scaled case. We show in Section \ref{section:experiments} a case where NTK training with scaled ResNet succeeds to achieve non-trivial performance while the non-scaled NTK regime fails.
\begin{figure}
  \centering
  \includegraphics[width=.65\linewidth]{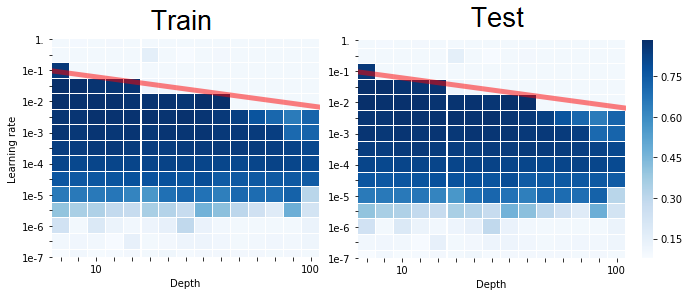}
\caption{Train/Test accuracy of an FFNN with ReLU activation on Fashion MNIST dataset for different depths and learning rates, trained with SGD for 10 epochs. The plot is in log-log scale.}
 \label{fig:heatmaps}
\end{figure}
A similar scaling was studied in \cite{huang_ntk}, where authors scale the blocks with $1/L$ instead of our scaling $1/\sqrt{l}$, and show that it also stabilizes the NTK of ResNet. It is also possible to force the NTK regime to remain `expressive', i.e. non-trivial, in the infinite depth regime by using more aggressive scaling factors (\cite{hayou2020stable}).

\section{Learning Rate Passband}\label{section:learning_rate}
Tuning the learning rate (LR) is crucial for the training of DNNs with SGD; a large/small LR could cause the training to fail. Empirically, the optimal LR tends to decrease as the network depth grows. In this section, we use the NTK linear model presented in Section \ref{section:ntk} to establish the existence of an LR \emph{passband}, i.e. an interval of values for the LR where training occurs. The idea is to use the infinite-width NTK to approximate the largest eigenvalue of the exact NTK (finite-width NTK) which governs gradient updates. Recall the dynamics of the linear model in the infinite-width limit
\begin{equation}\label{equation:ode_f_linear}
df_t(\mathcal{X}) = - \frac{1}{N} \hat{K}^L (f_t(\mathcal{X}) -  \mathcal{Z}) dt .
\end{equation}
The GD update with learning rate $\eta$ is given by 
\begin{equation}\label{equation:update_rule}
f_{t+1}(\mathcal{X}) = (I - \frac{\eta}{N} \hat{K}^L)f_{t}(\mathcal{X}) - \frac{\eta}{N} \hat{K}^L \mathcal{Z} .
\end{equation}
% Similarly, the GD `jump' is given by 
% \begin{equation}\label{equation:jump}
% f_{t+1}(\mathcal{X}) - f_{t}(\mathcal{X}) = - \frac{\eta}{N} \hat{K}^L (f_{t}(\mathcal{X}) - \mathcal{Z}).
% \end{equation}
To ensure stability in Eq. \eqref{equation:update_rule}, a necessary condition is that $\| I - \frac{\eta}{N} \hat{K}^L\|_F < 1$, which implies having
$$
\eta < \frac{2}{ \mu_{\max}(\frac{1}{N} \hat{K}^L) }
$$
where $\mu_{\max}$ refers to the largest eigenvalue.\\
% To obtain a lower bound on the learning rate, we use the GD jump equation \eqref{equation:jump}. Indeed, forcing the GD jumps to be of order $\Theta(1)$, we obtain the lower bound 
% $$
% \eta \geq \Theta \left( \frac{1}{\sqrt{N} \mu_{\max}(\frac{1}{N} \hat{K}^L)}\right)
% $$
Consider the case of an FFNN with Tanh activation for instance, initialized in the EOC. Then, as $L$ grows we have that $\hat{K}^L = q L ((1-\lambda)I + \lambda U) + \bigO(\log(L))$ (Theorem \ref{thm:ntk_eoc}). Therefore, for large $L$ and $N$, we have that $\mu_{\max}(\frac{1}{N} \hat{K}^L) \sim q \lambda L$. \\
The upper bound on $\eta$ scales as $1/L$, therefore, we expect the passband to have a linear upper bound when we plot $\log(LR)$ vs $\log(L)$. To validate this hypothesis, we train an FFNN on Fashion-MNIST dataset. 
Figure \ref{fig:heatmaps} shows the train/test accuracy for a grid of LRs and depths. The slope of the red line is $-1$ which confirms our prediction that the upper bound of the LR passband grows as $L^{-1}$. A similar bound has been introduced recently in \cite{hayase2020spectrum} in the different context of networks achieving dynamical isometry with Hard-Tan activation function. Figure \ref{fig:heatmaps} also shows that the lower bound of the passband and depths are almost uncorrelated.\footnote{We currently do not have an explanation for this effect. }

\section{Experiments}\label{section:experiments}
% \begin{figure}
%   \centering
%   \includegraphics[width=1.\linewidth]{ntk_decomposition.jpg}
%  \caption{Normalized eigenvalues of $K_L$ on the 2D sphere for an FFNN with different initializations, activations, and depths.}
%  \label{fig:ntk_decomposition}
% \end{figure}
\begin{figure}
  \centering
  \includegraphics[width=.6\linewidth]{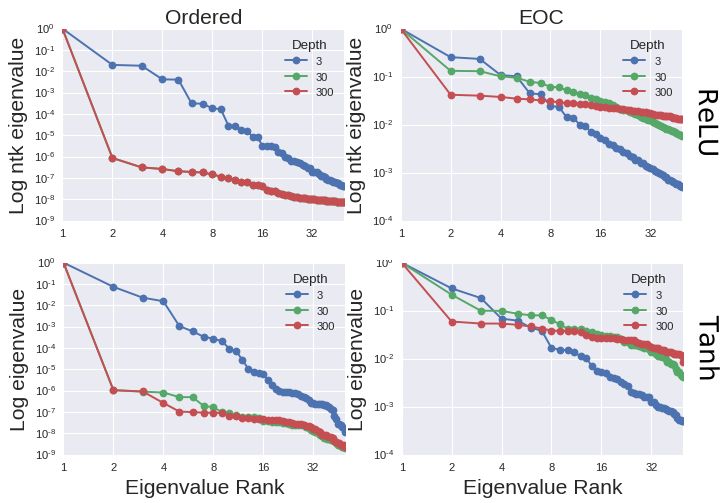}
 \caption{\small{Normalized eigenvalues of $K_L$ on the 2D sphere for an FFNN with different initializations, activations, and depths.}}
 \label{fig:ntk_decomposition}
\end{figure}
\subsection{Behaviour of $K^L$ as $L$ goes to infinity}
Proposition \ref{prop:ordered_chaotic_ntk}, and Theorems \ref{thm:ntk_eoc} and \ref{thm:ntk_residual} show that the NTK (or scaled NTK) converges to a trivial kernel.
% give theoretical convergence rates for quantities of the form $\big|\kappa^L - \kappa^{\infty}|$. 
% We illustrate these results in Figure \ref{fig:convergence_rate}. Figure \ref{fig:convergence_rate_eoc} shows a convergence rate of approximately order $L^{-1}$ for ReLU and Tanh. Recall that the exact rate is $\mathcal{O}(\log(L) L^{-1})$, the logarithmic factor is the reason why the curves are not perfectly linear. With NTK training, Tanh performs better than ReLU for depth $L=30$ (see Table \ref{table:test_accuracies})\footnote{We do not have an explanation for this at the moment, but we believe that the smoothness of Tanh might play a role, see e.g. \cite{hayou}}. 
% Figure \ref{fig:convergence_rate_ord} demonstrates that this convergence occurs at an exponential convergence rate in the Ordered phase for both ReLU and Tanh, and Figure \ref{fig:convergence_rate_res} shows the convergence rate in the case of FFNN with residual connections. As predicted by Theorem \ref{thm:ntk_residual}, the convergence rate $\mathcal{O}(L^{-1} \log(L))$ is independent of the parameter $\sigma_w$ in that case.
Figure \ref{fig:ntk_decomposition} shows the normalized eigenvalues of the NTK of an FFNN on 2D sphere. In the Ordered phase, the eigenvalues converge quickly to zero as the depth grows, while with an EOC initialization, the eigenvalues converge to zero at a slower rate. For $L=300$, the NTK in the EOC is `richer' than the NTK in the Ordered phase, in the sense that the small eigenvalues with EOC are relatively much bigger than those with the Ordered phase intialization. This reflects directly on the RKHS of the NTK, and allows the NTK regime solution to be more expressive since it is a linear combination of the eigenfunctions of the NTK.

\subsection{Can NTK regime explain DNN performance? }
We train FFNN, Vanilla CNN (stacked convolutional layers without pooling, followed by a dense layer), Vanilla ResNet (ResNet with FFNN blocks) with different depths using two training methods:

\begin{table*}
  \caption{Test accuracy for varying architectures and  depths on MNIST and CIFAR10 dataset. Test accuracy is reported after 100 training epochs for $L \in \{3,30\}$ and $160$ epochs for $L=300$.}
  \label{table:test_accuracies}
  \vskip 0.05in
  \centering
  \resizebox{14cm}{!}{
  \begin{tabular}{lllllllllll}
    \multicolumn{1}{c}{} & \multicolumn{4}{c}{MNIST} & \multicolumn{4}{c}{CIFAR10}\\
    \cmidrule(r){2-5}
    \cmidrule(r){6-9}
    \multicolumn{1}{c}{} & \multicolumn{2}{c}{NTK Training} & \multicolumn{2}{c}{SGD Training} & \multicolumn{2}{c}{NTK Training} & \multicolumn{2}{c}{SGD Training}\\
    \cmidrule(r){2-3}
    \cmidrule(r){4-5}
    \cmidrule(r){6-7}
    \cmidrule(r){8-9}
             & EOC         &  Ordered   & EOC       &  Ordered  & EOC         &  Ordered   & EOC       &  Ordered  \\
    \midrule
    \textbf{L=3}\\
    FFNN-ReLU           & $96.64_{\pm 0.11}$  &  $ 96.57_{\pm 0.12}$   & $97.05_{\pm 0.27}$   &  $97.11_{\pm 0.31}$  & $48.13_{\pm 0.10}$  &  $ 48.45_{\pm 0.14}$   & $55.13_{\pm 0.23}$   &  $54.10_{\pm 0.12}$   \\
    FFNN-Tanh                & $95.34_{\pm 1.04}$  &  $96.32_{\pm 0.41}$   & $97.19_{\pm 0.11}$  &  $97.03_{\pm 0.29}$ & $48.32_{\pm 0.15}$  &  $ 48.10_{\pm 0.10}$   & $56.13_{\pm 0.34}$   &  $54.10_{\pm 0.23}$ \\
    CNN-ReLU                & $97.13_{\pm 0.31}$  &  $97.23_{\pm 0.22}$   & $98.95_{\pm 0.12}$  &  $98.89_{\pm 0.18}$ & $49.11_{\pm 0.16}$  &  $ 42.76_{\pm 3.32}$   & $60.23_{\pm 0.45}$   &  $59.05_{\pm 0.15}$\\
    V-ResNet    & $96.73_{\pm 0.05}$  &  $96.71_{\pm 0.16}$   & $97.19_{\pm 0.23}$  &  $97.12_{\pm 0.14}$ & $47.82_{\pm 0.73}$  &  $48.01_{\pm 0.20}$   & $54.40_{\pm 0.24}$  &  $54.28_{\pm 0.33}$ \\
    \midrule
    \textbf{L=30}\\
    FFNN-ReLU           & $96.95_{\pm 0.22}$  &  \textemdash   & $97.55_{\pm 0.09}$   &  \textemdash   & $48.32_{\pm 0.10}$  &  \textemdash   & $56.10_{\pm 0.41}$   &  \textemdash   \\
    FFNN-Tanh                & $97.30_{\pm 0.15}$  &  \textemdash   & $97.87_{\pm 0.17}$  &  \textemdash & $48.40_{\pm 0.12}$  &  \textemdash   & $57.39_{\pm 0.08}$   &  \textemdash  \\
    CNN-ReLU                & $98.60_{\pm 0.13}$  &  \textemdash   & $99.02_{\pm 0.07}$  &  \textemdash & $48.42_{\pm 0.10}$  &  \textemdash   & $75.39_{\pm 0.31}$   &  \textemdash \\
    V-ResNet                & \textemdash  &  \textemdash   & $98.17_{\pm 0.03}$  &  $98.13_{\pm 0.08}$ & \textemdash  &  \textemdash   & $57.09_{\pm 0.47}$  &  $58.13_{\pm 0.18}$ \\
    \midrule
    \textbf{L=300}\\
    FFNN-ReLU          & \textemdash &  \textemdash  & $98.14_{\pm 0.12}$   &  \textemdash & \textemdash &  \textemdash  & $30.25_{\pm 3.23}$   &  \textemdash   \\
    FFNN-Tanh               & \textemdash  &  \textemdash  & $98.54_{\pm 0.18}$  &  \textemdash  &  \textemdash &  \textemdash  & $58.25_{\pm 0.43}$   &  \textemdash\\
    CNN-ReLU                & \textemdash  &  \textemdash   & $99.43_{\pm 0.04}$  &  \textemdash & \textemdash &  \textemdash  & $76.25_{\pm 0.21}$   &  \textemdash\\
    V-ResNet                & \textemdash  &  \textemdash   & $98.23_{\pm 0.09}$  &  $98.19_{\pm 0.06}$  & \textemdash  &  \textemdash   & $58.87_{\pm 0.44}$  &  $59.25_{\pm 0.10}$\\
    S-ResNet                & $97.10_{\pm 0.13}$  &  $97.15_{\pm 0.10}$    & $98.57_{\pm 0.06}$  &  $98.76_{\pm 0.8}$  & $46.77_{\pm 0.19}$   &  $47.13_{\pm 0.10}$    & $59.14_{\pm 0.37}$  &  $59.91_{\pm 0.51}$\\
    \bottomrule
  \end{tabular}
  }
\end{table*}
\paragraph{SGD training.} We use SGD with a batchsize of 128 and a learning rate $10^{-1}$ for $L\in \{3,30\}$ and $10^{-2}$ for $L=300$ (this learning rate was found by a grid search of exponential step size 10; note that the optimal learning rate with NTK parameterization is usually bigger than the optimal learning rate with standard parameterization). We use $100$ training epochs for $L \in \{3,30\}$, and $150$ epochs for $L=300$, with 1 Tesla v100 GPU for all our experiments, which ran for (approximately) 2 weeks.\\
\paragraph{NTK training.} We use the Python library Neural-Tangents introduced by \citet{neuraltangents2020} with $10K$ samples from MNIST/CIFAR10\footnote{CIFAR10: \url{https://www.cs.toronto.edu/~kriz/cifar.html}\\MNIST: \url{http://yann.lecun.com/exdb/mnist/}}. This corresponds to the inversion of a $10K \times 10K$ matrix to obtain the NTK regime solution discussed in Section \ref{section:ntk}.

For the EOC initialization, we use $(\sigma_b,\sigma_w)=(0,\sqrt{2})$ for ReLU, and $(\sigma_b,\sigma_w)=(0.2,1.298)$ for Tanh. For the Ordered phase initialization, we use $(\sigma_b,\sigma_w)=(1,0.1)$ for both ReLU and Tanh (We use the algorithm presented in \cite{hayou} to compute EOC values). Table \ref{table:test_accuracies} displays the test accuracies for both NTK training and SGD training. The dashed lines refer to the trivial test accuracy $\sim 10\%$, which is the test accuracy of a uniform random classifier with 10 classes i.e. in these cases the model does not learn. For $L=300$, NTK training fails for all architectures and initializations confirming the results of Theorems \ref{thm:ntk_eoc} and \ref{thm:ntk_residual}, and Proposition \ref{prop:ordered_chaotic_ntk}; while SGD succeeds in training FFNN and CNN with an EOC initialization and fails with an Ordered initialization, and succeeds in training ResNet with both initialization schemes (which confirms findings in \cite{yang2017meanfield} that ResNet `live' in the EOC). This shows that the NTK regime cannot explain DNN performance trained with SGD.  We also run experiments with a scaled ResNet (S-ResNet) architecture and report the results for depth $L=300$. S-ResNet is NTK-trainable even with depth $L=300$, which is not the case with standard ResNet. Further results on the performance of scaled ResNet with SGD are provided in the next section.

\subsection{Does Scaled ResNet outperform ResNet with SGD?}
Theorem \ref{thm:scaled_resnet} shows that scaled ResNet is better than standard ResNet in preserving the expressivity of NTK as the depth grows. However, it is not clear whether this scaling has an impact on the performance of the ResNet trained with SGD. This section provides an empirical invistigation of this question.\\
\begin{table}[htp]
\caption{\small{Test accuracy on CIFAR100 for ResNet.}}
\label{table:accuracies_resnet}
\vskip 0.05in
\begin{center}
\begin{small}
\resizebox{6.5cm}{!}{
\begin{tabular}{lcccr}
 &  & Epoch 10 & Epoch 160 \\
\midrule
\multirow{2}{*}{ResNet32} &  standard &  \textbf{54.18$\pm$1.21} & 72.49$\pm$0.18 \\
&scaled& 53.89$\pm$2.32 & \textbf{74.07$\pm$0.22}\\
\midrule
\multirow{2}{*}{ResNet50} &  standard  & 51.09$\pm$1.73 & 73.63$\pm$1.51 \\
&scaled& \textbf{55.39$\pm$1.52} & \textbf{75.02$\pm$0.44}\\
\midrule
\multirow{2}{*}{ResNet104} &  standard  & 47.02$\pm$3.23 & 74.77$\pm$0.29\\
&scaled& \textbf{56.38$\pm$2.54} & \textbf{76.14$\pm$0.98}\\
\bottomrule
\end{tabular}
}
\end{small}
\end{center}
\vskip -0.1in
\end{table} 

We train standard and scaled ResNets with depths 32, 50, and 104 on CIFAR100 with SGD. We use a decaying learning rate schedule; we start with 0.1 and divide by 10 after $n_e/2$ epochs, where $n_e$ is the total number of epochs; we scale again, by 10, after $n_e/4$ epochs. We use a batch size of 128, and we train the model for 160 epochs. Table \ref{table:accuracies_resnet} displays test accuracy for standard ResNet and scaled ResNet after 10 and 160 epochs of SGD training; Scaled ResNet outperforms ResNet and converges faster. However, it is not clear whether this is linked to the NTK, or caused by something else. The impact of scaling ResNet blocks is well understood in the NTK regime but remains an open question in the feature learning regime (the regime where the NTK changes with training time) which occurs with SGD training.

% With $L=30$, NTK training fails with Vanilla ResNet, while it yields good performance with scaled ResNet; this also confirms the benefits of the scaling introduced in Proposition \ref{prop:scaled_resnet}. However, even with scaled ResNet, the NTK training fails for depth $L=300$.

% \begin{table}
% \caption{Test accuracy on CIFAR100 for ResNet.}
% \label{table:accuracies_resnet}
% \vskip 0.05in
% \begin{center}
% \begin{small}
% \resizebox{7cm}{!}{
% \begin{tabular}{lcccr}
%  &  & Epoch 10 & Epoch 160 \\
% \midrule
% \multirow{2}{*}{ResNet32} &  standard &  \textbf{54.18$\pm$1.21} & 72.49$\pm$0.18 \\
% &scaled& 53.89$\pm$2.32 & \textbf{74.07$\pm$0.22}\\
% \midrule
% \multirow{2}{*}{ResNet50} &  standard  & 51.09$\pm$1.73 & 73.63$\pm$1.51 \\
% &scaled& \textbf{55.39$\pm$1.52} & \textbf{75.02$\pm$0.44}\\
% \midrule
% \multirow{2}{*}{ResNet104} &  standard  & 47.02$\pm$3.23 & 74.77$\pm$0.29\\
% &scaled& \textbf{56.38$\pm$2.54} & \textbf{76.14$\pm$0.98}\\
% \bottomrule
% \end{tabular}
% }
% \end{small}
% \end{center}
% \vskip -0.1in
% \end{table}

\section{Conclusion and Limitations}\label{sec:limitations}
In this paper, we have shown that the infinite depth limit of the NTK regime is trivial and cannot explain the performance of DNNs. However, the convergence occurs with hyper-parameters dependent rate, which is has a direct impact on the empirical performance (Table \ref{table:test_accuracies}).
These findings add to a recent line of research which shows that the infinite-width approximation of the NTK does not fully capture the training dynamics of DNNs. Indeed, recent works have shown that the NTK for finite width neural networks changes with time \citep{chizat,ghorbani2019linearized,huang}. \cite{hanin} showed that it can even be random in the limit $n, L \rightarrow \infty$ (where $n$ is a width of the network) with fixed ratio $\gamma = \frac{L}{n}$. An interesting property in this regime is the ``feature learning'' which the NTK regime lacks.  Further research is needed in order to understand the difference between the two regimes.

\section{Proof techniques}\label{section:proof_techniques}
The techniques used in the proofs range from simple algebraic manipulation to more specialized inequalities. A summary of the key ideas is given below.

\paragraph{Lemmas \ref{lemma:ffnn_ntk}, \ref{lemma:cnn_ntk}, \ref{lemma:resnet_ntk}, \ref{lemma:resnet_cnn_ntk}.}The proofs of these lemmas are simple and follow the same inductive argument as in the proof of the original NTK result in \cite{jacot}. Note that these results can also be obtained using results from \cite{yang_tensor3_2020}.

\paragraph{Proposition \ref{prop:ordered_chaotic_ntk}, Theorems \ref{thm:ntk_eoc}, \ref{thm:ntk_residual}, \ref{thm:scaled_resnet}.} The proof of these results follow two steps; firstly, we derive the asymptotic behaviour of the NTK in the limit of large depth $L$; secondly, we control this asymptotics using uniform upper/lower bounds. We analyse the asymptotic behaviour of the NTK of FFNN using existing results on signal propagation in deep FFNN. For CNNs, the dynamics are more complicated as they involve convolution operators. We use some results from the theory of Circulant Matrices for this purpose. The use of  Assumption \ref{assumption:cnn} also simplifies the analysis for CNNs with $(\sigma_b,\sigma_w)$ chosen in the EOC.\\
It is relatively easy to control the dynamics of the NTK in the Ordered/Chaotic phase. However, these dynamics become much more sophisticated when the network is initialized in the EOC; technical lemmas which we call Appendix Lemmas are introduced for this purpose. A key ingredient in the proofs of Theorems \ref{thm:ntk_eoc}, \ref{thm:ntk_residual}, and \ref{thm:scaled_resnet}, is the following lemma, which we state and prove in the appendix.\\

\noindent\textbf{Lemma. }[Uniform bounds]
\textit{ Let $A \subset \mathbb{R}$ be a compact set and $g$ a non-decreasing function on $A$. Define the sequence $\zeta_l$ by $\zeta_l = g(\zeta_{l-1})$ and $\zeta_0 \in A$. Assume that there exist $\alpha_l, \beta_l$ that do not depend on $\zeta_0$ (in the sense that sequences $\alpha_l, \beta_l$  are the same for all $\zeta_0 \in A$), with $\beta_l = o(\alpha_l)$, such that for all $\zeta_0 \in A$, 
$$
\zeta_l = \alpha_l + \bigO_{\zeta_0}(\beta_l)
$$
where $\bigO_{\zeta_0}$ means that the $\bigO$ bound depends on $\zeta_0$. Then, we have that 
$$
\sup_{\zeta_0 \in A} |\zeta_l - \alpha_l| = \bigO(\beta_l),
$$
i.e. we can choose the bound $\bigO$ to be independent of $\zeta_0$.
}

\paragraph{Proposition \ref{prop:spectral_decomposition_Sd}.} The spectral decomposition of zonal kernels on the sphere is a classical result in spectral theory which was recently applied to NTK  \citep{geifman2020similarity, cao2020understanding, bietti_inductive_bias_ntk}. In order to prove the convergence of the eigenvalues, we use the Dominated Convergence theorem, leveraging the asymptotic results in Proposition \ref{prop:ordered_chaotic_ntk} and Theorems \ref{thm:ntk_eoc}, \ref{thm:ntk_residual}.\\

\section{Proofs}\label{proofsSection}
In this section, we provide proofs for the main results. We use several technical lemmas which we state and prove in the appendix. To distinguish between lemmas in the main text and those in the appendix, we call the later Appendix Lemmas. We refer to some results in the Appendix by the name `Fact'. These are simple results that have appeared (or are similar to) in previous works.

\subsection{Proofs on the limiting NTK:  \Cref{prop:ordered_chaotic_ntk}, Theorems \ref{thm:ntk_eoc}, \ref{thm:ntk_residual} and \ref{thm:scaled_resnet}} \label{sec:pr:limNTK}
In this section we prove the four theorems describing the limiting behaviour of $K^L$ or scaled versions of it.

\subsubsection{Proof of \Cref{prop:ordered_chaotic_ntk}: NTK in the ordered/chaotic phase}\label{sec:prop:ordered}
We prove the result for an FFNN architecture. The proof easily extends to a CNN architecture under \Cref{assumption:cnn}, using Appendix Lemmas \ref{lemma:correlation_convergence_convnet_tanh} and \ref{lemma:correlation_convergence_convnet_relu}.

Let $x,x' \in \mathbb{R}^d$ be two inputs. From lemma \ref{lemma:ffnn_ntk}, we have that
$$
K^l(x, x') = K^{l-1}(x, x') \dot{q}^l(x, x') + q^l(x,x')
$$
where $q^1(x,x') = \sigma_b^2 + \frac{\sigma_w^2}{d} x^Tx'$ and $q^{l}(x,x') = \sigma_b^2 + \sigma_w^2 \mathbb{E}_{f \sim \mathcal{N}(0, q^{l-1})}[\phi(f(x))\phi(f(x'))]$ and $\dot{q}^l(x, x') = \sigma_w^2\mathbb{E}_{f \sim \mathcal{N}(0, q^{l-1})}[\phi'(f(x))\phi'(f(x'))]$.
From facts \ref{fact:convergence_variance_ffnn_tanh}, \ref{fact:convergence_correlation_ffnn_ordered_tanh}, \ref{fact:convergence_correlation_ffnn_chaotic_tanh}, \ref{fact:convergence_correlation_relu_ordered}, and \ref{fact:convergence_derivative_f_ordered_chaotic_ffnn} in the appendix, in the ordered/chaotic phase, there exist $k, \beta, \eta, l_0>0$ and $\alpha \in (0,1)$ such that for all $l \geq l_0$ we have  
$$
\sup_{(x, x') \in B_\epsilon} |q^l(x,x') - k | \leq e^{-\beta l}
,
\quad \text{and} \quad \sup_{(x, x') \in B_\epsilon} |\dot{q}^l(x,x') - \alpha| \leq  e^{-\eta l}. 
$$
Therefore, there exists $M>0$ such that for any $l\geq l_0$ and $x, x' \in \mathbb{R}^d$
$$
K^l(x, x') \leq M.
$$
Letting $r_l = \sup_{(x,x') \in B_\epsilon} |K^l(x,x') - \frac{k}{1-\alpha}|$, we have 
$$
r_l \leq \alpha r_{l-1} + M e^{-\eta l} + e^{-\beta l}.
$$
We conclude using Appendix Lemma \ref{lemma:uppercounded_seq}.\\

\subsubsection{Proof of Theorem \ref{thm:ntk_eoc} : FFNN and CNN in the EOC} \label{sec:pr:thEOC}

We  prove the results for an FFNN architecture. In the case of a CNN architecture, under \Cref{assumption:cnn}, the NTK of a CNN is the same as that of an FFNN. Therefore, the results on the NTK of FFNN are all valid to the NTK of CNN $K^l_{\alpha,\alpha'}$for any $\alpha, \alpha'$.

%\paragraph{Case 1: FFNN.} 
Let $\epsilon \in (0,1)$, $E \subset \mathbb{R}^d$, $(\sigma_b, \sigma_w) \in$ EOC, and $x,x' \in \mathbb{R}^d$. Recall that $c^l(x,x') = \frac{q^l(x,x')}{\sqrt{q^l(x,x)q^l(x',x')}}$. Let $\gamma_l := 1 - c^l(x,x') $ and $f$ be the \emph{correlation function} defined by the recursive equation $c^{l+1} = f(c^l)$ (See appendix \ref{app:warmup_mean_field}). By definition, we have that 
$\dot{q}^l(x, x)=f'( c^{l-1}(x,x')) $. Let us first prove the result for ReLU. 

\begin{itemize}
\item \emph{$\phi=$ReLU}: From fact \ref{fact:variance_relu_eoc} in the appendix, we know that, when choosing the hyper-parameters $(\sigma_w, \sigma_b)$ in the EOC for ReLU, the variance $q^l(x,x)$ is constant w.r.t $l$ and is given by $q^l(x,x) = q^1(x,x)= \frac{\sigma_w^2}{d} ||x||^2$. Moreover, from fact \ref{fact:properties_correlation_function}, we have that $\dot{q}^l(x, x) = 1$. Therefore  
$$K^l(x, x) = K^{l-1}(x, x) + \frac{\sigma_w^2}{d} ||x||^2 = l \frac{\sigma_w^2}{d} ||x||^2 = l \tilde{K}^{\infty}(x,x)$$
which concludes the proof for $K^L(x,x)$. Note that the results is 'exact' for ReLU, which means the upper bound $\mathcal{O}(L^{-1})$ is valid but not optimal in this case. However, we will see that this bound is optimal for Tanh.\\

From Appendix Lemma \ref{lemma:asymptotic_expnasion_correlation_relu_ffnn}, we have that 

    $$
    \sup_{(x,x') \in B_\epsilon}\left|c^l(x,x') - 1 + \frac{\kappa}{l^2} -\kappa' \frac{\log(l)}{l^3} \right|= \mathcal{O}(l^{-3})
    $$
    and 
    $$
\sup_{(x,x') \in B_\epsilon} \left|f'(c^l(x,x')) - 1 + \frac{3}{l} - \kappa'' \frac{\log(l)}{l^2}  \right| = \mathcal{O}(l^{-2}).
$$

Using Appendix Lemma \ref{lemma:upper_bound_eoc} with $a_l = K^{l+1}(x,x'), b_l=q^{l+1}(x,x'), \lambda_l = f'(c^l(x,x'))$, we conclude that

$$\sup_{(x,x') \in B_\epsilon}\left|\frac{K^{l+1}(x,x')}{l}-\frac{1}{4} \frac{\sigma_w^2}{d} \|x\| \|x'\|\right|= \Theta(\log(l)l^{-1}).$$
Using the compactness of $B_{\epsilon}$, we conclude that 
$$\sup_{(x,x') \in B_\epsilon}\left|\frac{K^{l}(x,x')}{l}-\frac{1}{4} \frac{\sigma_w^2}{d} \|x\| \|x'\|\right|= \Theta(\log(l)l^{-1}).$$

\item $\phi=Tanh$: The proof in the case of Tanh is slightly different from that of ReLU. We use different technical lemmas to conclude. 

From Appendix Lemma \ref{lemma:asymptotic_expnasion_correlation_tanh_ffnn}, we have that
$$
    \sup_{(x,x') \in B_\epsilon}\left|c^l(x,x') - 1 + \frac{\kappa}{l} - \kappa (1 - \kappa^2 \zeta) \frac{\log(l)}{l^3} \right|= \mathcal{O}(l^{-3})
    $$
    where $\kappa = \frac{2}{f''(1)}>0$ and $\zeta = \frac{f^{3}(1)}{6}>0$. Moreover, we have that 
    $$
\sup_{(x,x') \in B_\epsilon} \left|f'(c^l(x,x')) - 1 + \frac{2}{l} - 2(1 - \kappa^2 \zeta) \frac{\log(l)}{l^2}  \right| = \mathcal{O}(l^{-2}).
$$

We conclude in the same way as in the case of ReLU using Appendix Lemma \ref{lemma:upper_bound_eoc}. The only difference is that, in this case, the limit of the sequence $b_l = q^{l+1}(x,x')$ is the limiting variance $q$ (from facts \ref{fact:convergence_correlation_ffnn_eoc_tanh}, \ref{fact:convergence_variance_ffnn_tanh}) does not depend on $(x,x')$.
\end{itemize}

% Let us prove now that $R_{L,N} = \Theta(N^2)$. There two cases depending on the activation function (ReLU-like or \mathcal{S}-smooth), however, the proof is similar. Let us prove the result for smooth activation functions belonging to class $\mathcal{S}$. Recall that
% $$
% f_{\infty}(x) = f_0(x) + \gamma(x, \mathcal{X}) (\mathcal{Z} - f_0(\mathcal{X}))
% $$
% In this limit, the only term that controls the generalization function is $\gamma(x, \mathcal{X})$. In the limit of large $L$, we have that 
% $$
% \gamma(x, \mathcal{X}) = \lambda e_1^T J^{-1} + \Theta(\log(L) L^{-1})
% $$
% where $e_1 = (1, ..., 1) \in \mathbb{R}^N$. This yields
% \begin{align*}
%     R_{L,N} &= \mathbb{E}_{x,y}[ (f_0(x) - y)^2] + a^2 \mathbb{E}_{\mathcal{D}}[(\sum_{i=1}^N f_0(x_i) - y_i)^2] \\
%     &+ 2 a \mathbb{E}_{x,y}[ (f_0(x) - y)]  \mathbb{E}_{\mathcal{D}}[\sum_{i=1}^N (f_0(x_i) - y_i)]  + \Theta(N \log(L) L^{-1})\\
%     &= (1 + a^2 N^2+ 2 a N) \mathbb{E}_{x,y}[(f_0(x) - y)^2] + \Theta((N \log(L) L^{-1}) = \Theta(N^2)
% \end{align*}
% which concludes the proof.

\subsubsection{Proof of Theorem \ref{thm:ntk_residual}: NTK for ResNET}\label{sec:pr:resNTK}

Similarly to the previous theorem, we  prove the result for a ResNet architecture with fully-connected layers and we use  \Cref{assumption:cnn} so that the dynamics of the correlation and NTK are exactly the same for FFNN, hence all results on FFNN apply to CNN.\\

%\paragraph{Case 1: ResNet with fully-connected layers.}
Let $\epsilon \in (0,1)$, $E \subset \mathbb{R}^d$, and $x, x' \in \mathbb{R}^d$. We first prove the result for the diagonal terms $K^{L}_{res}(x, x)$, we deal afterwards with off-diagonal terms $K^{L}_{res}(x, x')$. 
\begin{itemize}
    \item Diagonal terms: from fact \ref{fact:correlatin_function_relu}, we have that $\dot{q}^l(x, x)=\frac{\sigma_w^2}{2} f(1) = \frac{\sigma_w^2}{2}$. Moreover, it is easy to see that the variance terms for a ResNet follow the recursive formula $q^l(x,x) = q^{l-1}(x,x) + \sigma_w^2/2 \times q^{l-1}(x,x)$, hence
    
    \begin{equation}\label{equation:variance_resnet}
        q^l(x,x)= (1 + \sigma_{w}^2/2)^{l-1} \frac{\sigma_w^2}{d} \|x\|^2.
    \end{equation}
    
    Recall that the recursive formula of NTK of a ResNet with fully-connected layers is given by (Appendix Lemma \ref{lemma:resnet_ntk})
    $$
K_{res}^l(x, x') = K_{res}^{l-1}(x, x') (\dot{q}^l(x, x')+1) + q^l(x,x').
$$
Hence, for the diagonal terms we obtain
$$
K_{res}^l(x, x) = K_{res}^{l-1}(x, x) \left(\frac{\sigma_w^2}{2}+1\right) + q^l(x,x).
$$
Letting $\hat{K}_{res}^l = K_{res}^l / \left(1 + \frac{\sigma_w^2}{l}\right)^{l-1}$ yields
$$
\hat{K}_{res}^l(x, x) = \hat{K}_{res}^{l-1}(x, x) + \frac{\sigma_w^2}{d} \|x\|^2.
$$
Therefore, $\Bar{K}_{res}^l(x,x) = \frac{\hat{K}_{res}^1(x,x)}{l} + \left(1 - 1/l\right)\frac{\sigma_w^2}{d} \|x\|^2$, the conclusion is straightforward since $E$ is compact and $\hat{K}_{res}^1(x,x)$ is continuous which implies that it is uniformly bounded on $E$.

    \item Off-diagonal terms: the argument is similar to that of Theorem \ref{thm:ntk_eoc} with few key differences. From Appendix Lemma \ref{lemma:asymptotic_expansion_resnet} we have that 
    
    $$
    \sup_{(x,x') \in B_\epsilon}\left|c^l(x,x') - 1 + \frac{\kappa_{\sigma_w}}{l^2} - \kappa_{\sigma_w}' \frac{\log(l)}{l^3} \right|= \mathcal{O}(l^{-3})
    $$
    where $\kappa_{\sigma_w}, \kappa_{\sigma_w}'>0$. Moreover, we have that 
    $$
\sup_{(x,x') \in B_\epsilon} \left|f'(c^l(x,x')) - 1 + \frac{3 (1 + \frac{2}{\sigma_w^2})}{l} - \kappa_{\sigma_w}'' \frac{\log(l)}{l^2}  \right| = \mathcal{O}(l^{-2}).$$

Let $\alpha = \frac{\sigma_w^2}{2}$. We also have $\dot{q}^{l+1}(x,x') = \alpha f'(c^{l}(x,x'))$ where $f$ is the ReLU correlation function given in fact \ref{fact:correlatin_function_relu}. It follows that for all $(x,x') \in B_\epsilon$

$$
1 + \dot{q}^{l+1}(x,x') =  (1 + \alpha) (1 - 3 l^{-1} + \zeta \frac{\log(l)}{l^2} + \bigO(l^{-3}))
$$
for some constant $\zeta \neq 0$ that does not depend on $x,x'$. The bound $\bigO$ does not depend on $x,x'$ either.
Now let $a_l = \frac{K^{l+1}_{res}(x,x')}{(1 + \alpha)^{l}}$. Using the recursive formula of the NTK, we obtain
$$
a_l = \lambda_l a_{l-1} + b_l
$$
where $\lambda_l = 1 - 3 l^{-1} + \zeta \frac{\log(l)}{l^2} + \bigO(l^{-3})$, $b_l = \frac{\sigma_w^2}{d}\sqrt{\|x\|\|x'\|} f(c^l(x,x')) = q(x,x') + \bigO(l^{-2})$ with $q(x,x') =  \frac{\sigma_w^2}{d}\sqrt{\|x\|\|x'\|} $ and where we used the fact that $c^l(x,x') = 1 + \bigO( l^{-2})$ (Appendix Lemma \ref{lemma:asymptotic_expnasion_correlation_relu_ffnn}) and the formula for ResNet variance terms given by Eq. \eqref{equation:variance_resnet}. Observe that all bounds $\bigO$ are independent from the inputs $(x,x')$. Therefore, using Appendix Lemma \ref{lemma:upper_bound_eoc}, we have 
$$\sup_{x,x' \in B_\epsilon} \big|K^{L+1}_{res}(x,x')/L(1+\alpha)^L - \Bar{K}^\infty_{res}(x,x') \big| = \Theta(L^{-1} \log(L)),$$
which can also be written as
$$\sup_{x,x' \in B_\epsilon} \big|K^{L}_{res}(x,x')/(L-1)(1+\alpha)^{L-1} - \Bar{K}^\infty_{res}(x,x') \big| = \Theta(L^{-1} \log(L)),$$
We conclude by observing that $K^{L}_{res}(x,x')/(L-1)(1+\alpha)^{L-1} = K^{L}_{res}(x,x')/L(1+\alpha)^{L-1} + \mathcal{O}(L^{-1})$ where $\bigO$ can be chosen to depend only on $\epsilon$.
\end{itemize}

% Let $x, x'$ be two inputs. Using lemma \ref{lemma:resnet_cnn_ntk}, we have that for all $\alpha,\alpha'$

% \begin{align*}
%     K^{res, l}_{\alpha,\alpha'} &= K^{res, l-1}_{\alpha,\alpha'} + \frac{1}{2k+1} \sum_{\beta} \big [\dot{q}^l_{\alpha+\beta, \alpha'+\beta} K^{l-1}_{\alpha+\beta,\alpha'+\beta} + \hat{q}^{l}_{\alpha+\beta, \alpha'+\beta}\big]\\
% \end{align*} 

% where we have $\frac{1}{2k+1} \sum_{\beta}q^{l}_{\alpha+\beta, \alpha'+\beta} =  (1 + \alpha)^{l-1}(q + O(l^{-2}))$ by appendix lemma \ref{lemma:asymptotic_expansion_resnet}, where $q>0$ is a constant.\\

% Writing this with Hadamard product
% \begin{align*}
%     K^{res, l} &=  (1 + \alpha f'(C_{l-1}))\circ K^{res, l-1} + (1 + \alpha)^{l-1}(q + O(l^{-2}))e_1
% \end{align*} 

% Letting $\Theta_l = K^{res, l} / (1+\alpha)^{l-1}$, we have that 

% \begin{align*}
%     \Theta_l &=  \frac{1 + \alpha f'(C_{l-1})}{1 + \alpha} \circ \Theta_{l-1} + (q + O(l^{-2}))e_1
% \end{align*} 
% we apply appendix lemma \ref{lemma:asymptotic_expansion_resnet} to get
% $$
% \frac{1 + \alpha f'(C_{l-1})}{1 + \alpha} = (1 - 3 l^{-1})e_1 + \zeta l^{-2} + O(l^{-3}))
% $$where $\zeta = (\zeta_{\alpha, \alpha'})$. from here the proof is similar to the Fully connected layers case, we apply appendix lemma \ref{lemma:upper_bound_relu_ntk_eoc} element-wise. We conclude using the fact that the Taylor expansion bounds are uniform w.r.t $(x,x') \in B_\epsilon$, which concludes the proof for Convolutional layers.
% \end{itemize}

% The proof for the $R_{L,N}$ is exactly the same as in theorem \ref{thm:ntk_eoc}.

\subsubsection{Proof of Theorem \ref{thm:scaled_resnet}: NTK for Scaled ResNet}\label{sec:pr:scaledres}

We use the same techniques as in the non-scaled case. Let us prove the result for fully connected layers, the proof for convolutional layers follows the same analysis.

Let $\epsilon \in (0,1)$ and $x,x' \in B_{\epsilon}$ be two inputs. We first prove the result for the diagonal term $K^{L}_{res}(x, x)$ then $K^{L}_{res}(x, x')$. 
\begin{itemize}
    \item We have that $\dot{q}^l(x, x)=\frac{\sigma_w^2}{2l} f(1) = \frac{\sigma_w^2}{2l}$. Moreover, we have $q^l(x,x) = q^{l-1}(x,x) + \sigma_w^2/2l \times q^{l-1}(x,x) = \left[\prod_{k=1}^l(1 + \sigma_{w}^2/2k) \right]\frac{\sigma_w^2}{d} \|x\|^2$.  Recall that
    $$
K_{res}^l(x, x) = K_{res}^{l-1}(x, x) (1 + \frac{\sigma_w^2}{2l}) + q^l(x,x)
$$\\
letting $k_l' = \frac{K_{res}^l(x,x)}{\prod_{k=1}^l(1 + \sigma_{w}^2/2k)}$, we obtain
$$
k_l' = k_{l-1}'  + \frac{\sigma_w^2}{d} \|x\|.
$$
Using the fact that $\prod_{k=1}^l(1 + \sigma_{w}^2/2k) = \Theta(l^{\sigma_w^2/2})$, we conclude for $K_{res}^l(x,x)$.
    
    \item Recall that
    $$
K_{res}^l(x, x') = K_{res}^{l-1}(x, x') (\dot{q}^l(x, x')+1) + q^l(x,x').
$$\\
Let $c^l:= c^l(x,x')$. From Appendix Lemma \ref{lemma:asymptotic_expansion_scaled_resnet} we have that 
$$
1 - c^l = \frac{\zeta}{\log(l)^2} - \frac{\nabla}{\log(l)^3} + o\left(\frac{1}{\log(l)^3}\right),
$$
$\zeta = \frac{16}{s^2 \sigma_w^4}$ and $\nabla>0$. Using the Taylor expansion of $f'$ as in Appendix Lemma \ref{lemma:asymptotic_expnasion_correlation_relu_ffnn}, it follows that
$$
f'(c^l(x,x')) = 1 - \frac{6}{\sigma_w^2} \log(l)^{-1} + \zeta' \log(l)^{-2} + \bigO(\log(l)^{-3})
$$
where $\zeta' = \frac{\nabla}{\sqrt{2 \pi \zeta}}$. We obtain 
$$
1 + \dot{q}^l(x,x') =  1 + \frac{\sigma_w^2}{2l} - 3 l^{-1} \log(l)^{-1} + \zeta'' l^{-1} \log(l)^{-2} + \bigO(l^{-1} \log(l)^{-3})
$$
where $\zeta'' = \frac{\sigma_w^2}{2} \zeta'$. Letting $a_l = \frac{K^{l+1}_{res}(x,x')}{\prod_{k=1}^l(1 + \sigma_{w}^2/2k)}$, we obtain
$$
a_l = \lambda_l a_{l-1} + b_l
$$
where $\lambda_l = 1 - l^{-1} - 3 l^{-1} \log(l)^{-1} + \bigO(l^{-1} \log(l)^{-2})$, $b_l = \sqrt{q^1(x,x)}\sqrt{q^1(x',x')} f(c^l(x,x')) = q(x,x') + \bigO(\log(l)^{-2})$ with $q =  \sqrt{q^1(x,x)}\sqrt{q^1(x',x')} $ and where we used the fact that $c^l = 1 + \bigO( \log(l)^{-2})$ (Appendix Lemma \ref{lemma:asymptotic_expansion_scaled_resnet}). \\

Now we proceed in the same way as in the proof of Appendix Lemma \ref{lemma:upper_bound_eoc}. Let $x_l = \frac{a_l}{l} - q$, then there exists $M_1, M_2 > 0$ such that 
$$
x_{l-1} (1 - \frac{1}{l}) - M_1 l^{-1} \log(l)^{-1} \leq x_l \leq x_{l-1} (1 - \frac{1}{l}) - M_2 l^{-1} \log(l)^{-1}.
$$
Therefore, there exists $l_0$ independent of $(x,x')$ such that for all $l \geq l_0$

$$
x_l \leq x_{l_0} \prod_{k=l_0}^l(1-\frac{1}{k}) - M_2 \sum_{k=l_0}^l \prod_{j=k+1}^l(1 - \frac{1}{j}) k^{-1} \log(k)^{-1} 
$$
and 
$$
x_l \geq x_{l_0} \prod_{k=l_0}^l(1-\frac{1}{k}) - M_1 \sum_{k=l_0}^l \prod_{j=k+1}^l(1 - \frac{1}{j}) k^{-1} \log(k)^{-1}. 
$$

After simplification, we have that 

$$
\sum_{k=l_0}^l \prod_{j=k+1}^l(1 - \frac{1}{j}) k^{-1} \log(k)^{-1} = \Theta\left( \frac{1}{l} \int^l \frac{1}{\log(t)} dt\right) = \Theta( \log(l)^{-1}),
$$
where we have used the asymptotic approximation of the Logarithmic Intergal function $\text{Li}(x) = \int^t \frac{1}{\log(t)} \sim_{x\rightarrow \infty} \frac{x}{\log(x)}.$

We conclude that $\alpha_L = L \times  \prod_{k=1}^l(1 + \sigma_{w}^2/2k) \sim L^{1 + \frac{\sigma_w^2}{2}}$ and the convergence rate of the NTK is now $\Theta(\log(L)^{-1})$ which is better than $\Theta(L^{-1} \log(L))$. The convergence is uniform over the set $B_\epsilon$.\\

In the limit of large $L$, the matrix NTK of the scaled resnet has the following form
$$
\hat{AK}^l_{res} = q U + \log(L)^{-1} \Theta(M_L)
$$
where $U$ is the matrix of ones, and $M_L$ has all elements but the diagonal equal to 1 and the diagonal terms are $\mathcal{O}(L^{-1} \log(L)) \rightarrow 0$. Therefore, $M_L$ is inversible for large $L$ which makes $\hat{K}^l_{res}$ also inversible. Moreover, observe that the convergence rate for scaled resnet is $\log(L)^{-1}$ which means that for the same depth $L$, the NTK remains far more expressive for scaled resnet compared to standard resnet, this is particularly important for the generalization. 

\end{itemize}

\subsection{Proof of Proposition \ref{prop:spectral_decomposition_Sd}: Spectral analysis of the NTK} \label{sec:pr:spectral}

We start by presenting a brief review of the theory of Spherical Harmonics, see  \citep{macrobert} for more details. 

\subsubsection{Some facts on Spherical harmonics}
Let $\mathbb{S}^{d-1}$ be the unit sphere in $\mathbb{R}^d$ defined by $\mathbb{S}^{d-1} = \{ x \in \mathbb{R}^d : \|x\|_2 = 1\}$. For some $k \geq 1$, there exists a set $(Y_{k,j})_{1\leq j \leq N(d,k)}$ of Spherical Harmonics of degree $k$ with $N(d,k) = \frac{2k + d - 2}{k} {k + d-3 \choose d-2}$.\\

The set of functions $(Y_{k,j})_{k\geq1, j \in [1:N(d,k)]}$ form an orthonormal basis with respect to the uniform measure on the unit sphere $\mathbb{S}^{d-1}$.\\
For some function $g$, the Hecke--Funk formula is given by
$$
\int_{\mathbb{S}^{d-1}} g(\langle x, w \rangle)Y_{k,j}(w) d\nu_{d-1}(w) = \frac{\Omega_{d-1}}{\Omega_d} Y_{k,j}(x) \int_{-1}^1 g(t) P^d_k(t) (1-t^2)^{(d-3)/2} dt
$$
where $\nu_{d-1}$ is the uniform measure on the unit sphere $\mathbb{S}^{d-1}$, $\Omega_d$ is the volume of the unit sphere $\mathbb{S}^{d-1}$, and $P^d_{k}$ is the multi-dimensional Legendre polynomials given explicitly by Rodrigues' formula
$$
P^d_k(t) = \left(-\frac{1}{2}\right)^{k} \frac{\Gamma(\frac{d-1}{2})}{\Gamma(k + \frac{d-1}{2})} (1-t^2)^{\frac{3-d}{2}} \left( \frac{d}{dt}\right)^k (1-t^2)^{k + \frac{d-3}{2}}.
$$
$(P^d_k)_{k\geq0}$ form an orthogonal basis of $L^2([-1,1], (1-t^2)^{\frac{d-3}{2}}dt)$, i.e. 
$$
\langle P^d_k, P^d_{k'} \rangle_{L^2([-1,1], (1-t^2)^{\frac{d-3}{2}}dt)} = \delta_{k,k'} 
$$
where $\delta_{ij}$ is the Kronecker symbol. Moreover, we have 
$$
\| P^d_k \|^2_{L^2([-1,1], (1-t^2)^{\frac{d-3}{2}}dt)} = \frac{(k + d-3)!}{(d-3) (k - d + 3 )!}.
$$

Using the Heck--Funk formula, we can easily conclude that any dot product kernel on the unit sphere $\mathbb{S}^{d-1}$, i.e. kernel of the form $\kappa(x,x') = g(\langle x, x' \rangle)$ can be decomposed on the Spherical Harmonics basis. Indeed, for any $x,x' \in \mathbb{S}^{d-1}$, the decomposition on the spherical harmonics basis yields 
$$
\kappa(x,x') = \sum_{k\geq 0} \sum_{j=1}^{N(d,k)} \left[\int_{\mathbb{S}^{d-1}}g(\langle w, x' \rangle)Y_{k,j}(w)d\nu_{d-1}(w)\right] Y_{k,j}(x).
$$
Using the Hecke--Funk formula yields
$$
\kappa(x,x') = \sum_{k\geq0} \sum_{j=1}^{N(d,k)}\left[ \frac{\Omega_{d-1}}{\Omega_d} \int_{-1}^1 g(t) P^d_k(t) (1-t^2)^{(d-3)/2} dt\right] Y_{k,j}(x)Y_{k,j}(x'),
$$
we conclude that 
$$
\kappa(x,x') = \sum_{k \geq 0} \mu_k \sum_{j=1}^{N(d,k)} Y_{k,j}(x)Y_{k,j}(x')
$$
where $\mu_k = \frac{\Omega_{d-1}}{\Omega_d} \int_{-1}^1 g(t) P^d_k(t) (1-t^2)^{(d-3)/2} dt$.

We now use these result in the proof of the next theorem.

\subsubsection{Proof of Proposition \ref{prop:spectral_decomposition_Sd}} \label{sec:proof:scaled}

From the recursive formulas of the NTK for FFNN, CNN and ResNet architectures, it is straightforward that, on the unit sphere $\mathbb{S}^{d-1}$, the kernel $\kappa^L$ is zonal in the sense that it depends only on the scalar product. More precisely, for all $L\geq 1$, there exists a function $g^L$ such that for all $x,x' \in \mathbb{S}^{d-1}$
$$
\kappa^L(x,x') = g^L(\langle x,x'\rangle).
$$
Using the previous results on Spherical Harmonics, we have that for all $x,x' \in \mathbb{S}^{d-1}$
$$
\kappa^L(x,x') = \sum_{k\geq 0 } \mu^L_k \sum_{j=1}^{N(d,k)} Y_{k,j}(x) Y_{k,j}(x')
$$
where $\mu^L_k = \frac{\Omega_{d-1}}{\Omega_d} \int_{-1}^1 g^L(t) P^d_k(t) (1-t^2)^{(d-3)/2} dt$.

For $k=0$, we have that for all $L \geq 1$,
$\mu^L_0 = \frac{\Omega_{d-1}}{\Omega_d} \int_{-1}^1 g^L(t) (1-t^2)^{(d-3)/2} dt$. By a simple dominated convergence argument, we have that $\lim_{L \rightarrow \infty} \mu^L_0 = q \lambda \frac{\Omega_{d-1}}{\Omega_d} \int_{-1}^1 (1-t^2)^{(d-3)/2} dt > 0$, where $q , \lambda$ are given in Theorems \ref{thm:ntk_eoc}, \ref{thm:ntk_residual} and Proposition \ref{prop:ordered_chaotic_ntk} (where we take $q=1$ for the Ordered/Chaotic phase initialization in Proposition \ref{prop:ordered_chaotic_ntk}). Using the same argument, we have that for $k\geq 1$, $\lim_{L \rightarrow \infty} \mu^L_k = q\lambda \frac{\Omega_{d-1}}{\Omega_d} \int_{-1}^1 P^d_{k}(t)(1-t^2)^{(d-3)/2} dt = q\lambda \frac{\Omega_{d-1}}{\Omega_d} \langle P^d_0, P^d_k \rangle_{L^2([-1,1], (1-t^2)^{\frac{d-3}{2}} dt)} = 0$.

\section*{Acknowledgements}
The project leading to this work has received funding from the European Research Council
(ERC) under the European Union’s Horizon 2020 research and innovation programme (grant agreement No 834175).

\newpage
\bibliography{bibliography}
\bibliographystyle{plain}

\newpage
\onecolumn
\section*{Appendix}
\setcounter{equation}{0}
\setcounter{lemma}{0}
\setcounter{prop}{0}
\setcounter{corollary}{0}
\setcounter{definition}{0}
\setcounter{section}{-1}
\setcounter{assumption}{0}

\section{Some facts on the infinite-width regime and on NTKs}\label{section:infinite_width_limit}

%\section{The infinite-width limit}
\subsection{Gaussian process limit}
In this section we recall some results on the Gaussian process limit of the of the function $f_\theta(x) = s(y_\theta^L(x))$ modelled as a DNN, when the weights parameters are drawn from Gaussian distributions.
\subsubsection{Forward propagation}\label{subsection:aforward_prop}
\paragraph{FeedForward Neural Network.}
For some input $x \in \mathbb{R}^{d}$, the propagation of this input through the network is given by Eq. \eqref{equation:ffnn_net}. When we take the limit $n_{l-1} \rightarrow \infty$ recursively over $l$, this implies, using the Central Limit Theorem, that $y_i^{l} (x)$ is a Gaussian variable for any input $x$. This gives an error of order $\mathcal{O}(1/\sqrt{n_{l-1}})$ (standard Monte Carlo error). More generally, an approximation of the random process $y_i^l(.)$  by a Gaussian process was first proposed by \cite{neal} in the single layer case and has been extended to the multiple layer case by \cite{lee_nngp} and \cite{matthews}. The limiting Gaussian process kernels follow a recursive formula given by, for any inputs $x,x'\in \mathbb R^d$
\begin{align*}
\kappa^l(x,x') &= \mathbb{E}[y^l_i(x)y^l_i(x')]\\
&= \sigma^2_b + \sigma^2_w \mathbb{E}[\phi(y^{l-1}_i(x))\phi(y^{l-1}_i(x'))]\\
&= \sigma^2_b + \sigma^2_w \Psi_{\phi} (\kappa^{l-1}(x,x), \kappa^{l-1}(x,x'), \kappa^{l-1}(x',x')),
\end{align*}
where $\Psi_{\phi}$ is a function that only depends on the activation function $\phi$. This provides a simple recursive formula for the computation of the kernel $\kappa^l$; see, e.g., \cite{lee_nngp} for more details.

\paragraph{Convolutional Neural Networks.}
The propagation of the input through the network is described by Eq. \eqref{equation:convolutional_net}.
The infinite-width approximation with 1D CNN yields a recursion for the kernel. However, the infinite-width here means infinite number of channels, with a Monte Carlo error of $\mathcal{O}(1/\sqrt{n_{l-1}})$. The kernel in this case depends on the choice of the neurons in the channel and is given by 
$$
\kappa^l_{\alpha,\alpha'}(x,x') = 
\mathbb{E}[y^l_{i,\alpha}(x)y^l_{i,\alpha'}(x')] =   \sigma_b^2 + \frac{\sigma_w^2}{2k + 1} \sum_{\beta \in ker} \mathbb{E}[\phi(y^{l-1}_{1,\alpha + \beta}(x)) \phi(y^{l-1}_{1,\alpha' + \beta}(x'))]
$$
so that 
$$
\kappa^l_{\alpha,\alpha'}(x,x')
=   \sigma_b^2 + \frac{\sigma_w^2}{2k + 1} \sum_{\beta \in ker} F_{\phi} (\kappa^{l-1}_{\alpha+\beta,\alpha'+\beta}(x,x), \kappa^{l-1}_{\alpha+\beta,\alpha'+\beta}(x,x'), \kappa^{l-1}_{\alpha+\beta,\alpha'+\beta}(x',x')).
$$
The convolutional kernel $\kappa^l_{\alpha,\alpha'}$ has the `self-averaging' property; i.e. it is an average over the kernels corresponding to different combination of neurons in the previous layer. However, it is easy to simplify the analysis in this case by studying the average kernel per channel defined by $\hat{\kappa}^l = \frac{1}{N^2} \sum_{\alpha,\alpha'} \kappa^l_{\alpha,\alpha'}$. Indeed, by summing terms in the previous equation and using the fact that we use circular padding, we obtain
$$
\hat{\kappa}^l(x,x') = \sigma_b^2 + \sigma_w^2 \frac{1}{N^2} \sum_{\alpha,\alpha'} F_\phi(\kappa^{l-1}_{\alpha,\alpha'}(x,x), \kappa^{l-1}_{\alpha,\alpha'}(x,x'), \kappa^{l-1}_{\alpha,\alpha'}(x',x')).
$$
This expression is similar in nature to that of FFNN. We will use this observation in the proofs.\\

Note that our analysis only requires the approximation that, in the infinite-width limit, for any two inputs $x,x'$, the variables $y^l_i(x)$ and $y^l_i(x')$ are Gaussian with covariance $\kappa^l(x,x')$ for FFNN, and $y^l_{i,\alpha}(x)$ and $y^l_{i,\alpha'}(x')$ are Gaussian with covariance $\kappa^l_{\alpha,\alpha'}(x,x')$ for CNN. We do not need the much stronger approximation that the process $y^l_i(x)$ ($y^l_{i,\alpha}(x)$ for CNN) is a Gaussian process.

\paragraph{Residual Neural Networks.}

The propagation for ResNet architectures is described by \eqref{resNN:def}.
The infinite-width limit approximation for ResNet yields similar results with an additional residual terms. It is straighforward to see that, in the case of a ResNet with FFNN-type layers, we have that
\begin{align*}
\kappa^l(x,x') = \kappa^{l-1}(x,x') + \sigma^2_b + \sigma^2_w F_{\phi} (\kappa^{l-1}(x,x), \kappa^{l-1}(x,x'), \kappa^{l-1}(x',x')),
\end{align*}
whereas for ResNet with CNN-type layers, we have that 
\begin{equation*}
\begin{split}
\kappa^l_{\alpha,\alpha'}(x,x')
&=  \kappa^{l-1}_{\alpha,\alpha'}(x,x') +  \sigma_b^2 \\
&+ \frac{\sigma_w^2}{2k + 1} \sum_{\beta \in ker} F_{\phi} (\kappa^{l-1}_{\alpha+\beta,\alpha'+\beta}(x,x), \kappa^{l-1}_{\alpha+\beta,\alpha'+\beta}(x,x'), \kappa^{l-1}_{\alpha+\beta,\alpha'+\beta}(x',x')).
\end{split}
\end{equation*}

\subsubsection{Gradient Independence}\label{section:gradient_independence}
% \textcolor{red}{Is it really useful for the paper?}
In the literature of signal propagation in DNNs, an ubiquitous approximation is that of the gradient independence which is similar in nature to the practice of feedback alignment \citep{timothy_feedback}. This approximation states that, for wide neural networks, the weights used for forward propagation are independent from those used for back-propagation. When used for the computation of Neural Tangent Kernel, this approximation was proven to give the exact computation for standard architectures such as FFNN, CNN and ResNets \cite{yang_tensor3_2020} (Theorem D.1). 

This result has been extensively used in the literature as an approximation before being proved to yields exact computation for the NTK, and theoretical results derived under this approximation were verified empirically; see references below. 

\paragraph{Gradient Covariance back-propagation.} Analytical formulas for gradient covariance back-propagation were derived using this result, in \citep{hayou, samuel, yang2017meanfield, lee_nngp, poole, xiao_cnnmeanfield, yang2019scaling}. Empirical results showed an excellent match for FFNN in \cite{samuel}, for Resnets in \cite{yang2019scaling} and for CNN in \cite{xiao_cnnmeanfield}. 

\paragraph{Neural Tangent Kernel.}The Gradient Independence approximation was implicitly used in \cite{jacot} to derive the infinite-width Neural Tangent Kernel (See \cite{jacot}, Appendix A.1). Authors have found that this infinite-width NTK computed with the Gradient Independence approximation yields excellent match with empirical (exact) NTK. 

We use this result in our proofs and we refer to it simply by the Gradient Independence.

\subsection{On the theory of signal propagation in DNNs}\label{app:warmup_mean_field}
%\subsection{Notation}
For FFNN layers, let $q^l(x):=q^l(x,x)$ be the variance of $y^l_1(x)$ (the choice of the index $1$ is not important since, in the infinite-width limit, the random variables $(y^l_i(x))_{i \in [1:N_l]}$ are iid). Let $q^l(x,x')$, resp. $c^l_1(x,x')$ be the covariance, resp. the correlation between $y^l_1(x)$ and $y^l_1(x')$. For Gradient back-propagation, let $\Tilde{q}^l(x,x')$ be the Gradient covariance defined by  $\Tilde{q}^l(x,x')= \mathbb{E}\left[ \frac{\partial \mathcal{L}}{\partial y^l_1}(x) \frac{\partial \mathcal{L}}{\partial y^l_1}(x')\right]$ where $\mathcal{L}$ is some loss function. Similarly, let $\Tilde{q}^l(x)$ be the Gradient variance at point $x$. We also define $\Dot{q}^l(x,x') =\sigma_w^2 \mathbb{E}[\phi'(y^{l-1}_1(x))\phi'(y^{l-1}_1(x'))] $.\\
For CNN layers, we use similar notation across channels. Let $q^l_{\alpha}(x)$ be the variance of $y^l_{1,\alpha,}(x)$ (the choice of the index $1$ is not important here either since, in the limit of infinite number of channels, the random variables $(y^l_{i,\alpha}(x))_{i \in [1:N_l]}$ are iid). Let $q^l_{\alpha, \alpha'}(x,x')$ the covariance between $y^l_{1,\alpha}(x)$ and $y^l_{1,\alpha'}(x')$, and $c^l_{\alpha,\alpha'}(x,x')$ the corresponding correlation. We also define the pseudo-covariance $\hat{q}^l_{\alpha,\alpha'}(x,x') = \sigma_b^2 + \sigma_w^2 \mathbb{E}[\phi(y^{l-1}_{1,\alpha}(x))\phi(y^{l-1}_{1,\alpha'}(x'))]$ and $\Dot{q}^l_{\alpha,\alpha'}(x,x') = \sigma_w^2 \mathbb{E}[\phi(y^{l-1}_{1,\alpha}(x))\phi(y^{l-1}_{1,\alpha'}(x'))]$.\\
The Gradient covariance  is defined by $\Tilde{q}^l_{\alpha,\alpha'}(x,x')= \mathbb{E}\left[ \frac{\partial \mathcal{L}}{\partial y^l_{1,\alpha}}(x) \frac{\partial \mathcal{L}}{\partial y^l_{1,\alpha'}}(x')\right]$.

\subsubsection{Covariance propagation and the Edge of Chaos}
\paragraph{Covariance propagation for FFNN.}
In Section \ref{subsection:aforward_prop}, we derived the covariance kernel propagation in an FFNN. For two inputs $x,x' \in \mathbb{R}^d$, we have
\begin{equation}\label{equation:covariance_prop_ffnn}
q^l(x,x') = \sigma^2_b + \sigma^2_w \mathbb{E}[\phi(y^{l-1}_i(x))\phi(y^{l-1}_i(x'))]    
\end{equation}
this can be written as 
$$
q^l(x,x') = \sigma^2_b + \sigma^2_w \mathbb{E}\left[\phi\left(\sqrt{q^l(x)} Z_1\right)\phi\left(\sqrt{q^l(x')}(c^{l-1} Z_1 + \sqrt{1 - (c^{l-1})^2} Z_2\right)\right], \quad Z_1, Z_2 \overset{iid}{\sim} \mathcal{N}(0,1),
$$
with $c^{l-1} := c^{l-1}(x,x')$.\\
With ReLU, and since ReLU is positively homogeneous (i.e. $\phi(\lambda x) = \lambda \phi(x)$ for $\lambda\geq0$), we have that 
$$
q^l(x,x') = \sigma^2_b + \frac{\sigma^2_w}{2} \sqrt{q^l(x)}\sqrt{q^l(x')} f(c^{l-1})
$$
where $f$ is the ReLU correlation function given by \cite{hayou}
\begin{align*}
    f(c) = \frac{1}{\pi}(c \arcsin{c} + \sqrt{1- c^2}) + \frac{1}{2}c.
\end{align*}
\paragraph{Covariance propagation for CNN.}
The only difference with FFNN is that the independence is across channels and not neurons. Simple calculus yields
$$
q^l_{\alpha,\alpha'}(x,x') = 
\mathbb{E}[y^l_{i,\alpha}(x)y^l_{i,\alpha'}(x')] =   \sigma_b^2 + \frac{\sigma_w^2}{2k + 1} \sum_{\beta \in ker} \mathbb{E}[\phi(y^{l-1}_{1,\alpha + \beta}(x)) \phi(y^{l-1}_{1,\alpha' + \beta}(x'))]
$$
Observe that 
\begin{equation}\label{equation:covariance_prop_cnn}
q^l_{\alpha,\alpha'}(x,x') = \frac{1}{2k+1} \sum_{\beta in ker} \hat{q}^l_{\alpha+\beta,\alpha'+\beta}(x,x')
\end{equation}
With ReLU, we have
$$
q^l_{\alpha,\alpha'}(x,x')=   \sigma_b^2 + \frac{\sigma_w^2}{2k + 1} \sum_{\beta \in ker} \sqrt{q^l_{\alpha+\beta}(x)}\sqrt{q^l_{\alpha'+\beta}(x')}f(c^{l-1}_{\alpha+\beta,\alpha'+\beta}(x,x')).
$$
\paragraph{Covariance propagation for ResNet with ReLU.}
In the case of ResNet, only an added residual term shows up in the recursive formula. For a ResNet with FFNN layers, the recursion reads

\begin{equation}\label{equation:covariance_prop_resnet_ffnn}
q^l(x,x') =  q^{l-1}(x,x') +  \sigma^2_b + \frac{\sigma^2_w}{2} \sqrt{q^l(x)}\sqrt{q^l(x')} f(c^{l-1})
\end{equation}
with CNN layers, we have instead
\begin{equation}\label{equation:covariance_prop_resnet_cnn}
q^l_{\alpha,\alpha'}(x,x') = q^{l-1}_{\alpha,\alpha'}(x,x') +   \sigma_b^2 + \frac{\sigma_w^2}{2k + 1} \sum_{\beta \in ker} \sqrt{q^l_{\alpha+\beta}(x)}\sqrt{q^l_{\alpha'+\beta}(x')}f(c^{l-1}_{\alpha+\beta,\alpha'+\beta}(x,x'))
\end{equation}

\paragraph{Edge of Chaos (EOC)}
Let $x \in \mathbb{R}^d$ be an input. The convergence of $q^l(x)$ as $l$ increases has been studied by \cite{samuel} and \cite{hayou}. In particular, under weak regularity conditions, it is proven that $q^l(x)$ converges to a point $q(\sigma_b, \sigma_w)>0$ independent of $x$ as $l \rightarrow \infty$. The asymptotic behaviour of the correlations $c^l(x,x')$ between $y^l(x)$ and $y^l(x')$ for any two inputs $x$ and $x'$ is also driven by $(\sigma_b, \sigma_w)$: the dynamics of $c^l$ is controlled by a function $f$ i.e. $c^{l+1} = f(c^l)$ called the correlation function. The authors define the EOC as the set of parameters $(\sigma_b, \sigma_w)$ such that $\sigma_w^2 \mathbb{E}[\phi'(\sqrt{q(\sigma_b, \sigma_w)}Z)^2] =1$ where $Z \sim \mathcal{N}(0, 1)$. Similarly the Ordered, resp. Chaotic, phase is defined by $\sigma_w^2 \mathbb{E}[\phi'(\sqrt{q(\sigma_b, \sigma_w)}Z)^2] < 1$, resp. $\sigma_w^2 \mathbb{E}[\phi'(\sqrt{q(\sigma_b, \sigma_w)}Z)^2] > 1$. In the Ordered phase, the gradient will vanish as it backpropagates through the network, and the correlation $c^l(x,x')$ converges exponentially to $1$. Hence the output function becomes constant (hence the name 'Ordered phase'). In the Chaotic phase, the gradient explodes and the correlation converges exponentially to some limiting value $c<1$ which results in the output function being discontinuous everywhere (hence the 'Chaotic' phase name). In the EOC, the second moment of the gradient remains constant throughout the backpropagation and the correlation converges to 1 at a sub-exponential rate, which allows deeper information propagation. Hereafter, \emph{$f$ will always refer to the correlation function}.

\subsubsection{Gradient Covariance back-propagation}\label{section:gradient_backprop}
\paragraph{Gradient back-propagation for FFNN.}
The gradient back-propagation is given by 
\begin{align*}
    \frac{\partial \mathcal{L}}{\partial y^l_i} &= \phi'(y^l_i) \sum_{j=1}^{N_{l+1}} \frac{\partial \mathcal{L}}{\partial y^{l+1}_j} W^{l+1}_{ji}.
\end{align*}
where $\mathcal{L}$ is some loss function.
Using the Gradient Independence \ref{section:gradient_independence}, we have as in \cite{samuel}
\begin{equation*}
    \Tilde{q}^l(x) = \Tilde{q}^{l+1}(x) \frac{N_{l+1}}{N_l} \chi(q^l(x)).
\end{equation*}
where $\chi(q^l(x)) = \sigma_w^2 \mathbb{E}[\phi(\sqrt{q^l(x)}Z)^2]$.

\paragraph{Gradient Covariance back-propagation for CNN.} We have that
$$
\frac{\partial{\mathcal{L}}}{\partial W^l_{i,j, \beta}} = \sum_{\alpha} \frac{\partial{\mathcal{L}}}{\partial y^l_{i,\alpha}} \phi(y^{l-1}_{j, \alpha+\beta}) 
$$
Moreover,
$$
\frac{\partial{\mathcal{L}}}{\partial y^l_{i, \alpha}} = \sum_{j = 1}^n \sum_{\beta \in ker} \frac{\partial{\mathcal{L}}}{\partial y^{l+1}_{j,\alpha-\beta}} W^{l+1}_{i,j,\beta} \phi'(y^l_{i,\alpha}).
$$

Using the Gradient Independence \ref{section:gradient_independence}, and taking the average over the number of channels we have that 
$$
\mathbb{E}\left[\frac{\partial{\mathcal{L}}}{\partial y^l_{i, \alpha}}^2\right] = \frac{\sigma_w^2 \mathbb{E}\left[\phi'(\sqrt{q^l_{\alpha}(x)}Z)^2\right]}{2k+1} \sum_{\beta \in ker} \mathbb{E}\left[\frac{\partial{\mathcal{L}}}{\partial y^{l+1}_{i, \alpha-\beta}}^2\right].
$$
We can get similar recursion to that of the FFNN case by summing over $\alpha$ and using the periodic boundary condition, this yields
$$
\sum_{\alpha} \mathbb{E}\left[\frac{\partial{\mathcal{L}}}{\partial y^l_{i, \alpha}}^2\right] = \chi(q^l_{\alpha}(x)) \sum_{\alpha} \mathbb{E}\left[\frac{\partial{\mathcal{L}}}{\partial y^{l+1}_{i, \alpha}}^2\right].
$$

% Recall the two types of activation functions considered in the paper: 
% \begin{definition2}
% Let $\phi : \mathbb{R} \rightarrow \mathbb{R}$ be a measurable function. Then
% \begin{enumerate}
%     \item $\phi$ is said to be ReLU-like if there exist $\lambda, \beta \in \mathbb{R}$ such that $\phi(x)=\lambda x$ for $x>0$ and $\phi(x)=\beta x$ for $x\leq0$.
%     \item $\phi$ is said to be in $\mathcal{S}$ if $\phi(0)=0$, $\phi$ has polynomial growth, and $\phi \in \mathcal{C}^4(\mathbb{R}^d)$.
% \end{enumerate}
% \end{definition2}

\subsubsection{Some facts on the theory of information propagation}\label{subsection:results_mf}

The results presented here are based on an initialization of the model with $w^l_{ij}, b^l_i \stackrel{iid}\sim \mathcal{N}(0, 1)$ and under the following conditions:
\begin{itemize}
    \item The input data is a subset of a compact set $E$ of $\mathbb{R}^d$, and no two inputs are co-linear.
    \item All calculations are done in the limit of infinitely wide networks.\\
\end{itemize}
These results are compiled from previous papers and recalled here for the sake of completeness.
\paragraph{Results for FFNN with Tanh activation.}
\begin{fact}\label{fact:convergence_variance_ffnn_tanh}
For any choice of $\sigma_b, \sigma_w \in \mathbb{R}^+$, there exist $q, \lambda>0$ such that for all $l\geq 1$, $\sup_{x \in \mathbb{R}^d} |q^{l}(x,x)-q| \leq e^{-\lambda l}$. (Eq. (3) and conclusion right after in \cite{samuel}).
\end{fact}

\begin{fact}\label{fact:convergence_correlation_ffnn_ordered_tanh}
In the Ordered phase, there exists $\gamma >0$ such that $\sup_{x, x' \in \mathbb{R}^d} |c^{l}(x,x')-1| \leq e^{-\gamma l}$. (Eq. (8) in \cite{samuel})
\end{fact}

\begin{fact}\label{fact:convergence_correlation_ffnn_eoc_tanh}
Let $(\sigma_b, \sigma_w)\in$ EOC. Using the same notation as in fact \ref{fact:convergence_correlation_ffnn_chaotic_tanh}, we have that $\sup_{(x,x')\in B_\epsilon} |1 - c^l(x,x')| = \mathcal{O}(l^{-1})$. (Proposition 3 in \cite{hayou}).
\end{fact}

\begin{fact}\label{fact:convergence_correlation_ffnn_chaotic_tanh}
Let $B_{\epsilon} = \{(x,x') \in \mathbb{R}^d : c^1(x,x') < 1-\epsilon\}$. In the chaotic phase, there exist $c <1$ such that for all $\epsilon \in (0,1)$, there exists $\gamma >0$ such that $\sup_{(x,x') \in B_\epsilon} |c^{l}(x,x')-c| \leq e^{-\gamma l}$. (Eqs. (8) and (9) in \cite{samuel})
\end{fact}

\begin{fact}[Correlation function]\label{fact:correlation_function_tanh}
The correlation function $f$ is defined by 
$$f(x) = \frac{\sigma_b^2 + \sigma_w^2 \mathbb{E}[\phi(\sqrt{q} Z_1) \phi(\sqrt{q} (xZ_1 + \sqrt{1 - x^2} Z_2))]}{q}$$ where $q$ is given in Fact \ref{fact:convergence_variance_ffnn_tanh} and $Z_1, Z_2$ are iid standard Gaussian variables. 
\end{fact}

\begin{fact}\label{fact:derivatives_f_tanh}
$f$ has a derivative of any order $j \geq 1$ given by
$$f^{(j)}(x) = \sigma_w^2 q^{j-1} \mathbb{E}[\phi^{(j)}(Z_1)\phi^{(j)}(xZ_1 + \sqrt{1 - x^2} Z_2)], \quad \forall x \in [-1,1]$$ 
As a result, we have that $f^{(j)}(1) = \sigma_w^2 q^{j-1} \mathbb{E}[\phi^{(j)}(Z_1)^2] > 0$ for all $j\geq 1$.\\
\end{fact}
The proof of the previous fact is straightforward following the same integration by parts technique as in the proof of lemma 1 in \cite{hayou}. The result follows by induction.

\begin{fact}\label{fact:taylor_exp_tanh}
Let $(\sigma_b, \sigma_w)\in$ EOC. We have that $f'(1)=1$ (by definition of EOC). As a result, the Taylor expansion of $f$ near 1 is given by
\begin{equation*}
    f(c) = c+ \alpha (1-c)^2- \zeta (1-c)^3 +  O((1-c)^{4}).
\end{equation*}
where $\alpha, \zeta>0$.
\end{fact}
\begin{proof}
The proof is straightforward using fact \ref{fact:derivatives_f_tanh}, and integral-derivative interchanging.
\end{proof}

\paragraph{Results for FFNN with ReLU activation.}
\begin{fact}\label{fact:variance_relu_ordered}
The ordered phase for ReLU is given by $\textrm{Ord}=\{ (\sigma_b,\sigma_w) \in (\mathbb{R}^+)^2 :\sigma_w<\sqrt{2}\}$. Moreover, for any $(\sigma_b,\sigma_w) \in \mathrm{Ord}$, there exist $\lambda$ such that for all $l\geq 1$, $\sup_{x \in \mathbb{R}^d} |q^{l}(x,x)-q| \leq e^{-\lambda l}$, where $q = \frac{\sigma_b^2}{1 - \sigma_w^2 /2}$.
\end{fact}
The proof is straightforward using Eq. \eqref{equation:covariance_prop_ffnn}.
\begin{fact}\label{fact:convergence_correlation_relu_ordered}
For any $(\sigma_b,\sigma_w)$ in the Ordered phase, there exist $\lambda$ such that for all $l\geq 1$, $\sup_{(x,x') \in \mathbb{R}^d} |c^{l}(x,x')-1| \leq e^{-\lambda l}$. 
\end{fact}
The proof of this claim follows from standard Banach Fixed point theorem in the same fashion as for Tanh in \cite{samuel}.

\begin{fact}\label{fact:variance_relu_chaotic}
The Chaotic phase for ReLU is given by $\textrm{Ch}=\{ (\sigma_b,\sigma_w) \in (\mathbb{R}^+)^2 :\sigma_w>\sqrt{2}\}$. Moreover, for any $(\sigma_b,\sigma_w) \in \mathrm{Ch}$, for all $l\geq 1$, $x \in \mathbb{R}^d, q^{l}(x,x) \gtrsim (\sigma_w^2 / 2)^l$.
\end{fact}
The variance explodes exponentially in the Chaotic phase, which means the output of the Neural Network can grow arbitrarily in this setting. Hereafter, when no activation function is mentioned, and when we choose "$(\sigma_b, \sigma_w)$ in the Ordered/Chaotic phase", it should be interpreted as "$(\sigma_b, \sigma_w)$ in the Ordered phase" for ReLU and "$(\sigma_b, \sigma_w)$ in the Ordered/Chaotic phase" for Tanh.
\begin{fact}\label{fact:variance_relu_eoc}
For ReLU FFNN in the EOC, we have that $q^{l}(x,x) = \frac{\sigma_w^2}{d} ||x||^2$ for all $l \geq 1$.
\end{fact}
The proof is straightforward using Eq. \ref{equation:covariance_prop_ffnn} and that $(\sigma_b,\sigma_w)=(0,\sqrt{2})$ in the EOC.

\begin{fact}\label{fact:correlatin_function_relu}
The EOC of ReLU is given by the singleton $\{(\sigma_b, \sigma_w) = (0, \sqrt{2})\}$. In this case, the correlation function of an FFNN with ReLU is given by
$$
f(x) = \frac{1}{\pi}(x\arcsin{x} + \sqrt{1-x^2}) + \frac{1}{2}x
$$ (Proof of Proposition 1 in \cite{hayou}).
\end{fact}

\begin{fact}\label{fact:convergence_correlation_ffnn_eoc_relu}
Let $(\sigma_b, \sigma_w)\in$ EOC. Using the same notation as in fact \ref{fact:convergence_correlation_ffnn_chaotic_tanh}, we have that $$\sup_{(x,x')\in B_\epsilon} |1 - c^l(x,x')| = \mathcal{O}(l^{-2})$$ (Follows straightforwardly from Proposition 1 in \cite{hayou}).
\end{fact}

\begin{fact}\label{fact:taylor_expansion_relu}
We have that
\begin{equation}
f(c) = c +  s (1 - c)^{3/2} + b (1-c)^{5/2} + O((1-c)^{7/2}
\end{equation}
with $s=\frac{2\sqrt{2}}{3 \pi}$ and $b=\frac{\sqrt{2}}{30\pi}$.
\end{fact}
This result was proven in \cite{hayou} (in the proof of Proposition 1) for order $3/2$, the only difference is that here we push the expansion to order $5/2$.

\paragraph{Results for CNN with Tanh activation function.}
\begin{fact}\label{fact:convergence_variance_cnn_tanh}
For any choice of $\sigma_b, \sigma_w \in \mathbb{R}^+$, there exist $q, \lambda>0$ such that for all $l\geq 1$, $\sup_{\alpha, \alpha'}\sup_{x \in \mathbb{R}^d} |q^{l}_{\alpha, \alpha'}(x,x)-q| \leq e^{-\lambda l}$. (Eq. (2.5) in \cite{xiao_cnnmeanfield} and variance convergence result in \cite{samuel}).
\end{fact}
The behaviour of the correlation $c^l_{\alpha,\alpha'}(x,x')$ was studied in \cite{xiao_cnnmeanfield} only in the case $x'=x$. We give a comprehensive analysis of the asymptotic behaviour of $c^l_{\alpha,\alpha'}(x,x')$ in the next section. 

\paragraph{General results on the correlation function.}
\begin{fact}\label{fact:properties_correlation_function}
Let $f$ be either the correlation function of Tanh or ReLU. We have that  
\begin{itemize}
    \item $f(1)=1$ (lemma 2 in \cite{hayou}).
    \item In the ordered phase $0<f'(1)<1$ (By definition).
    \item In the Chaotic phase $f'(1)>1$ (By definition).
    \item In the EOC, $f'(1)=1$ (By definition).
    \item In the Ordered phase and the EOC, $1$ is the unique fixed point of $f$ (\cite{hayou}).
    \item In the Chaotic phase, $f$ has two fixed points, $1$ which is unstable, and $c<1$ which is a stable fixed point \cite{samuel}.
\end{itemize}
\end{fact}

\begin{fact}\label{fact:convergence_derivative_f_ordered_chaotic_ffnn}
Let $\epsilon \in (0,1)$. In the Ordered/Chaotic phase, with either ReLU or Tanh, there exists $\alpha \in (0,1)$,$\gamma >0$ such that 
$$
\sup_{(x,x') \in B_\epsilon} |f'(c^l(x,x')) - \alpha| \leq  e^{- \gamma l}
$$
\end{fact}
\begin{proof}
This result follows from a simple first order expansion inequality. For Tanh with $(\sigma_b,\sigma_w)$ in the Ordered phase, we have that 
$$
\sup_{(x,x') \in B_\epsilon} |f'(c^l(x,x')) - f'(1)|  \leq \zeta_l  \sup_{(x,x') \in B_\epsilon} |c^l(x,x') - 1|
$$
where $\zeta_l = \sup_{t \in (\min_{(x,x' \in B_\epsilon)}c^l(x,x'), 1)} |f''(t)| \rightarrow |f''(1)|$. We conclude for Ordered phase with Tanh using fact \ref{fact:convergence_correlation_ffnn_ordered_tanh}. The same argument can be used for Chaotic phase with Tanh using fact \ref{fact:convergence_correlation_ffnn_chaotic_tanh}; in this case, $\alpha = f'(c)$ where $c$ is the unique stable fixed point of the correlation function $f$.\\

In the Ordered phase with ReLU, let $\Tilde{f}$ be the correlation function. It is easy to see that $\Tilde{f}'(c) = \frac{\sigma_w^2}{2} f'(c)$ where $f$ is given in fact \ref{fact:correlatin_function_relu}. 
$f'(x)=1 - \frac{\sqrt{2}}{\pi} (1-x)^{1/2} + \mathcal{O}((1-x)^{3/2})$. Therefore, there exists $l_0, \zeta>0$ such that for $l>l_0$,
$$
\sup_{(x,x') \in B_\epsilon} |f'(c^l(x,x')) - f'(1)|  \leq \zeta  \sup_{(x,x') \in B_\epsilon} |c^l(x,x') - 1|^{1/2}
$$
We conclude using fact \ref{fact:convergence_correlation_relu_ordered}.
\end{proof}

\section{Technical lemmas}\label{app:technical_lemmas}
In this section we provide a number of technical lemmas used to control certain types of bounds. 

\subsection{A technical lemma for the derivation of uniform bounds}\label{app:uniform_bound_lemma}
Results in theorem \ref{thm:ntk_eoc} and \ref{thm:ntk_residual} and Proposition \ref{prop:ordered_chaotic_ntk} involve a supremum over the set $B_\epsilon$. To obtain such results, we need a 'uniform' Taylor analysis of the correlation $c^l(x,x')$ (see the next section) where uniformity is over $(x,x') \in B_{\epsilon}$. It turns out that such result is trivial when the correlation follows a dynamical system that is controlled by a non-decreasing function. We clarify this in the next lemma.

\begin{lemma2}[Uniform Bounds]\label{lemma:uniform_bounds}
Let $A \subset \mathbb{R}$ be a compact set and $g$ a non-decreasing function on $A$. Define the sequence $\zeta_l$ by $\zeta_l = g(\zeta_{l-1})$ and $\zeta_0 \in A$. Assume that there exist $\alpha_l, \beta_l$ that do not depend on $\zeta_0$ (in the sense that $\alpha_l, \beta_l$ are the same for all $\zeta_0 \in A$), with $\beta_l = o(\alpha_l)$, such that for all $\zeta_0 \in A$, 
$$
\zeta_l = \alpha_l + \bigO_{\zeta_0}(\beta_l)
$$
where $\bigO_{\zeta_0}$ means that the $\bigO$ bound depends on $\zeta_0$. Then, we have that 
$$
\sup_{\zeta_0 \in A} |\zeta_l - \alpha_l| = \bigO(\beta_l)
$$
i.e. we can choose the bound $\bigO$ to be independent of $\zeta_0$.
\end{lemma2}
\begin{proof}
Let $\zeta_{0,m}=\min A$ and $\zeta_{0,M}=\max A$. Let $(\zeta_{m,l})$ and $(\zeta_{M,l})$ be the corresponding sequences. Since $g$ is non-decreasing, we have that for all $\zeta_0 \in A$, $\zeta_{m,l} \leq \zeta_{l} \leq \zeta_{M,l}$. Moreover, by assumption, there exists $M_1, M_2 > 0$ such that 
$$
|\zeta_{m,l}- \alpha_l | \leq M_1 |\beta_l|
$$ and
$$
|\zeta_{M,l}- \alpha_l | \leq M_2 |\beta_l|
$$
therefore, 
$$
|\zeta_{l}- \alpha_l | \leq \max(|\zeta_{m,l}- \alpha_l |, |\zeta_{M,l}- \alpha_l |) \leq \max(M_1, M_2) |\beta_l|
$$
which concludes the proof.
\end{proof}
Note that Appendix Lemma \ref{lemma:uniform_bounds} can be easily extended to Taylor expansions with `$o$' instead of `$\bigO$'. We will use this result in the proofs, by refereeing to Appendix Lemma \ref{lemma:uniform_bounds}.

\subsection{On the convergence of sequences under recursive formulations}
In this section we describe three types of convergence of sequences $(a_l)_l$ verifying some recursive inequalities. 

The next lemma is used in the proof of Proposition \ref{prop:ordered_chaotic_ntk}.

\begin{lemma2}\label{lemma:uppercounded_seq}
Let $(a_l)$ be a sequence of non-negative real numbers such that $\forall l\geq 0, a_{l+1} \leq \alpha a_l + k e^{-\beta l}$, where $\alpha \in (0,1)$ and $k, \beta>0$. Then there exists $\gamma>0$ such that $\forall l\geq 0, \,  a_l \leq e^{-\gamma l}$.
\end{lemma2}

\begin{proof}
Using the inequality on $a_l$, we can easily see that 
\begin{align*}
a_l &\leq a_0 \alpha^l + k \sum_{j=0}^{l-1} \alpha^j e^{-\beta (l-j)} \\
&\leq a_0 \alpha^l + k \frac{l}{2} e^{-\beta l/2} + k\frac{l}{2} \alpha^{l/2}
\end{align*}
where we divided the sum into two parts separated by index $l/2$ and upper-bounded each part. The existence of $\gamma$ is straightforward. 
\end{proof}

The next lemma gives uniform bounds for a special type of sequences where the uniform bound here is over the initial point in the sequence.

\begin{lemma2}\label{lemma:convergence_sequence_a_b_lambda}
Let $A, B, \Lambda \subset \mathbb{R+}$ be three compact sets, and $(a_l), (b_l), (\lambda_l)$ be three sequences of non-negative real numbers such that for all $(a_0,b_0,\lambda_0) \in A\times B\times \Lambda$
\begin{equation*}
  a_l = a_{l-1} \lambda_l + b_l, \quad \lambda_l = 1 - \frac{\alpha}{l} + \mathcal{O}(l^{-1-\beta}),\quad
  b_l = q(b_0) + o(l^{-1}),
\end{equation*}
where $\alpha \in \mathbb{N}^*$ independent of $a_0,b_0,\lambda_0$, $q(b_0)\geq 0 $ is a limit that depends on $b_0$, and $\beta \in (0,1)$.\\
Assume the `$\mathcal{O}$' and `$o$' depend only on $A, B, \Lambda \subset \mathbb{R}$. Then, we have
\begin{equation*}
    \sup_{(a_0,b_0,\lambda_0) \in A\times B\times \Lambda}\left|\frac{a_l}{l} - \frac{q}{1+\alpha}\right| = \mathcal{O}(l^{-\beta}).
\end{equation*}

\end{lemma2}
\begin{proof}
Let $A, B, \Lambda \subset \mathbb{R}$ be three compact sets and $(a_0,b_0,\lambda_0) \in A\times B\times \Lambda$.
It is easy to see that there exists a constant $G>0$ independent of $a_0,b_0,\lambda_0$ such that $|a_l| \leq G \times l + |a_0|$ for all $l\geq0$. Letting $r_l = \frac{a_l}{l}$, we have that for $l\geq 2$
\begin{align*}
    r_l &= r_{l-1} (1 - \frac{1}{l})(1 - \frac{\alpha}{l} + \mathcal{O}(l^{-1-\beta})) + \frac{q}{l} + o(l^{-2}) \\
    &= r_{l-1}(1 - \frac{1+\alpha}{l}) + \frac{q}{l} + \mathcal{O}(l^{-1-\beta}).
\end{align*}
where $\mathcal{O}$ bound depends only on $A, B, \Lambda$.
Letting $x_l = r_l - \frac{q}{1+\alpha} $, there exists $M>0$ that depends only on $A, B, \Lambda$, and $l_0>0$ that depends only on $\alpha$ such that for all $l \geq l_0$
\begin{align*}
    x_{l-1}(1 - \frac{1+\alpha}{l}) - \frac{M}{l^{1+\beta}}\leq x_l &\leq x_{l-1} (1 - \frac{1+\alpha}{l}) + \frac{M}{l^{1+\beta}}.
\end{align*}
Let us deal with the right hand inequality first. By induction, we have that
$$
x_l \leq x_{l_0-1} \prod_{k=l_0}^l(1-\frac{1+\alpha}{k}) + M \sum_{k=l_0}^l \prod_{j=k+1}^l(1 - \frac{1+\alpha}{j}) \frac{1}{k^{1+\beta}}.
$$

By taking the logarithm of the first term in the right hand side and using the fact that $\sum_{k=l_0}^l \frac{1}{k} = \log(l) + O(1)$, we have
$$
\prod_{k=l_0}^l(1-\frac{1+\alpha}{k})= \Theta(l^{-1-\alpha}).
$$
where the bound $\Theta$ does not depend on $l_0$.
For the second part, observe that 
\begin{align*}
    \prod_{j=k+1}^l(1 - \frac{1+\alpha}{j}) = \frac{(l-\alpha-1)!}{l!} \frac{k!}{(k-\alpha-1)!} 
\end{align*}
and 
$$
\frac{k!}{(k-\alpha-1)!} \frac{1}{k^{1+\beta}}\sim_{k \rightarrow \infty} k^{\alpha - \beta}.
$$
Since $\alpha\geq 1$ ($\alpha \in \mathbb{N}^*$), then the serie with term $k^{\alpha-\beta}$ is divergent and we have that 
\begin{align*}
\sum_{k=l_0}^l \frac{k!}{(k-\alpha-1)!} \frac{1}{k^{2}} &\sim \sum_{k=1}^l k^{\alpha - \beta}\\
&\sim \int_{1}^l t^{\alpha -\beta} dt\\
&\sim \frac{1}{\alpha-\beta+1} l^{\alpha-\beta+1}.
\end{align*}
Therefore, it follows that
\begin{align*}
    \sum_{k=l_0}^l \prod_{j=k+1}^l(1 - \frac{1+\alpha}{j}) \frac{1}{k^{1+\beta}} &= \frac{(l-\alpha-1)!}{l!} \sum_{k=l_0}^l \frac{k!}{(k-\alpha-1)!} \frac{1}{k^{1+\beta}}\\
    &\sim \frac{1}{\alpha } l^{-\beta}.
\end{align*}
This proves that 
$$
x_l \leq \frac{M}{\alpha} l^{-\beta} + o(l^{-\beta}).
$$
where the `$o$' bound depends only on $A, B, \Lambda$. Using the same approach for the left-hand inequality, we prove that 
$$
x_l \geq -\frac{M}{\alpha} l^{-\beta} + o(l^{-\beta}).
$$
This concludes the proof.

\end{proof}

The next lemma is a different version of the previous lemma and will be useful for other applications. 

\begin{lemma2}\label{lemma:upper_bound_eoc}
Let $A, B, \Lambda \subset \mathbb{R+}$ be three compact sets, and $(a_l), (b_l), (\lambda_l)$ be three sequences of non-negative real numbers such that for all $(a_0,b_0,\lambda_0) \in A\times B\times \Lambda$
\begin{align*}
   a_l &= a_{l-1} \lambda_l + b_l, \quad  b_l = q(b_0) + \mathcal{O}(l^{-1}), \\
   \lambda_l &= 1 - \frac{\alpha}{l} + \kappa \frac{\log(l)}{l^2}+ \mathcal{O}(l^{-2}),
\end{align*}
where $\alpha \in \mathbb{N}^*, \kappa\neq0$ both do not depend on $a_0,b_0,\Lambda_0$, $q(b_o) \in \mathbb{R}^+ $ is a limit that depends on $b_0$.\\
Assume the `$\mathcal{O}$' and `$o$' depend only on $A, B, \Lambda \subset \mathbb{R}$. Then, we have
\begin{equation*}
     \sup_{(a_0,b_0,\lambda_0) \in A\times B\times \Lambda} \left|\frac{a_l}{l} - \frac{q}{1+\alpha} \right| = \Theta(\log(l)l^{-1})
\end{equation*}

\end{lemma2}

\begin{proof}
Let $A, B, \Lambda \subset \mathbb{R}$ be three compact sets and $(a_0,b_0,\lambda_0) \in A\times B\times \Lambda$. Similar to the proof of Appendix Lemma \ref{lemma:convergence_sequence_a_b_lambda}, there exists a constant $G>0$ independent of $a_0,b_0,\lambda_0$ such that $|a_l| \leq G \times l + |a_0|$ for all $l\geq0$, therefore $(a_l/l)$ is bounded. Let $r_l = \frac{a_l}{l}$. We have
\begin{align*}
    r_l &= r_{l-1} (1 - \frac{1}{l})(1 - \frac{\alpha}{l} + \kappa \frac{\log(l)}{l^2} + \mathcal{O}(l^{-2})) + \frac{q}{l} + \mathcal{O}(l^{-2}) \\
    &= r_{l-1}(1 - \frac{1+\alpha}{l}) + r_{l-1} \kappa \frac{\log(l)}{l^2} + \frac{q}{l} + \mathcal{O}(l^{-2}).
\end{align*}
Let $x_l = r_l - \frac{q}{1+\alpha} $. It is clear that $\lambda_l = 1 - \alpha/l + \mathcal{O}(l^{-3/2})$. Therefore, using Appendix Lemma \ref{lemma:convergence_sequence_a_b_lambda} with $\beta=1/2$, we have $r_l \rightarrow \frac{q}{1+\alpha}$ uniformly over $a_0,b_0,\lambda_0$. Thus, assuming $\kappa >0$ (for $\kappa<0$, the analysis is the same), there exists $\kappa_1, \kappa_2, M, l_0>0$ that depend only on $A, B, \Lambda$ such that for all $l\geq l_0$ 
\begin{align*}
    x_{l-1} (1 - \frac{1+\alpha}{l}) + \kappa_1 \frac{\log(l)}{l^2} - \frac{M}{l^2} \leq x_l \leq x_{l-1} (1 - \frac{1+\alpha}{l}) + \kappa_2 \frac{\log(l)}{l^2} + \frac{M}{l^2}.
\end{align*}
It follows that 
$$
x_l \leq x_{l_0} \prod_{k=l_0}^l(1-\frac{1+\alpha}{k}) +  \sum_{k=l_0}^l \prod_{j=k+1}^l(1 - \frac{1+\alpha}{j}) \frac{\kappa_2 \log(k) + M}{k^2} 
$$
and 
$$
x_l \geq x_{l_0} \prod_{k=l_0}^l(1-\frac{1+\alpha}{k}) + \sum_{k=l_0}^l \prod_{j=k+1}^l(1 - \frac{1+\alpha}{j}) \frac{\kappa_1 \log(k) - M}{k^2}. 
$$

Recall that we have
$$
\prod_{k=l_0}^l(1-\frac{1+\alpha}{k}) = \Theta(l^{-1-\alpha})
$$
and 
\begin{align*}
    \prod_{j=k+1}^l(1 - \frac{1+\alpha}{j}) = \frac{(l-\alpha-1)!}{l!} \frac{k!}{(k-\alpha-1)!} 
\end{align*}
so that
$$
\frac{k!}{(k-\alpha-1)!} \frac{\kappa_1 \log(k) - M}{k^2}\sim_{k \rightarrow \infty} \log(k) k^{\alpha -1}.
$$
Therefore, we obtain
\begin{align*}
\sum_{k=l_0}^l \frac{k!}{(k-\alpha-1)!} \frac{\kappa_1 \log(k) - M}{k^2} &\sim \sum_{k=1}^l \log(k) k^{\alpha -1}\\
&\sim \int_{1}^l \log(t) t^{\alpha - 1} dt\\
&\sim C_1 l^{\alpha} \log(l),
\end{align*}
where $C_1>0$ is a constant. Similarly, there exists a constant $C_2>0$ such that 
\begin{align*}
\sum_{k=1}^l \frac{k!}{(k-\alpha-1)!} \frac{\kappa_2 \log(k) + M}{k^2} &\sim C_2 l^{\alpha} \log(l).
\end{align*}
Moreover, having that $\frac{(l-\alpha-1)!}{l!} \sim l^{-1 - \alpha}$ yields
$$
x_l \leq C' l^{-1} \log(l) + o(l^{-1} \log(l))
$$
where $C'$ and `$o$' depend only on $A, B, \Lambda$.
Using the same analysis, we get 
$$
x_l \geq C'' l^{-1} \log(l) + o(l^{-1} \log(l))
$$
where $C''$ and `$o$' depend only on $A, B, \Lambda$, which concludes the proof.

\end{proof}

\subsection{Technical Lemmas for FFNN and CNN}

\begin{lemma2}[Asymptotic behaviour of $c^l$ for ReLU]\label{lemma:asymptotic_expnasion_correlation_relu_ffnn}
Let $(\sigma_b, \sigma_w) \in EOC$ and $\epsilon \in (0,1)$. There exist universal constants $\kappa,\kappa', \kappa'' > 0$ (that do not depend on any parameter) such that
    $$
    \sup_{(x,x') \in B_\epsilon}\left|c^l(x,x') - 1 + \frac{\kappa}{l^2} - \kappa' \frac{\log(l)}{l^3} \right|= \mathcal{O}(l^{-3})
    $$
    and,
    $$
\sup_{(x,x') \in B_\epsilon} \left|f'(c^l(x,x')) - 1 + \frac{3}{l} - \kappa'' \frac{\log(l)}{l^2}  \right| = \mathcal{O}(l^{-2}).
$$
\end{lemma2}

\begin{proof}
Let $(x,x') \in B_\epsilon$ and $s=\frac{2 \sqrt{2}}{3 \pi}$. From the preliminary results, we have that $\lim\limits_{l \rightarrow \infty} \sup_{x,x' \in \mathbb{R}^d} 1 - c^l(x,x') = 0$ (fact \ref{fact:convergence_correlation_ffnn_eoc_relu}). Using fact \ref{fact:taylor_expansion_relu}, we have uniformly over $B_{\epsilon}$,

$$    \gamma_{l+1} = \gamma_{l} - s \gamma^{3/2}_{l} - b \gamma^{5/2}_{l} + O(\gamma^{7/2}_{l})$$
where $s, b>0$, this yields
\begin{align*}
    \gamma^{-1/2}_{l+1} = \gamma^{-1/2}_{l} + \frac{s}{2} + \frac{3s^2}{8} \gamma_l^{1/2} + (\frac{b}{2} + \frac{5}{16}s^3 )\gamma_l +  O(\gamma_{l}^{3/2}).
\end{align*}
Thus, letting $b'=\frac{b}{2} + \frac{5}{16}s^3 $, as $l$ goes to infinity
\begin{equation*}
    \gamma^{-1/2}_{l+1} - \gamma^{-1/2}_l \sim \frac{s}{2},
\end{equation*}
and by summing and equivalence of positive divergent series
\begin{equation*}
    \gamma^{-1/2}_{l} \sim \frac{s}{2} l.
\end{equation*}
Moreover, since $ \gamma^{-1/2}_{l+1} = \gamma^{-1/2}_{l} + \frac{s}{2} + \frac{3 s^2}{8} \gamma_l^{1/2} + b' \gamma_l+ O(\gamma_{l}^{3/2})$, using the same argument multiple times and inverting the formula yields
$$c^l(x,x') = 1 - \frac{\kappa}{l^2} + \kappa' \frac{\log(l)}{l^3} + \mathcal{O}(l^{-3}) $$ where $\kappa = \frac{9 \pi^2}{2}$.
Note that, by Appendix Lemma \ref{lemma:uniform_bounds} (Section \ref{app:uniform_bound_lemma}), the $\mathcal{O}$ bound can be chosen in a way that it does not depend on $(x,x')$, it depends only on $\epsilon$; this concludes the proof for the first part of the result.\\
Using fact \ref{fact:correlatin_function_relu}, we have that
\begin{align*}
    f'(x) &= \frac{1}{\pi} \arcsin(x) + \frac{1}{2}\\
    &= 1 - \frac{\sqrt{2}}{\pi} (1 - x)^{1/2} + O((1-x)^{3/2}).
\end{align*}
Thus, it follows that
$$
f'(c^l(x,x')) = 1 - \frac{3}{l} +  \kappa'' \frac{\log(l)}{l^2} + \mathcal{O}(l^{-2}).
$$
for some universal constant $\kappa''$ uniformly over the set $B_\epsilon$, which concludes the proof.\\
\end{proof}

We prove a similar result for an FFNN with Tanh activation.

\begin{lemma2}[Asymptotic behaviour of $c^l$ for Tanh]\label{lemma:asymptotic_expnasion_correlation_tanh_ffnn}
Let $(\sigma_b, \sigma_w) \in EOC$ and $\epsilon \in (0,1)$. We have 
    $$
    \sup_{(x,x') \in B_\epsilon}\left|c^l(x,x') - 1 + \frac{\kappa}{l} - \kappa (1 - \kappa^2 \zeta) \frac{\log(l)}{l^3} \right|= \mathcal{O}(l^{-3})
    $$
    where $\kappa = \frac{2}{f''(1)}>0$ and $\zeta = \frac{f^{3}(1)}{6}>0$. Moreover, we have that 
    $$
\sup_{(x,x') \in B_\epsilon} \left|f'(c^l(x,x')) - 1 + \frac{2}{l} - 2(1 - \kappa^2 \zeta) \frac{\log(l)}{l^2}  \right| = \mathcal{O}(l^{-2}).
$$
\end{lemma2}
\begin{proof}
Let $(x,x') \in B_{\epsilon}$ and $\lambda_l := 1 - c^l(x,x')$. Using a Taylor expansion of $f$ near 1 (fact \ref{fact:taylor_exp_tanh}), there exist $\alpha, \zeta >0$ such that
\begin{align*}
\lambda_{l+1} &= \lambda_l - \alpha \lambda_l^2 + \zeta \lambda_l^3 + O(\lambda_l^{4}) 
\end{align*}
Here also, we use the same technique as in the previous lemma. We have that
\begin{align*}
\lambda_{l+1}^{-1} &= \lambda_l^{-1}(1 - \alpha \lambda_l + \zeta \lambda^2+ O(\lambda_l^{3}))^{-1} = \lambda_l^{-1}(1 + \alpha \lambda_l + (\alpha^2 -\zeta) \lambda_l^2+ O(\lambda_l^{3})) \\
&= \lambda_l^{-1} + \alpha + (\alpha^2 - \zeta) \lambda_l + O(\lambda_l^{2}). \\
\end{align*}
By summing (divergent series), we have that $\lambda_l^{-1} \sim \alpha l$. Therefore,
\begin{align*}
\lambda_{l+1}^{-1}- \lambda_l^{-1} - \alpha &= (\alpha^2 - \beta) \alpha^{-1} l^{-1} + o(l^{-1})
\end{align*}
By summing a second time, we obtain
$$
\lambda_l^{-1} = \alpha l + (\alpha- \beta \alpha^{-1}) \log(l) + o(\log(l)), 
$$
Using the same technique once again, we obtain
$$
\lambda_l^{-1} = \alpha l +(\alpha- \beta \alpha^{-1}) \log(l) + O(1). 
$$
This yields
$$\lambda_l = \alpha^{-1} l^{-1} - \alpha^{-1}(1 - \alpha^{-2} \beta) \frac{\log(l)}{l^2} + O(l^{-2}).$$
In a similar fashion to the previous proof, we can force the upper bound in $\mathcal{O}$ to be independent of $x$ using Appendix Lemma \ref{lemma:uniform_bounds}. This way, the bound depends only on $\epsilon$. This concludes the first part of the proof.\\

For the second part, observe that $f'(x) = 1 + (x-1)f''(1) + O((x-1)^2)$, hence $$f'(c^l(x,x')) = 1 - \frac{2}{l} + 2(1 - \alpha^{-2} \zeta )\frac{\log(l)}{l^2} + O(l^{-2})$$
which concludes the proof.
\end{proof}

\subsection{Large depth behaviour of the correlation in CNNs}
For CNNs, the infinite-width will always mean the limit of infinite number of channels. Recall that, by definition, $\hat{q}^{l}_{\alpha, \alpha'}(x,x') = \sigma_b^2 + \sigma_w^2 \mathbb{E}[\phi(y^{l-1}_{1,\alpha}(x))\phi(y^{l-1}_{1,\alpha'}(x'))]$ and  $q^l_{\alpha,\alpha'}(x,x') = 
\mathbb{E}[y^l_{i,\alpha}(x)y^l_{i,\alpha'}(x')]$. 

Unlike FFNN, neurons in the same channel are correlated since they share the same filters. Let $x,x'$ be two inputs and $\alpha, \alpha'$ two nodes in the same channel $i$. Using Central Limit Theorem in the limit of large $n_l$ (number of channels), we have
$$
q^l_{\alpha,\alpha'}(x,x') = 
\mathbb{E}[y^l_{i,\alpha}(x)y^l_{i,\alpha'}(x')] =   \frac{\sigma_w^2}{2k + 1} \sum_{\beta \in ker} \mathbb{E}[\phi(y^{l-1}_{1,\alpha + \beta}(x)) \phi(y^{l-1}_{1,\alpha' + \beta}(x'))] + \sigma_b^2
$$

Let $c^l_{\alpha, \alpha'}(x,x')$ be the corresponding correlation. Since $q^l_{\alpha,\alpha}(x,x)$ converges exponentially to $q$ which  depends neither on $x$ nor on $\alpha$, the mean-field correlation as in \cite{samuel, hayou} is given by 
$$
c^l_{\alpha,\alpha'}(x,x') =  \frac{1}{2k + 1} \sum_{\beta \in ker} f(c^{l-1}_{\alpha+\beta, \alpha'+\beta}(x,x'))
$$

where $f(c) =  \frac{ \sigma_w^2 \mathbb{E}[\phi(\sqrt{q}Z_1) \phi(\sqrt{q}(c Z_1 + \sqrt{1 - c^2}Z_2))] + \sigma_b^2}{q} $ and $Z_1, Z_2$ are independent standard normal variables. The dynamics of $c^l_{\alpha,\alpha'}$ become similar to those of $c^l$ in an FFNN under assumption \ref{assumption:cnn}. We show this in the proof of Appendix Lemma \ref{lemma:correlation_convergence_convnet_tanh}.
\\
In \cite{xiao_cnnmeanfield}, authors studied only the limiting behaviour of correlations $c^l_{\alpha, \alpha'}(x,x)$  (same input $x$), however, they do not study $c^l_{\alpha, \alpha'}(x,x')$ when $x\neq x'$. We do this in the following lemma, which will prove also useful for the main results of the paper. 

%%%Before moving to the proof, recall the definition of two classes of activation functions.

\begin{lemma2}[Asymptotic behaviour of the correlation in CNN with Tanh]\label{lemma:correlation_convergence_convnet_tanh}
We consider a CNN with Tanh activation function. Let $(\sigma_b, \sigma_w) \in (\mathbb{R}^+)^2$ and $\epsilon \in (0,1)$. Let $B_{\epsilon} = \{(x,x') \in \mathbb{R}^d : \sup_{\alpha,\alpha'} c^1_{\alpha,\alpha'}(x,x') < 1-\epsilon\}$. The following statements hold 
\begin{enumerate}
    \item If $(\sigma_b, \sigma_w)$ are in the Ordered phase, then there exists $\beta>0$ such that
    $$
    \sup_{(x,x') \in \mathbb{R}^d} \sup_{\alpha,\alpha'} |c^l_{\alpha,\alpha'}(x,x') - 1| = \mathcal{O}(e^{-\beta l})
    $$
    \item If $(\sigma_b, \sigma_w)$ are in the Chaotic phase, then for all $\epsilon>0$ there exists $\beta>0$ and $c\in(0,1)$ such that
    $$
    \sup_{(x,x') \in B_\epsilon} \sup_{\alpha,\alpha'} |c^l_{\alpha,\alpha'}(x,x') - c| = \mathcal{O}(e^{-\beta l})
    $$
    \item Under Assumption \ref{assumption:cnn}, if $(\sigma_b, \sigma_w) \in EOC$, then we have
    $$
    \sup_{(x,x') \in B_\epsilon} \sup_{\alpha, \alpha'}\left|c^l_{\alpha, \alpha'}(x,x') - 1 + \frac{\kappa}{l} - \kappa (1 - \kappa^2 \zeta) \frac{\log(l)}{l^3} \right|= \mathcal{O}(l^{-3})
    $$
    where $\kappa = \frac{2}{f''(1)}>0$, $\zeta = \frac{f^{3}(1)}{6}>0$, and $f$ is the correlation function given in Fact \ref{fact:correlation_function_tanh}.
\end{enumerate}

\end{lemma2}
We prove statements $1$ and $2$ for general inputs, i.e. without using Assumption \ref{assumption:cnn}. The third statement requires Assumption \ref{assumption:cnn}. 
\begin{proof}
Let $(x,x') \in \mathbb{R}^d$. Without using assumption \ref{assumption:cnn}, we have that
$$
c^l_{\alpha,\alpha'}(x,x') =  \frac{1}{2k + 1} \sum_{\beta \in ker} f(c^{l-1}_{\alpha+\beta, \alpha'+\beta}(x,x'))
$$
Writing this in matrix form yields
$$
C_l = \frac{1}{2k+1} U f(C_{l-1})
$$
where $C_l = ((c^l_{\alpha,\alpha+ \beta}(x,x'))_{\alpha \in [0:N-1]})_{\beta \in [0: N-1]}$ is a vector in $\mathbb{R}^{N^2}$, $U$ is a convolution matrix and $f$ is applied element-wise. As an example, for $k=1$, $U$ is given by
$$
U = 
\begin{bmatrix}
1 & 1 & 0 & ... & 0 & 1\\
1  & 1 & 1& 0 & \ddots & 0 \\
0  & 1 & 1 & 1  & \ddots & 0 \\
0  & 0 & 1 & 1 & \ddots & 0 \\
 & \ddots & \ddots & \ddots & \ddots &  \\
1 & 0 & \hdots & 0 & 1 &1\\
\end{bmatrix}
$$
For general $k$, $U$ is a Circulant symmetric matrix  with eigenvalues $\lambda_1>\lambda_2 \geq \lambda_3 ... \geq \lambda_{N^2}$. The largest eigenvalue of $U$ is given by $\lambda_1 = 2k+1$ and its equivalent eigenspace is generated by the vector $e_1 = \frac{1}{N} (1,1,...,1) \in \mathbb{R}^{N^2}$. This yields
$$
(1 + 2k)^{-l} U^l = e_1 e_1^T + O(e^{-\beta l})
$$
where $\beta = \log( \frac{\lambda_1}{\lambda_2})$.\\
This provides another justification to Assumption 1; as $l$ grows, and assuming that $C_l \rightarrow e_1$ (which we show in the remainder of this proof), $C_l$ exhibits a self-averaging property since $C_l \approx \frac{1}{2k+1} U C_{l-1}$. This system concentrates around the average value of the entries of $C_l$ as $l $ grows. Since the variances converge to a constant $q$ as $l$ goes to infinity (fact \ref{fact:convergence_variance_cnn_tanh}), this approximation implies that the entries of $C_l$ become almost equal as $l$ goes to infinity, thus making assumption \ref{assumption:cnn} almost satisfied in deep layers. Let us now prove the statements.
\begin{enumerate}
    \item Let $(\sigma_b, \sigma_w)$ be in the Ordered phase, $(x,x') \in \mathbb{R}^d$ and $c^l_m = \min_{\alpha,\alpha'} c^l_{\alpha, \alpha'}(x,x')$. Using the fact that $f$ is non-decreasing, we have that $c^l_{\alpha,\alpha'}(x,x') \geq \frac{1}{2k + 1} \sum_{\beta \in ker} c^{l-1}_{\alpha+\beta, \alpha'+\beta}(x,x')) \geq c^{l-1}_m$. Taking the minimum again over $\alpha, \alpha'$, we have $c^l_m \geq c^{l-1}_m$, therefore $c^l_m$ is non-decreasing and converges to the unique fixed point of $f$ which is $c=1$. This proves that $\sup_{\alpha,\alpha'} |c^l_{\alpha,\alpha'}(x,x') - 1| \rightarrow 0$. Moreover, the convergence rate is exponential using the fact that (fact \ref{fact:properties_correlation_function}) $0<f'(1) < 1$. To see this, observe that 
    $$
    \sup_{\alpha,\alpha'} |1 - c^{l}_{\alpha, \alpha'}(x,x')| \leq \left(\sup_{\zeta \in [c^{l-1}_m, 1]} f'(\zeta)\right) \times  \sup_{\alpha,\alpha'}|1 - c^{l}_{\alpha, \alpha'}(x,x') |
    $$
    Knowing that $\sup_{\zeta \in [c^{l-1}_m, 1]} f'(\zeta) \rightarrow f'(1)<1$, we conclude. 
    Moreover, the convergence is uniform in $(x,x')$ since the convergence rate depends only on $f'(1)$. 
    \item Let $\epsilon \in (0,1)$. In the chaotic phase, the only difference is the limit $c=c_1 <1$ and the Supremum is taken over $B_\epsilon$ to avoid points where $c^1(x,x') = 1$. In the Chaotic phase (fact \ref{fact:properties_correlation_function}), $f$ has two fixed points, 1 is an unstable fixed point and $c_1 \in (0,1)$ which is the unique stable fixed point. We conclude by following the same argument.
    
    \item Let $\epsilon \in (0,1)$ and $(\sigma_b,\sigma_w) \in$ EOC. Using the same argument of monotony as in the previous cases and that $f$ has 1 as unique fixed point, we have that $\lim_{l\rightarrow \infty} \sup_{x,x'} \sup_{\alpha,\alpha'}|1- c^l_{\alpha,\alpha'}(x,x')|=0$. \\
    From fact \ref{fact:taylor_exp_tanh}, the Taylor expansion of $f$ near 1 is given by
    \begin{equation*}
    f(c) = c+ \alpha (1-c)^2 - \zeta (1-c)^3 +  \mathcal{O}((1-c)^{4}).
\end{equation*}
where $\alpha = \frac{f''(1)}{2}$ and $\zeta = \frac{f^{(3)}(1)}{6}$. Using fact \ref{fact:derivatives_f_tanh}, we know that $f^{(k)}(1) = \sigma_w^2 q^{k-1} \mathbb{E}[\phi^{(k)}(\sqrt{q}Z)^2]$. Therefore, we have $\alpha >0$, and $ \zeta < 0$.\\
Under assumption \ref{assumption:cnn}, it is straightforward that for all $\alpha, \alpha'$, and $l\geq 1$
$$
c^l_{\alpha, \alpha'}(x,x') = c^l(x,x')
$$
i.e. $c^l_{\alpha, \alpha'}$ are equal for all $\alpha, \alpha'$. The dynamics of $c^l(x,x')$ are exactly the dynamics of the correlation in an FFNN. We conclude using Appendix Lemma \ref{lemma:asymptotic_expnasion_correlation_tanh_ffnn}.

% Since $\lim_{l\rightarrow \infty} \sup_{x,x'} \sup_{\alpha,\alpha'}|1- c^l_{\alpha,\alpha'}(x,x')|=0$, then, uniformly over $B_\epsilon$, we have that
% $$
% C_l = \frac{1}{2k+1} U (C_{l-1} + \alpha (1 - C_{l-1})^{2} +\zeta (1 - C_{l-1})^{3} + \mathcal{O}((1 - C_{l-1})^4) )
% $$
% where the vector sum and power is applied element-wise, i.e.  for some vector $v \in \mathbb{R}^m, v^k = (v_i^k)_{1\leq i \leq m}$ and $1 - v = (1-v_i)_{1\leq i \leq m}$.\\
% Let $A_l = ((2k+1)^{-1} U)^{-l}$ and  $\Lambda_l = A_l(1 -  C_l)$, then we have 
% $$
% \Lambda_l = \Lambda_{l-1} - \alpha A_l^{-1} \Lambda_{l-1}^2 - \zeta A_l^{-2} \Lambda_{l-1}^3 + \mathcal{O}(A_l^{-3} \Lambda_{l-1}^4)
% $$
% where $E^k$ should be understood as a matrix multiplication when $E$ is a matrix, and element-wise multiplication when $E$ is a vector.

% Using Taylor expansion multiple times, and using $A_l^{-m} = e_1e_1^T  + O(e^{-\beta l})$ for all $m\geq1$, we obtain uniformly over $B_\epsilon$
% $$
% \Lambda_l \sim (\alpha^{-1} l^{-1} - \alpha^{-1} \zeta \frac{\log(l)}{l^2} + \mathcal{O}(l^{-2})) e_1 
% $$

% This yields
% $$
% C_l = (1 - \alpha^{-1} l^{-1} + \alpha^{-1} \zeta \frac{\log(l)}{l^2} )e_1  + \mathcal{O}(l^{-2}) 
% $$
% uniformly over $B_\epsilon$, which concludes the proof.
\end{enumerate}
\end{proof}

It is straightforward that the previous Appendix Lemma extend to ReLU activation, with slightly different dynamics. In this case, we use Appendix Lemma \ref{lemma:asymptotic_expnasion_correlation_relu_ffnn} to conclude for the third statement.
\begin{lemma2}[Asymptotic behaviour of the correlation in CNN with ReLU-like activation functions]\label{lemma:correlation_convergence_convnet_relu}
We consider a CNN with ReLU activation. Let $(\sigma_b, \sigma_w) \in (\mathbb{R}^+)^2$. Let $(\sigma_b, \sigma_w) \in (\mathbb{R}^+)^2$ and $\epsilon \in (0,1)$. The following statements hold 
\begin{enumerate}
    \item If $(\sigma_b, \sigma_w)$ are in the Ordered phase, then there exists $\beta>0$ such that
    $$
    \sup_{(x,x') \in \mathbb{R}^d} \sup_{\alpha,\alpha'} |c^l_{\alpha,\alpha'}(x,x') - 1| = \mathcal{O}(e^{-\beta l})
    $$
    \item If $(\sigma_b, \sigma_w)$ are in the Chaotic phase, then there exists $\beta>0$ and $c\in(0,1)$ such that
    $$
    \sup_{(x,x') \in B_\epsilon} \sup_{\alpha,\alpha'} |c^l_{\alpha,\alpha'}(x,x') - c| = \mathcal{O}(e^{-\beta l})
    $$
    \item Under Assumption \ref{assumption:cnn}, if $(\sigma_b, \sigma_w) \in EOC$, then 
    $$
    \sup_{(x,x') \in B_\epsilon}\sup_{\alpha, \alpha'}\left|c^l(x,x') - 1 + \frac{\kappa}{l^2} - \kappa' \frac{\log(l)}{l^3} \right|= \mathcal{O}(l^{-3})
    $$
    where $\kappa, \kappa' >0$ are universal constants.
\end{enumerate}
\end{lemma2}
\begin{proof}
The proof is similar to the case of Tanh in Appendix Lemma \ref{lemma:correlation_convergence_convnet_tanh}. The only difference is that we use Appendix Lemma \ref{lemma:asymptotic_expnasion_correlation_relu_ffnn} to conclude for the third statement.
\end{proof}

\subsection{A technical lemma for ResNet}
% In this section, we provide proofs for lemmas \ref{lemma:resnet_ntk} and \ref{lemma:resnet_cnn_ntk} together with Theorem \ref{thm:spectral_decomposition_Sd} and proposition \ref{prop:scaled_resnet} on ResNets. 

Now we prove a result on the asymptotic behaviour of $c^l$ for a ResNet architecture. \\

\begin{lemma2}[Asymptotic expansion of $c^l$ for ResNet]\label{lemma:asymptotic_expansion_resnet}
Let $\epsilon \in (0,1)$ and $\sigma_w>0$. We have for FFNN
    $$
    \sup_{(x,x') \in B_\epsilon}\left|c^l(x,x') - 1 + \frac{\kappa_{\sigma_w}}{l^2} - \kappa_{\sigma_w}' \frac{\log(l)}{l^3} \right|= \mathcal{O}(l^{-3})
    $$
    where $\kappa_{\sigma_w}, \kappa_{\sigma_w}'>0$ are two constants that depend on $\sigma_w$. \\
    Moreover, we have that 
    $$
\sup_{(x,x') \in B_\epsilon} \left|f'(c^l(x,x')) - 1 + \frac{3 (1 + \frac{2}{\sigma_w^2})}{l} - \kappa_{\sigma_W}'' \frac{\log(l)}{l^2}  \right| = \mathcal{O}(l^{-2}).
$$ where $f$ is the ReLU correlation function given in fact \ref{fact:correlatin_function_relu} and $\kappa_{\sigma_W}''>0$ is a constant that depends on $\sigma_w$.\\
Moreover, this result holds also for CNNs where the supremum should be replaced by $\sup_{(x,x') \in B_\epsilon} \sup_{\alpha, \alpha'}$.
\end{lemma2}
\begin{proof}
We first prove the result for ResNet with fully connected layers, the extension to convolutional layers is straightforward. Let $\epsilon \in (0,1)$.
\begin{itemize}
    \item Let $x\neq x' \in \mathbb{R}^d$, and $c^l := c^l(x,x')$. It is straightforward that the variance terms follow the recursive form 
    $$q^l(x,x) = q^{l-1}(x,x) + \sigma_w^2/2 q^{l-1}(x,x) = (1 + \sigma_{w}^2/2)^{l-1} q^1(x,x)$$
    Leveraging this observation, we have that 
$$
c^{l+1} = \frac{1}{1 + \alpha} c^l + \frac{\alpha}{1  + \alpha} f(c^l), 
$$
where $f$ is the ReLU correlation function given in fact \ref{fact:correlatin_function_relu} and $\alpha = \frac{\sigma_w^2}{2}$. Recall that
$$
f(c) = \frac{1}{\pi} c \, \arcsin(c) + \frac{1}{\pi} \sqrt{1 - c^2} + \frac{1}{2} c.
$$
As in the proof of Appendix Lemma \ref{lemma:asymptotic_expnasion_correlation_relu_ffnn}, let $\gamma_l = 1 - c^l$, therefore, using Taylor expansion of $f$ near 1 given in fact \ref{fact:taylor_expansion_relu} yields
$$
\gamma_{l+1} = \gamma_{l} - \frac{\alpha s}{1 + \alpha} \gamma_l^{3/2} - \frac{\alpha b}{1 + \alpha} \gamma_l^{5/2} + O(\gamma_{l}^{7/5}).
$$

This asymptotic expansion is similar to the one in the proof of Appendix Lemma \ref{lemma:asymptotic_expnasion_correlation_relu_ffnn} with different coefficients $s' = \frac{\alpha s}{1 + \alpha}$ and $b' = \frac{\alpha b}{1 + \alpha}$. We conclude by following the same machinery. The second result also follows from this similarity as it is also a result of this asymptotic expansion.

\item Under Assumption \ref{assumption:cnn}, the result holds for CNN.

\end{itemize}

\end{proof}

The next theorem shows that no matter what the choice of $\sigma_w>0$, the normalized NTK of a ResNet will always have a subexponential convergence rate to a limiting $\bar{K}^{\infty}_{res}$.

\subsection{A technical lemma for scaled ResNet}

\begin{lemma2}\label{lemma:asymptotic_expansion_scaled_resnet}
Consider a Residual Neural Network with the following forward propagation equations
\begin{equation}
y^l(x) = y^{l-1}(x) +\frac{1}{\sqrt{l}}\mathcal{F}(w^l, y^{l-1}(x)), \quad l\geq 2.
\end{equation}
where $\mathcal{F}$ is either a convolutional or dense layer (equations \ref{equation:ffnn_net} and \ref{equation:convolutional_net}) with ReLU activation. Then there exists $\zeta, \nabla > 0$ such that for all $\epsilon \in (0,1)$
$$
\sup_{(x,x') \in B_\epsilon} \left|1 - c^l(x,x') - \frac{\zeta}{\log(l)^2} + \frac{\nabla}{\log(l)^3} \right|= o\left(\frac{1}{\log(l)^3}\right)
$$ where the bound `$o$' depends only on $\epsilon$.\\
For CNN, under Assumption \ref{assumption:cnn}, the result holds and the supremum is taken also over $\alpha, \alpha'$, i.e. 
$$
\sup_{(x,x') \in B_\epsilon} \sup_{\alpha, \alpha'} \left|1 - c^l_{\alpha, \alpha'}(x,x') - \frac{\zeta}{\log(l)^2} + \frac{\nabla}{\log(l)^3} \right|= o\left(\frac{1}{\log(l)^3}\right)
$$
\end{lemma2}

\begin{proof}
As for the previous results, we  only prove the result for a ResNet with fully-connected layers, the proof is similar for convolutional layers under \Cref{assumption:cnn}.\\

Let $\epsilon \in (0,1)$ and $(x,x') \in B_\epsilon$ be two inputs and denote by $c^l := c^l(x,x')$. Following the same machinery as in the proof of Appendix Lemma \ref{lemma:asymptotic_expansion_resnet}, we have that
$$
c^{l} = \frac{1}{1+\alpha_l} c^{l-1} + \frac{\alpha_l}{1+\alpha_l} f(c^{l-1})
$$
where $\alpha_l = \frac{\sigma_w^2}{2l}$. Using Fact \ref{fact:correlatin_function_relu}, it is straightforward that $f'\geq 0$, hence $f$ is non-decreasing. Hence, $c^{l}\geq c^{l-1}$ and $c^l$ converges to a fixed point $c$. Let us prove that $c=1$. By contradiction, suppose $c<1$ so that $f(c)-c>0$ ($f$ has a unique fixed point which is $1$). This yields
$$
c^l - c = c^{l-1} - c + \frac{f(c)-c}{l} + \mathcal{O}\left(\frac{c^l-c}{l}\right) + \mathcal{O}\left(l^{-2}\right)
$$
by summing, this leads to $c^l - c \sim (f(c)-c) \log(l)$ which is absurd since $f(c) \neq c$ ( $f$ admits only 1 as a fixed point). We conclude that $c=1$. Using the fact that $f$ is non-decreasing, it is easy to conclude that the convergence is uniform over $B_\epsilon$.\\

Now let us find the asymptotic expansion of $1 - c^l$. Recall the Taylor expansion of the correlation function $f$ near 1 given in Fact \ref{fact:taylor_expansion_relu}
\begin{equation}\label{Taylorf}
f(c) \underset{c \rightarrow 1-}{=} c + s (1-c)^{3/2} + b (1-c)^{5/2} + \mathcal{O}((1 - c)^{7/2})
\end{equation}
where $s=\frac{2\sqrt{2}}{3\pi}$ and $b= \frac{\sqrt{2}}{30 \pi}$. Letting $\gamma_l = 1 - c^l$ and $\delta_l = \frac{\alpha_l}{1+\alpha_l}$, we obtain
$$
\gamma_{l} = \gamma_{l-1} - s \delta_l \gamma_{l-1}^{3/2} -b \delta_l \gamma_{l-1}^{5/2} + \bigO(\delta_l \gamma_{l-1}^{7/5}).
$$
which yields
\begin{equation}\label{equation:taylor_expansion_gamma_l_resnet}
     \gamma^{-1/2}_{l} = \gamma^{-1/2}_{l-1} + \frac{s}{2} \delta_{l} + \frac{3}{8} s^2 \delta_l^2 \gamma_{l-1}^{1/2} + \left(\frac{b}{2} \delta_l + \frac{5}{16}\delta_l^3 s^3\right) \gamma_{l-1}+ O(\delta_l \gamma_{l-1}^{3/2}).
\end{equation}
therefore, we have that 

$$
\gamma_l^{-1/2} \sim \frac{s \sigma_w^2}{4} \log(l)
$$
and $1 - c^l \sim \frac{\zeta }{\log(l)^2}$ where $\zeta = 16/s^2\sigma_w^4$.\\

By pushing the asymptotic expansion to the next order, we obtain
$$
1 - c^l = \frac{\zeta}{\log(l)^2} - \frac{\nabla}{\log(l)^3} + o\left(\frac{1}{\log(l)^3}\right)
$$
where $\nabla>0$. the `$o$' holds uniformly for $(x,x') \in B_\epsilon$ as in the proof of Appendix Lemma \ref{lemma:asymptotic_expnasion_correlation_relu_ffnn}.\\

This result holds for a ResNet with CNN layers under Assumption \ref{assumption:cnn} since the dynamics are the same in this case.

\end{proof}

\newpage

\section{Proofs of the lemmas \ref{lemma:ffnn_ntk}, \ref{lemma:cnn_ntk}, \ref{lemma:resnet_ntk} \ref{lemma:resnet_ntk_conv}}\label{proofsSection3}
In this section, we provide proofs for for the recursive formulas satisfied by the NTK, i.e. lemmas \ref{lemma:ffnn_ntk}, \ref{lemma:cnn_ntk}, \ref{lemma:resnet_ntk} \ref{lemma:resnet_ntk_conv}.

\subsection{Proofs of non-ResNet lemmas}

Recall that lemma \ref{lemma:ffnn_ntk} in the paper is a generalization of theorem 1 in \cite{jacot} and is reminded here. The proof is simple and follows similar induction techniques as in \cite{jacot}.
\begin{lemma3} [Generalization of Th. 1 in \cite{jacot}]
Consider an FFNN of the form \eqref{equation:ffnn_net}. Then, as $n_1, n_2, ..., n_{L-1} \rightarrow \infty$, we have for all $x,x' \in \mathbb{R}^d$, $i,i' \leq n_L$, $K^L_{ii'}(x,x') = \delta_{ii'} K^L(x,x')$, where $K^L(x,x')$ is given by the recursive formula
$$
K^{L} (x,x') = \dot{q}^L(x,x') K^{L-1} (x,x') + q^L(x,x'),
$$
where $q^{l} (x,x')= \sigma_b^2 + \sigma_w^2 \mathbb{E}[\phi(y_1^{l-1}(x))\phi(y_1^{l-1}(x'))]$ and $\dot{q}^l(x,x') = \sigma_w^2 \mathbb{E}[\phi'(y_1^{l-1}(x))\phi'(y_1^{l-1}(x'))]$.
\end{lemma3}

\begin{proof}
The proof for general $\sigma_w$ is analogous to the case when $\sigma_w = 1$ (\cite{jacot}). The result is proved by by induction. We denote by $\partial_\theta y$ the gradient of $y$ with respect to $\theta$.\\

It is straightforward that 
$$
K^1_{ii'}(x,x') = \delta_{ii'} K^1(x,x')
$$
where $K^1(x,x') = \frac{\sigma_w^2}{d} x \cdot x' + \sigma_b^2$. 
For $l\geq 2$ and $i \in [1: n_l]$
\begin{align*}
    \partial_{\theta_{1:l}} y^{l+1}_i(x) &=  \frac{\sigma_w}{\sqrt{n_{l}}} \sum_{j=1}^{n_{l}} w^{l+1}_{ij} \phi'(y^{l}_{j}(x)) \partial_{\theta_{1:l}} y^{l}_j(x).
\end{align*}
and,
\begin{align*}
    (\partial_{\theta_{1:l}} y^{l+1}_i(x))(\partial_{\theta_{1:l}} y^{l+1}_{i'}(x'))^t &=  \frac{\sigma_w^2}{n_l} \sum_{j, j'}^{n_{l}} w^{l+1}_{ij} w^{l+1}_{i'j'}\phi'(y^l_{j}(x)) \phi'(y^l_{j'}(x'))\partial_{\theta_{1:l}} y^l_j(x) (\partial_{\theta_{1:l}} y^l_{j'}(x'))^t
\end{align*}
Therefore, when $l=2$, we have that 

\begin{align*}
    (\partial_{\theta_{1}} y^{2}_i(x))(\partial_{\theta_{1}} y^{2}_{i'}(x'))^t &=  \frac{\sigma_w^2}{n_1} \sum_{j, j'}^{n_{1}} w^{2}_{ij} w^{2}_{i'j'}\phi'(y^1_{j}(x)) \phi'(y^1_{j'}(x'))\partial_{\theta_{1}} y^1_j(x) (\partial_{\theta_{1}} y^1_{j'}(x'))^t\\
    &=  \frac{\sigma_w^2}{n_1} K^1(x,x') \sum_{j}^{n_{1}} w^{2}_{ij} w^{2}_{i'j}\phi'(y^1_{j}(x)) \phi'(y^1_{j'}(x')) 
\end{align*}

Taking $n_1$ to infinity, the law of large numbers yields

\begin{align*}
    (\partial_{\theta_{1}} y^{2}_i(x))(\partial_{\theta_{1}} y^{2}_{i'}(x'))^t \to  \delta_{ii'} K^1(x,x')  \dot{q}^2(x,x')
\end{align*}

We also have that for all $l \geq 2$
\begin{align*}
(\partial_{w^{l+1}} y^{l+1}_i(x))(\partial_{w^{l+1}} y^{l+1}_i(x'))^t &+  (\partial_{b^{l+1}} y^{l+1}_i(x))(\partial_{b^{l+1}} y^{l+1}_i(x'))^t = \frac{\sigma_w^2}{n_l} \sum_{j} \phi(y^l_j(x)) \phi(y^l_j(x')) + \sigma_b^2\\
&\underset{n_l \rightarrow \infty}{\rightarrow} \sigma_w^2 \mathbb{E}[\phi(y^{l}_i(x))\phi(y^{l}_i(x'))] + \sigma_b^2 = q^{l+1}(x,x').
\end{align*}

we conclude that the statement is true for $l=2$. Now assume that the result is true for $l \in {2, 3, \dots, L}$, let us prove it for $L+1$. Using the induction hypothesis, namely that  as $n_0, n_1, ..., n_{L-1} \rightarrow \infty$ sequentially, for all $j,j' \leq n_{l}$ and all $x,x'$
$$\partial_{\theta_{1:L}}y^L_j(x) (\partial_{\theta_{1:L}} y^L_{j'}(x'))^t \rightarrow K^L(x,x')\mathbf{1}_{j=j'} $$
Hence, given $n_L$, as $n_0, n_1, ..., n_{l-1} \rightarrow \infty$ sequentially, we have that
\begin{align*}
      \frac{\sigma_w^2}{n_L} \sum_{j, j'}^{n_{L}} w^{L+1}_{ij} w^{L+1}_{ij'}\phi'(y^L_{j}(x)) \phi'(y^L_{j'}(x'))&\partial_{\theta_{1:L}} y^L_j(x) (\partial_{\theta_{1:L}} y^L_{j'}(x'))^t
     \rightarrow \\
     & \frac{\sigma_w^2}{n_L} \sum_{j}^{n_{L}} (w^{L+1}_{ij})^2 \phi'(y^L_{j}(x))\phi'(y^L_j(x')) K^L(x,x') 
\end{align*}

and letting $n_L$ go to infinity, the law of large numbers implies that 
 $$
\frac{\sigma_w^2}{n_L} \sum_{j}^{n_{L}} (w^{L+1}_{ij})^2 \phi'(y^L_{j}(x)) \phi'(y^L_j(x')) K^L(x,x') \rightarrow \dot{q}^{L+1}(x,x') K^L(x,x').
 $$
 
Moreover, we have that 
\begin{align*}
(\partial_{w^{L+1}} y^{L+1}_i(x))(\partial_{w^{L+1}} y^{L+1}_i(x'))^t &+  (\partial_{b^{L+1}} y^{L+1}_i(x))(\partial_{b^{L+1}} y^{L+1}_i(x'))^t = \frac{\sigma_w^2}{n_L} \sum_{j} \phi(y^L_j(x)) \phi(y^L_j(x')) + \sigma_b^2\\
&\underset{n_L \rightarrow \infty}{\rightarrow} \sigma_w^2 \mathbb{E}[\phi(y^{L}_i(x))\phi(y^{L}_i(x'))] + \sigma_b^2 = q^{L+1}(x,x').
\end{align*}

which completes the proof.

\end{proof}

We now provide the proof of recursive formula satisfied by the NTK of a CNN. The proof techniques are similar to \Cref{lemma:ffnn_ntk}.

\begin{lemma3}[Infinite-width dynamics of the NTK of a CNN]
Consider a CNN of the form \eqref{equation:convolutional_net}, then we have that for all $x,x' \in \mathbb{R}^d$, $i,i' \leq n_1$ and  $\alpha,\alpha' \in [0: M - 1]$
$$
K^1_{(i,\alpha), (i',\alpha')}(x,x') = \delta_{ii'} \big(\frac{\sigma_w^2}{n_0(2k+1)} [x,x']_{\alpha, \alpha'} + \sigma_b^2 \big)
$$

For $l\geq2$, as $n_1, n_2, ..., n_{l-1} \rightarrow \infty$ recursively, we have for all $i,i' \leq n_l$, $\alpha,\alpha' \in [0:M - 1]$,  $K^{l}_{(i,\alpha),(i',\alpha')}(x,x') = \delta_{ii'} K^{l}_{\alpha,\alpha'}(x,x')$, where $K^{ l}_{\alpha,\alpha'}$ is given by the recursive formula 
\begin{align*}
    K^l_{\alpha,\alpha'} &= \frac{1}{2k+1} \sum_{\beta \in ker_l } \Psi^{l-1}_{\alpha+\beta,\alpha'+\beta}
\end{align*} 
where  $\Psi^{l-1}_{\alpha,\alpha'} = \dot{q}^l_{\alpha, \alpha'} K^{ l-1}_{\alpha,\alpha'}
    + \hat{q}^{l}_{\alpha, \alpha'}$, and $\hat{q}^{l}_{\alpha,  \alpha},\dot{q}^l_{\alpha, \alpha'} $ are defined in lemma \ref{lemma:ffnn_ntk}, with $y_{1,\alpha}^{l-1}(x), y_{1,\alpha'}^{l-1}(x')$ in place of $y_{1}^{l-1}(x),y_{1}^{l-1}(x')$.
\end{lemma3}

\begin{proof}

Here we omit the first step in the induction argument, i.e. to show that the result holds for $l=2$. This is straightforward and analogous to the case of $l=2$ in the proof of \Cref{lemma:ffnn_ntk}. We just illustrate small differences in the induction hypothesis for CNNs.\\

Let $x,x'$ be two inputs. We have that 

\begin{equation*}
\begin{aligned}
    y^1_{i,\alpha}(x) &= \frac{\sigma_w}{\sqrt{v_1}} \sum_{j=1}^{n_{0}} \sum_{\beta \in ker_1} w^1_{i,j,\beta} x_{j,\alpha+\beta} + \sigma_b b^1_i\\
    y^l_{i,\alpha}(x) &= \frac{\sigma_w}{\sqrt{v_l}} \sum_{j=1}^{n_{l-1}} \sum_{\beta \in ker_l} w^l_{i,j,\beta} \phi(y^{l-1}_{j,\alpha+\beta}(x)) + \sigma_b b^l_i
\end{aligned}
\end{equation*}
therefore

\begin{align*}
K^1_{(i,\alpha), (i',\alpha')}(x,x') &= \sum_{r} \left( \sum_{j}  \sum_{\beta} \frac{\partial y^1_{i,\alpha}(x)}{\partial w^1_{r,j,\beta}} \frac{\partial y^1_{i',\alpha'}(x)}{\partial w^1_{r,j,\beta}} \right) + \frac{\partial y^1_{i,\alpha}(x)}{\partial b^1_{r}} \frac{\partial y^1_{i',\alpha'}(x)}{\partial b^1_{r}}\\
&= \delta_{ii'} \left(\frac{\sigma_w^2}{n_0(2k+1)}\sum_{j} \sum_{\beta} x_{j, \alpha+\beta} x_{j, \alpha'+\beta} + \sigma_b^2 \right)
\end{align*}

Assume the result is true for $l-1$, let us prove it for $l$. Let $\theta_{1:l-1}$ be model weights and bias in the layers 1 to $l-1$. Let $\partial_{\theta_{1:l-1}} y^{l}_{i,\alpha}(x) = \frac{\partial y^l_{i,\alpha}(x)}{\partial \theta_{1:l-1}}$. We have that

\begin{align*}
\partial_{\theta_{1:l-1}} y^{l}_{i,\alpha}(x) = \frac{\sigma_w}{\sqrt{n_{l-1}(2k+1)}} \sum_j \sum_\beta w^l_{i,j,\beta} \phi'(y^{l-1}_{j,\alpha+\beta}) \partial_{\theta_{1:l-1}} y^{l-1}_{i,\alpha+\beta}(x)
\end{align*}
this yields
\begin{align*}
&\partial_{\theta_{1:l-1}} y^{l}_{i,\alpha}(x) \partial_{\theta_{1:l-1}} y^{l}_{i',\alpha'}(x)^T =\\ &\frac{\sigma_w^2}{n_{l-1}(2k+1)} \sum_{j,j'} \sum_{\beta,\beta'} w^l_{i,j,\beta} w^l_{i',j',\beta'} \phi'(y^{l-1}_{j,\alpha+\beta})\phi'(y^{l-1}_{j',\alpha'+\beta}) \partial_{\theta_{1:l-1}} y^{l-1}_{j,\alpha+\beta}(x) \partial_{\theta_{1:l-1}} y^{l-1}_{j',\alpha'+\beta}(x)^T
\end{align*}

as $n_1, n_2, ..., n_{l-2} \rightarrow \infty$ and using the induction hypothesis, we have
\begin{align*}
&\partial_{\theta_{1:l-1}} y^{l}_{i,\alpha}(x) \partial_{\theta_{1:l-1}} y^{l}_{i',\alpha'}(x)^T \rightarrow\\ &\frac{\sigma_w^2}{n_{l-1}(2k+1)} \sum_{j} \sum_{\beta,\beta'} w^l_{i,j,\beta} w^l_{i',j,\beta'} \phi'(y^{l-1}_{j,\alpha+\beta})\phi'(y^{l-1}_{j,\alpha'+\beta}) K^{l-1}_{(j,\alpha+\beta),(j,\alpha'+\beta)}(x,x')
\end{align*}

note that $K^{l-1}_{(j,\alpha+\beta),(j,\alpha'+\beta)}(x,x') = K^{l-1}_{(1,\alpha+\beta),(1,\alpha'+\beta)}(x,x')$ for all $j$ since the variables are iid across the channel index $j$.

Now letting $n_{l-1} \rightarrow \infty$, we have that 

\begin{align*}
&\partial_{\theta_{1:l-1}} y^{l}_{i,\alpha}(x) \partial_{\theta_{1:l-1}} y^{l}_{i',\alpha'}(x)^T \rightarrow\\ & \delta_{ii'} \big(\frac{1}{(2k+1)} \sum_{\beta,\beta'} \dot{q}^l_{\alpha+\beta,\alpha'+\beta} K^{l-1}_{(1,\alpha+\beta),(1,\alpha'+\beta)}(x,x')\big)
\end{align*}
We conclude using the fact that 
$$
\partial_{\theta_{l}} y^{l}_{i,\alpha}(x) \partial_{\theta_{l}} y^{l}_{i',\alpha'}(x)^T \rightarrow \delta_{ii'} (\frac{\sigma_w^2}{2k+1} \sum_{\beta} \mathbb{E}[\phi(y^{l-1}_{\alpha+\beta}(x))\phi(y^{l-1}_{\alpha'+\beta}(x'))] + \sigma_b^2 )
$$
\end{proof}

\subsection{Lemmas for ResNet NTK}

Here we provide the proofs for the recursive formulas of the NTK of a ResNet. The techniques are the same, and our goal here is to show the role of the residual term in the induction argument.

\begin{lemma3}[NTK of a ResNet with Fully Connected layers in the infinite-width limit]
Let $x,x'$ be two inputs and $K^{res,1}$ be the exact NTK for the Residual Network with 1 layer. Then, we have
\begin{itemize}
    \item For the first layer (without residual connections), we have for all $x,x' \in \mathbb{R}^d$
$$
K^{res,1}_{ii'}(x,x') = \delta_{ii'}\left( \sigma_b^2 + \frac{\sigma_w^2}{d} x\cdot x'\right),
$$
where $x\cdot x'$ is the inner product in $\mathbb{R}^d$.
\item For $l\geq2$, as $n_1, n_2, ..., n_{L-1} \rightarrow \infty$, we have for all $i,i' \in [1:n_l]$, $K^{res,l}_{ii'}(x,x') = \delta_{ii'} K^l_{res}(x,x')$, where $K^l_{res}(x,x')$ is given by the recursive formula have for all $x,x' \in \mathbb{R}^d$ and $l\geq 2$, as $n_1, n_2, ..., n_l \rightarrow \infty$ recursively, we have 
$$
K_{res}^l(x, x') = K_{res}^{l-1}(x, x') (\dot{q}^l(x, x')+1) + \hat{q}^l(x,x').
$$
\end{itemize}

\end{lemma3}

\begin{proof}
The first result is the same as in the FFNN case since we assume there is no residual connections between the first layer and the input. We prove the second result by induction.
\begin{itemize}
    \item Let $x,x' \in \mathbb{R}^d$. We have 
    $$
    K^1_{res}(x,x') = \sum_{j} \frac{\partial y^1_1(x)}{\partial w^1_{1j}}\frac{\partial y^1_1(x)}{\partial w^1_{1j}}   +  \frac{\partial y^1_1(x)}{\partial b^1_1}\frac{\partial y^1_1(x)}{\partial b^1_1} = \frac{\sigma_w^2}{d} x\cdot x' + \sigma_b^2.
    $$
    
    \item 
    
    The proof is similar to the FeedForward network NTK. For $l\geq 2$ and $i \in [1: n_l]$
\begin{align*}
    \partial_{\theta_{1:l}} y^{l+1}_i(x) &= \partial_{\theta_{1:l}} y^{l}_i(x)  + \frac{\sigma_w}{\sqrt{n_{l}}} \sum_{j=1}^{n_{l}} w^{l+1}_{ij} \phi'(y^{l}_{j}(x)) \partial_{\theta_{1:l}} y^{l}_j(x).
\end{align*}
Therefore, we obtain
\begin{align*}
    (\partial_{\theta_{1:l}} y^{l+1}_i(x))&(\partial_{\theta_{1:l}} y^{l+1}_i(x'))^t = (\partial_{\theta_{1:l}} y^l_i(x))(\partial_{\theta_{1:l}} y^l_i(x'))^t  \\
    &+ \frac{\sigma_w^2}{n_l} \sum_{j, j'}^{n_{l}} w^{l+1}_{ij} w^{l+1}_{ij'}\phi'(y^l_{j}(x)) \phi'(y^l_{j'}(x'))\partial_{\theta_{1:l}} y^l_j(x) (\partial_{\theta_{1:l}} y^l_{j'}(x'))^t +I
\end{align*}
where 
$$I = \frac{\sigma_w}{\sqrt{n_l}} \sum_{j=1}^{n_l} w^{l+1}_{ij} (\phi'(y^l_{j}(x)) \partial_{\theta_{1:l}} y^l_i(x) (\partial_{\theta_{1:l}} y^l_j(x'))^t +\phi'(y^l_j(x')) \partial_{\theta_{1:l}} y^l_j(x) (\partial_{\theta_{1:l}} y^l_i(x'))^t ).$$
Using the induction hypothesis, as $n_0, n_1, ..., n_{l-1} \rightarrow \infty$, we have that 
\begin{align*}
    & (\partial_{\theta_{1:l}} y^{l+1}_i(x))(\partial_{\theta_{1:l}} y^{l+1}_i(x'))^t  + \frac{\sigma_w^2}{n_l} \sum_{j, j'}^{n_{l}} w^{l+1}_{ij} w^{l+1}_{ij'}\phi'(y^l_{j}(x)) \phi'(y^l_{j'}(x'))\partial_{\theta_{1:l}} y^l_j(x) (\partial_{\theta_{1:l}} y^l_{j'}(x'))^t
    +I \\
    &\rightarrow K^l_{res}(x,x')  + \frac{\sigma_w^2}{n_l} \sum_{j}^{n_{l}} (w^{l+1}_{ij})^2 \phi'(y^l_{j}(x))\phi'(y^l_j(x')) K^l_{res}(x,x') 
    +I',
\end{align*}
where $I' = \frac{\sigma_w^2}{n_l} w^{l+1}_{ii} (\phi'(y^l_{i}(x)) + \phi'(y^l_i(x'))) K^l_{res}(x,x').$\\

 As $n_l \rightarrow \infty$,  we have that $I' \rightarrow 0$. Using the law of large numbers, as $n_l \rightarrow \infty$ 
 $$
\frac{\sigma_w^2}{n_l} \sum_{j}^{n_{l}} (w^{l+1}_{ij})^2 \phi'(y^l_{j}(x)) \phi'(y^l_j(x')) K^l_{res}(x,x') \rightarrow \dot{q}^{l+1}(x,x') K^l_{res}(x,x').
 $$
Moreover, we have that 
\begin{align*}
(\partial_{w^{l+1}} y^{l+1}_i(x))(\partial_{w^{l+1}} y^{l+1}_i(x'))^t &+  (\partial_{b^{l+1}} y^{l+1}_i(x))(\partial_{b^{l+1}} y^{l+1}_i(x'))^t =\\
&\frac{\sigma_w^2}{n_l} \sum_{j} \phi(y^l_j(x)) \phi(y^l_j(x')) + \sigma_b^2\\
&\underset{n_l \rightarrow \infty}{\rightarrow} \sigma_w^2 \mathbb{E}[\phi(y^{l}_i(x))\phi(y^{l}_i(x'))] + \sigma_b^2 = q^{l+1}(x,x').
\end{align*}

\end{itemize}

\end{proof}

In a similar fashion, we prove the recursive formula for ResNets with Convolutional layers.

\begin{lemma3}[NTK of a ResNet with Convolutional layers in the infinite-width limit]\label{lemma:resnet_cnn_ntk}
Let $K^{res,1}$ be the exact NTK for the ResNet with 1 layer. Then\\
$\bullet$~For the first layer (without residual connections), we have for all $x,x' \in \mathbb{R}^d$
$$
K^{res,1}_{(i,\alpha), (i',\alpha')}(x,x') = \delta_{ii'} \Big(\frac{\sigma_w^2}{n_0(2k+1)}[x,x']_{\alpha,\alpha'} + \sigma_b^2 \Big)
$$
$\bullet$~For $l\geq2$, as $n_1, n_2, ..., n_{l-1} \rightarrow \infty$ recursively, we have for all $i,i' \in [1:n_l]$, $\alpha,\alpha' \in [0:M-1]$, $K^{res,l}_{(i,\alpha),(i',\alpha')}(x,x') = \delta_{ii'} K^{res, l}_{\alpha,\alpha'}(x,x')$, where $K^{res, l}_{\alpha,\alpha'}$ is given by the recursive formula for all $x,x' \in \mathbb{R}^d$, using the same notations as in lemma \ref{lemma:cnn_ntk},
$$
K^{res, l}_{\alpha,\alpha'} = K^{res, l-1}_{\alpha,\alpha'} + \frac{1}{2k+1} \sum_{\beta} \Psi^{l-1}_{\alpha+\beta, \alpha'+\beta}.
$$
where $\Psi^l_{\alpha,\alpha'} = \dot{q}^l_{\alpha, \alpha'} K^{res, l}_{\alpha,\alpha'}+ \hat{q}^{l}_{\alpha, \alpha'}$.

\end{lemma3}

\begin{proof}
Let $x,x'$ be two inputs. We have that 
\begin{align*}
K^1_{(i,\alpha), (i',\alpha')}(x,x') &= \sum_{j} \left( \sum_{\beta} \frac{\partial y^1_{i,\alpha}(x)}{\partial w^1_{i,j,\beta}} \frac{\partial y^1_{i',\alpha'}(x)}{\partial w^1_{i',j,\beta}}  + \frac{\partial y^1_{i,\alpha}(x)}{\partial b^1_{j}} \frac{\partial y^1_{i',\alpha'}(x)}{\partial b^1_{j}}\right)\\
&= \delta_{ii'} \left(\frac{\sigma_w^2}{n_0(2k+1)}\sum_{j} \sum_{\beta} x_{j, \alpha+\beta} x_{j, \alpha'+\beta} + \sigma_b^2 \right).
\end{align*}
Assume the result is true for $l-1$, let us prove it for $l$. Let $\theta_{1:l-1}$ be model weights and bias in the layers 1 to $l-1$. Let $\partial_{\theta_{1:l-1}} y^{l}_{i,\alpha}(x) = \frac{\partial y^l_{i,\alpha}(x)}{\partial \theta_{1:l-1}}$. We have that

\begin{align*}
\partial_{\theta_{1:l-1}} y^{l}_{i,\alpha}(x) = \partial_{\theta_{1:l-1}} y^{l-1}_{i,\alpha}(x) + \frac{\sigma_w}{\sqrt{n_{l-1}(2k+1)}} \sum_j \sum_\beta w^l_{i,j,\beta} \phi'(y^{l-1}_{j,\alpha+\beta}) \partial_{\theta_{1:l-1}} y^{l-1}_{i,\alpha+\beta}(x)
\end{align*}
this yields
\begin{align*}
&\partial_{\theta_{1:l-1}} y^{l}_{i,\alpha}(x) \partial_{\theta_{1:l-1}} y^{l}_{i',\alpha'}(x)^T =\partial_{\theta_{1:l-1}} y^{l-1}_{i,\alpha}(x) \partial_{\theta_{1:l-1}} y^{l-1}_{i',\alpha'}(x)^T+\\ 
&\frac{\sigma_w^2}{n_{l-1}(2k+1)} \sum_{j,j'} \sum_{\beta,\beta'} w^l_{i,j,\beta} w^l_{i',j',\beta'} \phi'(y^{l-1}_{j,\alpha+\beta})\phi'(y^{l-1}_{j',\alpha'+\beta}) \partial_{\theta_{1:l-1}} y^{l-1}_{j,\alpha+\beta}(x) \partial_{\theta_{1:l-1}} y^{l-1}_{j',\alpha'+\beta}(x)^T + I,
\end{align*}
where 
$$
I = \frac{\sigma_w}{\sqrt{n_{l-1}(2k+1)}} \sum_{j,\beta} w^l_{i,j,\beta} \phi'(y^{l-1}_{j,\alpha+\beta}) (\partial_{\theta_{1:l-1}} y^{l-1}_{i,\alpha}(x) \partial_{\theta_{1:l-1}} y^{l-1}_{i,\alpha+\beta}(x)^T + \partial_{\theta_{1:l-1}} y^{l-1}_{i,\alpha+\beta}(x)\partial_{\theta_{1:l-1}} y^{l-1}_{i,\alpha}(x)^T).
$$
As $n_1, n_2, ..., n_{l-2} \rightarrow \infty$ and using the induction hypothesis, we have
\begin{align*}
&\partial_{\theta_{1:l-1}} y^{l}_{i,\alpha}(x) \partial_{\theta_{1:l-1}} y^{l}_{i',\alpha'}(x)^T \rightarrow \delta_{ii'} K^{l-1}_{\alpha,\alpha'} (x,x') + \\ &\frac{\sigma_w^2}{n_{l-1}(2k+1)} \sum_{j} \sum_{\beta,\beta'} w^l_{i,j,\beta} w^l_{i',j,\beta'} \phi'(y^{l-1}_{j,\alpha+\beta})\phi'(y^{l-1}_{j,\alpha'+\beta}) K^{l-1}_{(j,\alpha+\beta),(j,\alpha'+\beta)}(x,x').
\end{align*}
Note that $K^{l-1}_{(j,\alpha+\beta),(j,\alpha'+\beta)}(x,x') = K^{l-1}_{(1,\alpha+\beta),(1,\alpha'+\beta)}(x,x')$ for all $j$ since the variables are iid across the channel index $j$.
Now letting $n_{l-1} \rightarrow \infty$, we have that 
\begin{align*}
&\partial_{\theta_{1:l-1}} y^{l}_{i,\alpha}(x) \partial_{\theta_{1:l-1}} y^{l}_{i',\alpha'}(x)^T \rightarrow\\ & \delta_{ii'} K^{l-1}_{\alpha,\alpha'} (x,x') + \delta_{ii'} \left(\frac{1}{(2k+1)} \sum_{\beta,\beta'} f'(c^{l-1}_{\alpha +\beta, \alpha'+\beta}(x,x')) K^{l-1}_{(1,\alpha+\beta),(1,\alpha'+\beta)}(x,x')\right),
\end{align*}
where $f'(c^{l-1}_{\alpha +\beta, \alpha'+\beta}(x,x')) = \sigma_w^2 \mathbb{E}[\phi'(y^{l-1}_{j,\alpha+\beta})\phi'(y^{l-1}_{j,\alpha'+\beta})]$.\\
We conclude using the fact that 
$$
\partial_{\theta_{l}} y^{l}_{i,\alpha}(x) \partial_{\theta_{l}} y^{l}_{i',\alpha'}(x)^T \rightarrow \delta_{ii'} \left(\frac{\sigma_w^2}{2k+1} \sum_{\beta} \mathbb{E}[\phi(y^{l-1}_{\alpha+\beta}(x))\phi(y^{l-1}_{\alpha'+\beta}(x'))] + \sigma_b^2 \right).
$$
\end{proof}

% The following proposition establishes that any initialization on the Ordered or Chaotic phase, leads to a trivial limiting NTK as the number of layers $L$ becomes large. 
%\cite{hayou}, \cite{samuel} and \cite{pool} studied an initialization scheme known as the Edge of Chaos and proved that it maximizes the signal propagation through the neural network. However, this has not been directly linked to the training dynamics, we address\\
% \begin{prop2}[Limiting Neural Tangent Kernel with Ordered/Chaotic Initialization]
% Let $(\sigma_b, \sigma_w)$ be  either in the ordered or in the chaotic phase. Then, there exist $\lambda>0$ such that for all $\epsilon \in (0,1) $, there exists $\gamma > 0$ such that  
% $$
% \sup_{(x, x') \in B_\epsilon} |K^L(x,x') - \lambda | \leq e^{-\gamma L}.
% $$

% \end{prop2}

% \section{Further experimental results}\label{app:deterioration}
% $\bullet~$\textbf{Results for $L$ between 30 and 300}: In our experiments, we observed degeneracy of the NTK regime when $L \sim 300$ (hence our choice of $L=300$ in the paper). The Figure included here shows the percentage drop in performance of the NTK regime for a 100x100 FFNN with ReLU on MNIST for the Ordered phase/EOC Init.
% \begin{figure}[h]
%     \centering
%     \includegraphics[width=0.4\linewidth]{deterioration_of_ntk.png}
%     \caption{Deterioration of NTK regime}
%     \label{fig:deterioration}
% \end{figure}

\newpage

\newpage

\bibliography{bibliography}
\bibliographystyle{plainnat}
\end{document}